\documentclass[10pt,twocolumn,letterpaper]{article}

\usepackage[pagenumbers]{cvpr} 

\usepackage{algorithm}
\usepackage[noend]{algpseudocode}
\usepackage{amsfonts}       
\usepackage{amsmath}
\usepackage{amssymb}
\usepackage{amsthm}
\usepackage[page]{appendix} 
\usepackage{array}
\usepackage[accsupp]{axessibility}  
\usepackage{bm}
\usepackage{bbm}
\usepackage{booktabs}
\usepackage{cite}
\usepackage{color}
\usepackage{comment}
\usepackage{enumitem}
\usepackage{graphicx}
\usepackage{listings}
\usepackage{makecell}
\usepackage{microtype}      
\usepackage{newfloat}
\usepackage{nicefrac}
\usepackage{placeins}
\usepackage{pifont}
\usepackage{subcaption}
\usepackage{tabularx}
\usepackage{url}
\usepackage{verbatim}
\usepackage{xcite}
\usepackage{xcolor}
\usepackage{xpatch}
\usepackage{xr}
\usepackage{xspace}

\makeatletter
\@namedef{ver@everyshi.sty}{}
\makeatother

\usepackage{tikz}
\usetikzlibrary{automata,positioning,arrows}
\usepackage[mode=buildnew]{standalone}
\usepackage{pgfplots}
\usepackage{tikz-3dplot}
\usetikzlibrary{positioning}
\usepgfplotslibrary{groupplots,units}
\pgfplotsset{compat=1.14}

\usepackage[pagebackref,breaklinks,colorlinks]{hyperref}
\newcommand\copyrighttext{%
	\footnotesize \copyright~2022 IEEE. Personal use of this material is permitted. Permission from IEEE must be obtained for all other uses, in any current or future media, including reprinting/republishing this material for advertising or promotional purposes, creating new collective works, for resale or redistribution to servers or lists, or reuse of any copyrighted component of this work in other works.}
\newcommand\copyrightnotice{%
	\begin{tikzpicture}[remember picture,overlay]
	\node[anchor=south,yshift=10pt] at (current page.south) {{\parbox{\dimexpr\textwidth-\fboxsep-\fboxrule\relax}{\copyrighttext}}};
	\end{tikzpicture}%
}

\usepackage[capitalize]{cleveref}
\crefname{section}{Sec.}{Secs.}
\Crefname{section}{Section}{Sections}
\Crefname{table}{Table}{Tables}
\crefname{table}{Tab.}{Tabs.}
\crefname{line}{l.}{ls.}
\Crefname{line}{Line}{Lines}
\Crefname{listing}{Algorithm}{Algorithms}
\crefname{listing}{Alg.}{Algs.}
\Crefname{algorithm}{Algorithm}{Algorithms}
\crefname{algorithm}{Alg.}{Algs.}
\Crefname{thm}{Theorem}{Theorems}
\crefname{thm}{Thm.}{Thms.}
\def\cvprPaperID{8230} 
\def\confName{CVPR}
\def\confYear{2022}

\newcommand*{\mydprime}{^{\prime\prime}\mkern-1.2mu}
\newcommand*{\mytprime}{^{\prime\prime\prime}\mkern-1.2mu}

\theoremstyle{plain}
\newtheorem{thm}{\protect\theoremname}
\theoremstyle{definition}

\theoremstyle{plain}
\newtheorem{lem}[thm]{\protect\lemmaname}
\theoremstyle{plain}
\newtheorem{cor}[thm]{\protect\cornaname}
\theoremstyle{remark}

\providecommand{\theoremname}{Theorem}
\providecommand{\lemmaname}{Lemma}
\providecommand{\cornaname}{Corollary}

\DeclareMathOperator{\modulo}{mod}
\DeclareMathOperator{\supp}{supp}
\DeclareMathOperator{\spa}{span}
\DeclareMathOperator{\topp}{TOP}

\begin{document}
\makeatletter 
\newcommand{\printfnsymbol}[1]{%
	\textsuperscript{\@fnsymbol{#1}}%
} 
\makeatother
\renewcommand{\thefootnote}{\fnsymbol{footnote}}	

\stepcounter{footnote}
\stepcounter{footnote}

\makeatletter
\xpatchcmd{\paragraph}{3.25ex \@plus1ex \@minus.2ex}{3pt plus 1pt minus 1pt}{\typeout{success!}}{\typeout{failure!}}
\makeatother
	
\global\long\def\F{\mathcal{F}}%
\global\long\def\Fl{\mathcal{F}^{(l)}}%
\global\long\def\fo{\mathcal{F}_0}
\global\long\def\Fb{\mathbf{F}}%
\global\long\def\fe{\mathbf{F}^{(1)}}%
\global\long\def\B{\mathcal{B}}%
\global\long\def\bo{\mathcal{B}_0}
\global\long\def\ff{\mathbf{F}}%
\global\long\def\R{\mathbb{R}}%
\global\long\def\1{\mathbbm{1}}%
\global\long\def\co{c_{out}}%
\global\long\def\ci{c_{in}}%
\global\long\def\a{\alpha}%
\global\long\def\b{\beta}%
\global\long\def\loss{\mathcal{L}}%
\global\long\def\L{\Lambda}%
\global\long\def\Le{\Lambda^{(1)}}%
\global\long\def\Lz{\Lambda^{(2)}}%
\global\long\def\l{\lambda}%
\global\long\def\lm{\lambda_m}%
\global\long\def\lpm{\lambda_{m^\prime}}%
\global\long\def\labl{\lambda^{(\a,\b;l)}}%
\global\long\def\lab{\lambda^{(\a,\b)}}%
\global\long\def\labln{\lambda^{(\a,\b;l)}_n}%
\global\long\def\lpabln{\lambda^{(\a^\prime,\b^\prime;l^\prime)}_{n^\prime}}%
\global\long\def\labn{\lambda^{(\a,\b)}_n}%
\global\long\def\mabl{m^{(\a,\b;l)}}%
\global\long\def\mab{m^{(\a,\b)}}%
\global\long\def\muabl{\mu^{(\a,\b;l)}}%
\global\long\def\muab{\mu^{(\a,\b)}}%
\global\long\def\Nl{N_{(l)}}%
\global\long\def\gn{g^{(n)}}%
\global\long\def\gnl{g^{(n)}}%
\global\long\def\en{e^{(n)}}%
\global\long\def\g#1{g^{(#1)}}%
\global\long\def\gl#1#2{g^{(#1;#2)}}%
\global\long\def\e#1{e^{(#1)}}%
\global\long\def\kl{K_{(l)}}%
\global\long\def\col{c_{out}^{(l)}}%
\global\long\def\cil{c_{in}^{(l)}}%
\global\long\def\f{\varphi}%
\global\long\def\fpm{\varphi_{m^\prime}}%
\global\long\def\fm{\varphi_m}%
\global\long\def\lpm{\lambda_{m^\prime}}%
\global\long\def\fl{\varphi^{(l)}}
\global\long\def\hl{h^{(l)}}
\global\long\def\fabl{\varphi^{(\a,\b;l)}}
\global\long\def\fab{\varphi^{(\a,\b)}}
\global\long\def\fabn{\varphi^{(\a,\b)}_{n}}
\global\long\def\habl{h^{(\a,\b;l)}}
\global\long\def\hab{h^{(\a,\b)}}
\global\long\def\Hab{H^{(\a,\b)}}
\global\long\def\fabln{\varphi^{(\a,\b;l)}_{n}}
\global\long\def\fpabln{\varphi^{(\a^\prime,\b^\prime;l^\prime)}_{n^\prime}}
\global\long\def\habln{h^{(\a,\b;l)}_{i_n,j_n}}
\global\long\def\habin{h^{(\a,\b)}_{i_n,j_n}}
\global\long\def\habn{h^{(\a,\b)}_{n}}
\global\long\def\glo{\texttt{global}}%
\global\long\def\mod{\texttt{module} wise}%
\global\long\def\blk{\texttt{block} wise}%
\global\long\def\lay{\texttt{layer} wise}%
\global\long\def\Phil{\Phi^{(l)}}
\global\long\def\Psil{\Psi^{(l)}}
\global\long\def\sl{\sigma_{(l)}}
\global\long\def\s{\sigma}
\global\long\def\re{R^{(1)}}
\global\long\def\rz{R^{(2)}}
\global\long\def\de{D^{(1)}}
\global\long\def\adds{\texttt{ADD}\text{s}}
\global\long\def\mults{{FLOPs}}
\global\long\def\mult{{FLOP}}
\global\long\def\id{\text{id}}

\global\long\def\m{\mathbf{m}}%
\global\long\def\mp{\mathbf{m}^\prime}%
\global\long\def\md{\mathbf{m}\mydprime}%
\global\long\def\mt{\mathbf{m}\mytprime}%

\global\long\def\P{\Phi}
\global\long\def\Pa{\Phi^\ast}
\global\long\def\La{R^\ast}
\global\long\def\Ra{R^\ast}
\global\long\def\Fa{\Fb^\ast}
\global\long\def\sc{\mathcal{S}}
\global\long\def\tops{\topp_s(U)}
\global\long\def\tij{\mathcal{T}_{i,j}}
\global\long\def\up{u_+}

\renewcommand\and{\end{tabular}\kern-\tabcolsep\ and\ \kern-\tabcolsep\begin{tabular}[t]{c}}
\let\origthanks\thanks
\renewcommand\thanks[1]{\begingroup\let\rlap\relax\origthanks{#1}\endgroup}

\newcommand{\cmark}{\ding{51}}%
\newcommand{\xmark}{\ding{53}}%
\title{Interspace Pruning: Using Adaptive Filter Representations to Improve Training of Sparse CNNs}

\author{Paul Wimmer\printfnsymbol{1}\printfnsymbol{2}\thanks{Corresponding author. \printfnsymbol{4}Equal contribution.}\printfnsymbol{4}, Jens Mehnert\printfnsymbol{1}\printfnsymbol{4} and Alexandru Paul Condurache\printfnsymbol{1}\printfnsymbol{2}\\
	{\small \printfnsymbol{1}Automated Driving Research, Robert Bosch GmbH, 70469 Stuttgart, Germany }\\
	{\small \printfnsymbol{2}Institute for Signal Processing, University of L{\"u}beck, 23562 L{\"u}beck, Germany}\\
	{\tt\small \{paul.wimmer,jensericmarkus.mehnert,alexandrupaul.condurache\}@de.bosch.com}
}
\maketitle

\begin{abstract}
	Unstructured pruning is well suited to reduce the memory footprint of convolutional neural networks (CNNs), both at training and inference time. CNNs contain parameters arranged in $K \times K$ filters. Standard unstructured pruning (SP) reduces the memory footprint of CNNs by setting filter elements to zero, thereby specifying a fixed subspace that constrains the filter. 
	Especially if pruning is applied before or during training, this induces a strong bias. To overcome this, we introduce interspace pruning (IP), a general tool to improve existing pruning methods. It uses filters represented in a dynamic interspace by linear combinations of an underlying adaptive filter basis (FB). 
	For IP, FB coefficients are set to zero while un-pruned coefficients and FBs are trained jointly. In this work, we provide mathematical evidence for IP's superior performance and demonstrate that IP outperforms SP on all tested state-of-the-art unstructured pruning methods. Especially in challenging situations, like pruning for ImageNet or pruning to high sparsity, IP greatly exceeds SP with equal runtime and parameter costs. Finally, we show that advances of IP are due to improved trainability and superior generalization ability.
\end{abstract}

\section{Introduction}\label{sec:intro}\copyrightnotice
Deep neural networks (DNNs) have shown state-of-the-art (SOTA) performance in many artificial intelligence applications \cite{park_2020,yuan_2020,zoph_2020,pham_2021,wang_2021}. In order to solve these tasks, large models with up to billions of parameters are required. However, training, transferring, storing and evaluating such large models is costly \cite{schwartz_2019,strubell_2020}.    
\emph{Pruning} \cite{mozer_1989,janowsky_1989,lecun_1990,han_2015,guo_2016,gao_2021}
sets parts of the network's weights to zero. This reduces
the model's complexity and memory requirements, speeds up inference \cite{blalock_2020} and may lead to an improved generalization ability \cite{bartoldson_2019,lecun_1990,hassibi_1993}.
In recent years, \emph{training} sparse models became of interest, providing the benefits of reduced memory requirements and runtime not only for inference but also for training \cite{frankle_2018,lee_2018,mocanu_2018,ramanujan_2019,evci_2020,malach_2020,tanaka_2020,wang_2020,wimmer_2021}.  
\begin{figure*}[ht!]
	\centering
	\begin{subfigure}[b]{0.335\linewidth}
		\centering
		\includegraphics[width=\textwidth]{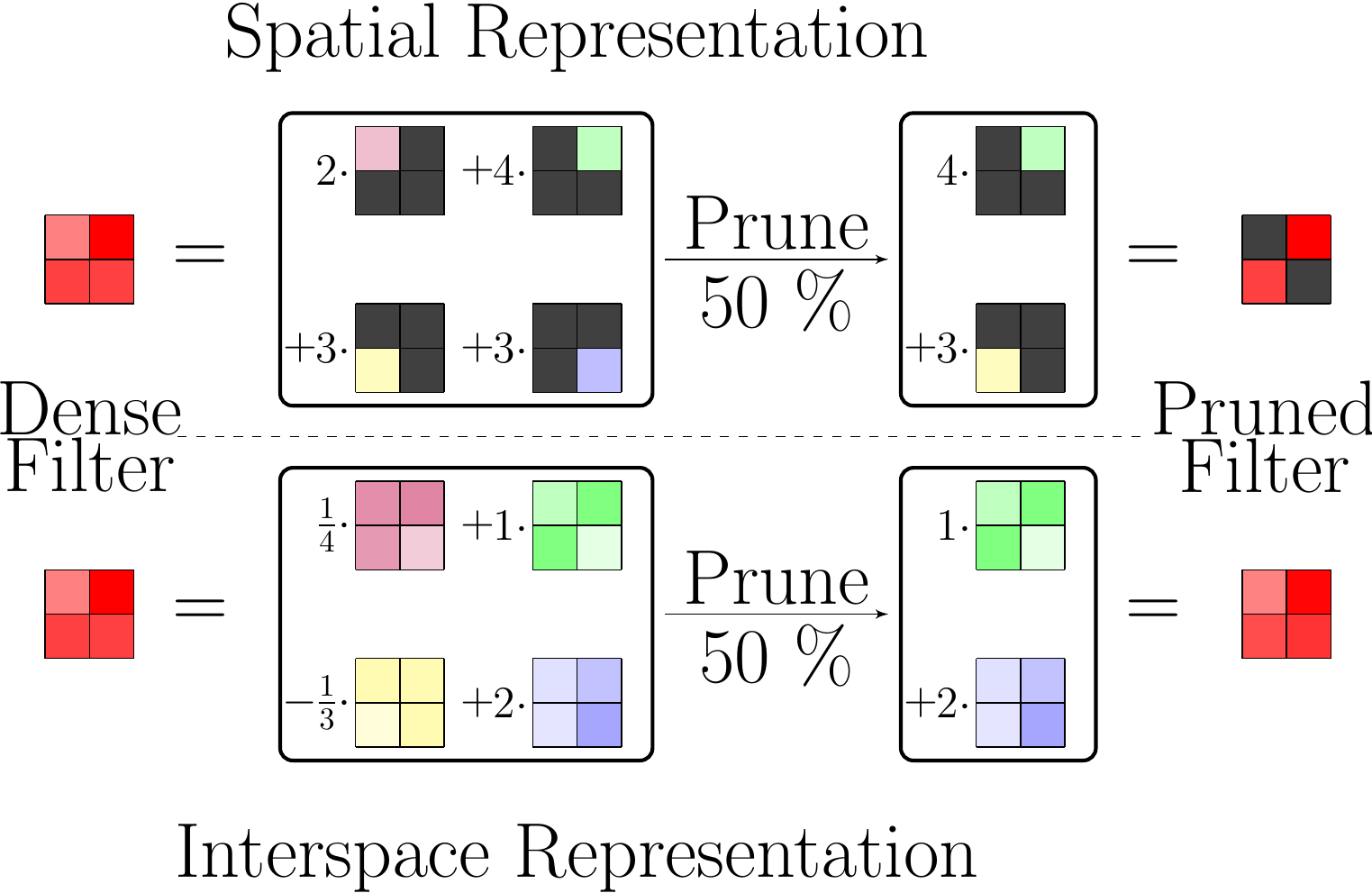}
		\caption{}
		\label{fig:pruning_scheme}
	\end{subfigure}
	\hspace{1em}
	\begin{subfigure}[b]{0.285\linewidth}
		\centering
		\includegraphics[width=\textwidth]{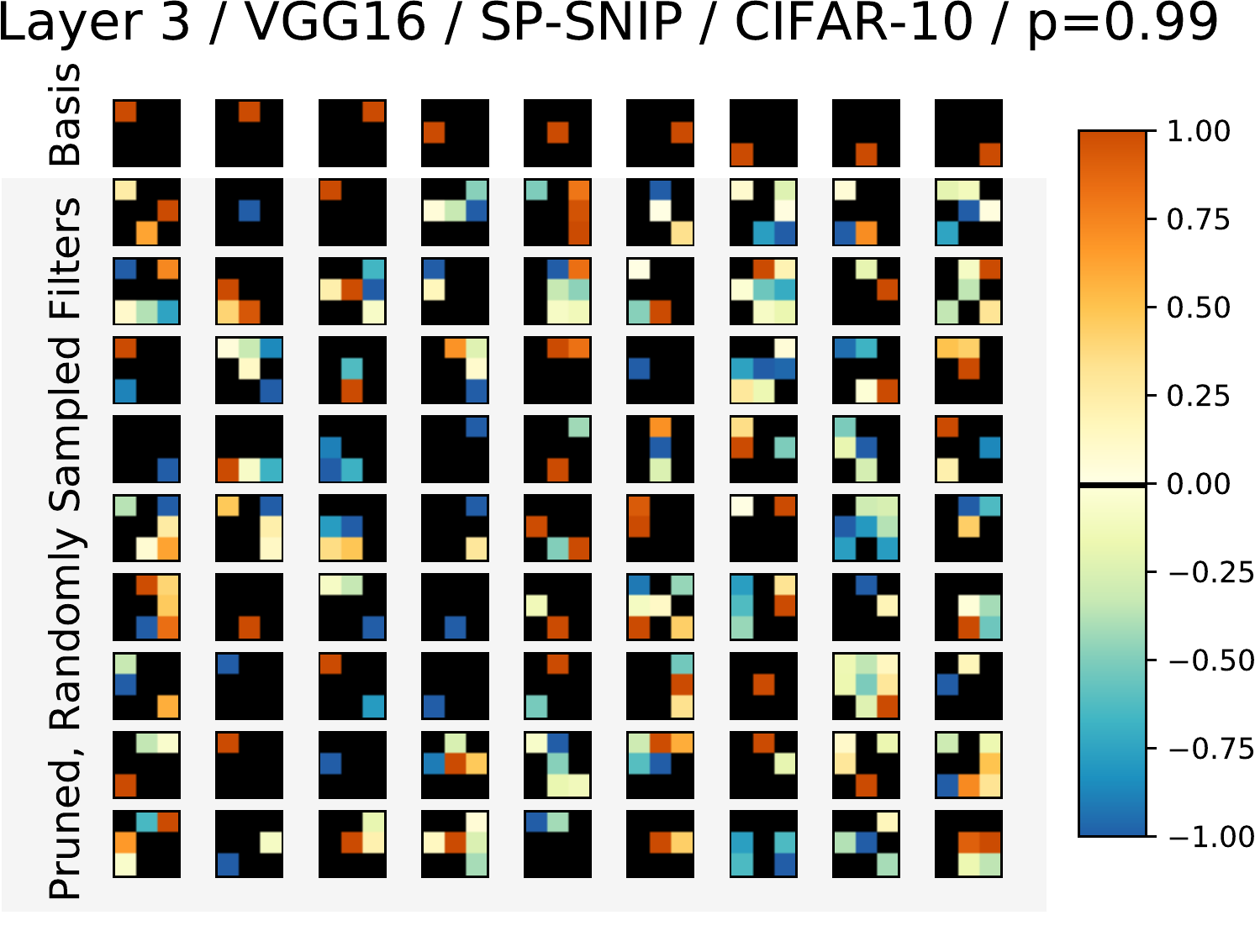}
		\caption{}
		\label{fig:c10_graphical_comparison_sp}
	\end{subfigure}
	\hspace{1em}
	\begin{subfigure}[b]{0.285\linewidth}
		\centering
		\includegraphics[width=\textwidth]{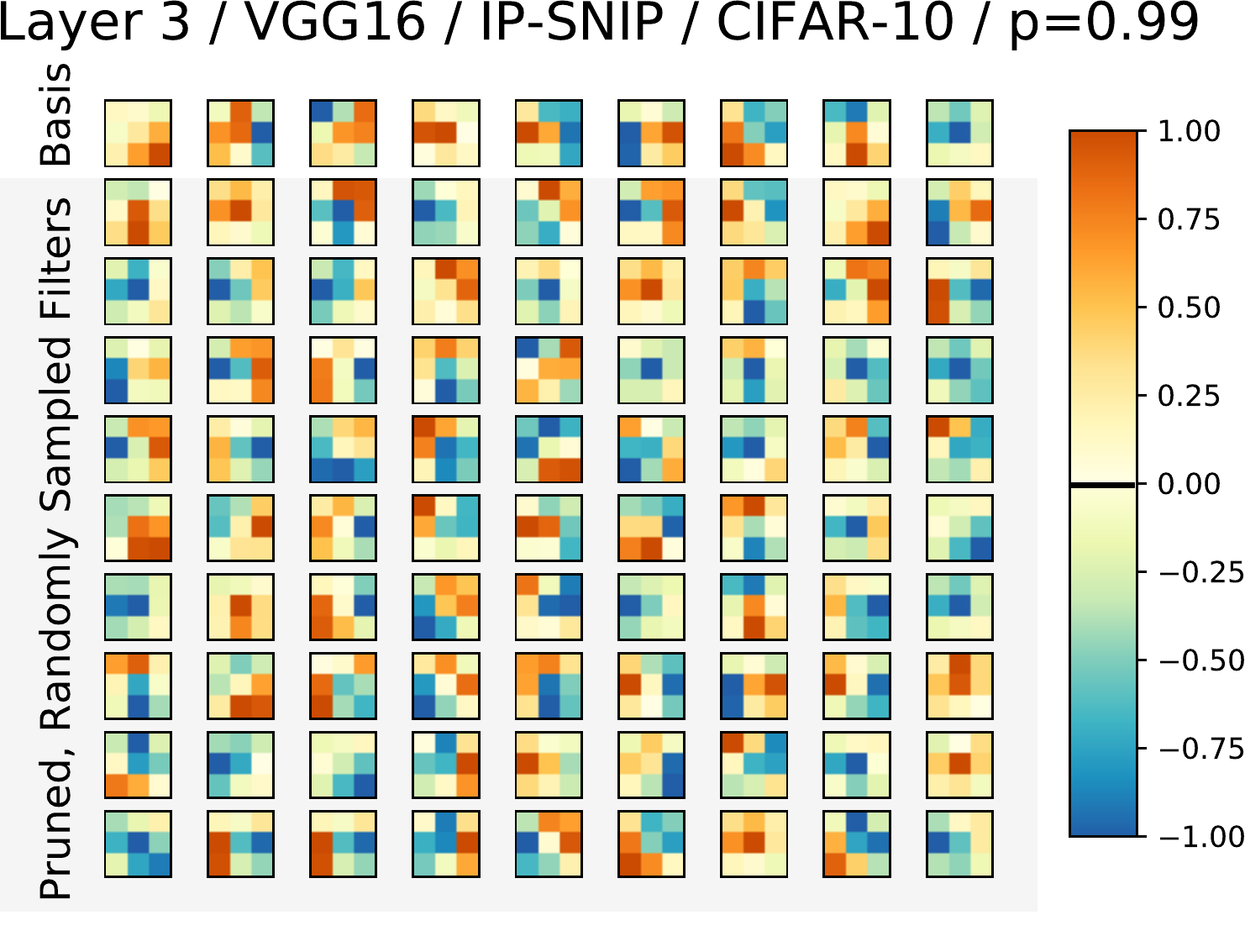}
		\caption{}
		\label{fig:c10_graphical_comparison_fbp}
	\end{subfigure}
	\caption{\subref{fig:pruning_scheme} Overview of SP and IP. Contrarily to SP \subref{fig:c10_graphical_comparison_sp}, IP \subref{fig:c10_graphical_comparison_fbp} produces spatially dense filters after training sparse networks. As for SP, sparsity in the interspace can be used to reduce memory requirements and, by the linearity of convolutions, also computational costs.} \label{fig:pruning}
\end{figure*}

In this work, we mainly focus on methods that prune individual parameters \emph{before} training, while the number of zeroed coefficients is kept fixed during training. With this \emph{unstructured pruning}, a network's memory footprint can be reduced. To lower the runtime in addition, specialized soft- and hardware is needed \cite{han_2016,parashar_2017,elsen_2020,gale_2020}. For training sparse networks, we distinguish between (i) \emph{pruning at initialization} (PaI) \cite{lee_2018,jorge_2020,tanaka_2020,wang_2020,wimmer_2021} which prunes the network at initialization and fixes zeroed parameters during training, (ii) finding the sparse architecture to be finally trained by iterative train-prune-reset cycles, a so called \emph{lottery ticket} (LT) \cite{frankle_2018,frankle_2020a}, and (iii) \emph{dynamic sparse training} (DST) \cite{mocanu_2018,evci_2020,liu_2021b} which prunes the network at initialization, but allows the pruning mask to be changed during training.

Convolutional neural networks (CNNs) are composed of layers, each having a certain number of input- and output channels. Every combination of input- and output channel is linked by a \emph{filter} $h \in \R^{K \times K}$ with \emph{kernel size} $K \times K$. A weight of $h$ is a \emph{spatial coefficient} $h_{i,j}$ for a {spatial coordinate} $(i,j)$. 
Filters $h$ can also be modeled in an \emph{interspace}, a linear space $\{\sum_{n=1}^{K^2} \l_n \cdot \gn : \l_n \in \R \}$ spanned by a \emph{filter basis} (FB) $\F := \{\g{1}, \ldots, \g{K^2}\} \subset \R^{K \times K}$ \cite{engan_1999,ullrich_2017}. One possibility for a FB is the standard basis $\B := \{\en : n=1,\ldots, K^2\}$ which yields the spatial representations. General interspace representations are more flexible since bases are not fixed. We represent $h$ in an interspace in order to learn the FB $\F$ spanning this space along with the \emph{FB coefficients} $\l$, and thereby obtain a better representation for $h$. Thus, setting coefficients of flexible, adaptive FBs to zero will improve results compared to prune spatial coefficients. 

For deep networks, where the layers' purposes are usually unknown to the experts but learnt during training, we believe that filters should train their bases along with their coefficients. A FB $\F$ is \emph{dynamic}, can be shared for any number of $K \times K$ filters, and is optimized jointly with its FB coefficients $\l$. By fitting an interspace to sparse filters during training, we overcome the lack of prior knowledge for a basis that is well suited to describe filters with few non-zero coefficients. If a filter is pruned to a single FB coefficient, $h = \lambda_n \cdot \g{n}$, it is not restricted since $\g{n}$ can change. Thus, pruning interspace coefficients of dynamic FBs keeps the CNN flexible and is called \emph{interspace pruning} (IP). A $1$-sparse filter $h = h_{i_n, j_n} \cdot \en$ directly predefines $h$ to stay on the fixed subspace $\spa \{\en\}$. Pruning spatial coefficients \wrt the standard basis $\B$ is called \emph{standard pruning} (SP). 

During training sparse CNNs, the problem of vanishing gradients due to spatial sparsity often occurs \cite{tanaka_2020,wang_2020,wimmer_2020}. In contrast, IP pruned networks are able to learn spatially dense FBs during training, even when using sparse interspace coefficients, see \cref{fig:pruning}. Therefore, IP leads to an improved information flow and better trainable models. 

Although IP yields dense spatial representations, the linearity of convolutions can be used to reduce the number of computations for CNNs with sparse interspace coefficients. Compared to SP, IP only increases the number of required computations by a small, constant count. However, as IP provides superior sparse models, IP generates CNNs with faster inference speed than SP while matching the dense performance. Further, the dynamic achieved by interspace representations is cheap in terms of memory. A FB $\F$ has $K^4$ parameters as it contains $K^2$ filters of size $K \times K$. A single FB can be shared for all $K \times K$ filters in a CNN. Also, more than one FB can be used with just a small increase in memory requirements. For cost reasons, we do not use more FBs than the number of layers in a CNN in our experiments, resulting in \emph{all} FBs creating an overhead of at most $0.01\%$ of the dense network's parameters. Despite adding only few additional costs compared to using spatial weights, interspace representations significantly improve results for sparse \emph{and} dense training.

\paragraph{Our core contributions are:}
\begin{itemize}[noitemsep,topsep=0pt]
	\item Representing and training convolutional filters in the interspace, a linear space spanned by a trainable FB. The FB is optimized jointly with the FB coefficients. 
	\item Formulating the concept of pruning for filters with interspace representation as general method to improve performance of CNNs with sparse coefficients.
	\item Theoretical proof of IP's improvements in \cref{thm:spd}.
	\item Experiments showing that IP exceeds SP for equal runtime and memory costs on SOTA sparse training methods and pruning methods which are applied during training or on pre-trained models. We demonstrate that IP's superiority is achieved by improved trainability, {and} at lower sparsity also due to better generalization. 
\end{itemize}

\section{Broader Impact} Pruning can lower costs for training, storing and evaluating DNNs. We are not aware of any negative outcome directly induced by this work. Nevertheless, as tool to improve pruning, and therefore to reduce costs for CNNs, IP could be used for any CNN based application with negative ethical or societal impact. As authors, we distance ourselves from such applications and the use of our method therein.  

As we show in the paper, IP improves unstructured pruning in general and is not restricted to a special scenario. We see IP as a tool which is applied in combination with SOTA SP techniques to lower costs further. Consequently, our work is to the advantage of everyone using pruning and, by the improved generalization ability obtained by training with interspace representations, deep learning in general.

\section{Related work}\label{sec:related_work}
Related work covers \emph{general pruning} and \emph{pruning before training and DST}. Training a sparse model allows to learn non-zero FB coefficients and FBs jointly from scratch. Such methods naturally benefit most from interspace representations and our experiments thus place a strong focus on them.
\paragraph{General pruning.}
Pruning is divided in \emph{structured} and \emph{unstructured} pruning. Structured pruning removes coarse structures of the network, like channels or neurons \cite{anwar_2017,huang_2018,zhuang_2020,wang_2021b,tang_2021,li_2021}. This yields lean architectures and thus reduces computation time. A more fine-grained approach is unstructured pruning where single, spatial weights are zeroed \cite{karnin_1990,lecun_1990,guo_2016,gale_2019,frankle_2018,lee_2018,mocanu_2018}. Unstructured pruning leads to better performance than structured pruning  \cite{li_2016,mao_2017} but requires soft- and hardware that supports sparse tensor computations to actually reduce runtime  \cite{han_2016,parashar_2017}. Also, storing sparse parameters in formats such as the \emph{compressed sparse row format} \cite{tinney_1967} creates additional overhead. This can lead to non-linear dependencies between the sparsity and actual memory/runtime costs, see also Appendix \cref{sec:memory,subsec:real_runtime}. 

Pruning can be applied at any time in training. The historically first approaches \cite{hanson_1988,janowsky_1989,mozer_1989,karnin_1990,lecun_1990} use trained networks and many prune and fine-tune cycles. Criteria are often based on expensive computations of the Hessian \wrt the loss function. Likewise, magnitudes of trained coefficients can be used as iterative pruning criterion \cite{han_2015,li_2016,gale_2019}. By adding sparsity forcing regularizations to the loss, pruning can be integrated dynamically into training \cite{cp_2018,louizos_2018,yang_2020}. 

Closest to our work are pruning coefficients in the frequency \cite{liu2_2018} and the Winograd domain \cite{liu3_2018}. Contrarily, we do not bind representations to a fixed basis but let the network learn its FBs self-reliantly. Moreover, IP is not a pruning method by itself, but is added on top of existing ones to boost them. Also to mention is \cite{li_2019}, a {low rank} approximation of CNNs. A dense, pre-trained network is approximated by learning undercomplete dictionaries for 3D filters. We, on the contrary, represent $2$D filters $h \in \R^{K \times K}$, prune the network instead of using low rank approximations and learn FBs jointly with the coefficients in one training. 

\paragraph{Pruning before training and dynamic sparse training.}
In \cite{frankle_2018}, an iterative procedure is proposed which consists of training un-pruned weights to convergence, applying magnitude pruning with a small pruning rate to the trained weights and resetting the non-zero weights to their initial value. Finally, this leads to sparse, randomly initialized networks which are well trainable -- so called lottery tickets. For SOTA CNNs, resetting un-pruned weights not to the initialization but a value from an early iteration improves performance significantly \cite{frankle_2020a,renda_2020}. By applying other criteria, like \emph{information flow} in the sparse network \cite{tanaka_2020,wang_2020,patil_2021,wimmer_2021} or influence of non-zero weights on changing the loss \cite{lee_2018,verdenius_2020,jorge_2020,wimmer_2021}, pruning can be successfully applied at initialization without pre-training the network. GraSP \cite{wang_2020}, SNIP \cite{lee_2018} and SynFlow \cite{tanaka_2020} are SOTA for PaI \cite{frankle_2021}. Dynamic sparse training  \cite{mocanu_2018,dettmers_2019,evci_2020,liu_2021b} adjusts pruning masks during training to ensure sparse networks while adapting the architecture to different conditions. SET \cite{mocanu_2018} frequently prunes the network based on magnitudes and activates as many un-trained parameters randomly. RigL \cite{evci_2020} improves this by recovering those weights with the biggest gradient magnitude.

\section{Filter bases and interspace pruning}\label{sec:fbs}
Inspired by \emph{sparse dictionary learning} (SDL) (\cref{subsec:spd}) we introduce interspace representations of convolutional filters and propose computations of resulting FB convolutions in \cref{subsec:fb_representation}. Further, \cref{subsec:fb_init_description} discusses FB sharing and the initialization of FBs and their coefficients. Finally, interspace pruning is formally defined in \cref{subsec:fb_pruning_description}.  

\subsection{Inspiration from sparse dictionary learning}\label{subsec:spd}
Sparse dictionary learning \cite{engan_1999,aharon_2006,mairal_2010} optimizes a dictionary $\Fb \in \R^{m \times m}$ jointly with coefficients $R \in \R^{m \times n}$ to approximate a target $U \in \R^{m \times n}$ by using only $s$ non-zero coefficients. Setting the \emph{pruning mask} $\supp R := \{(i,j) : R_{i,j} \neq 0\}$, this defines a non-convex optimization problem
\begin{equation}
\inf_{\Fb , R}\Vert U - \Fb \cdot R \Vert_F \;\; \text{s.t.} \; \Vert R \Vert_0 := \# \supp R \leq s \label{eq:spd_fb}\;.
\end{equation}
Usually, SDL allows $\Fb \in \R^{m \times M}$ with arbitrary $M$. Since FBs are bases, we restrict $\Fb$ to be quadratic. In our context, $U$ corresponds to all flattened filters of a convolutional layer, the dictionary $\Fb$ to the layer's flattened FB $\F$ and $R$ to the FB coefficients. For a layer $h \in \R^{\co \times \ci \times K \times K}$ with an associated FB $\F = \{g^{(1)}, \ldots, g^{(K^2)}\} \subset \R^{K \times K}$, we have $m=K^2$ and $n = \co \cdot \ci$. Standard magnitude pruning is a special case of SDL where $\Fb$ is fixed to form the standard basis $\Fb = \id_{\R^m}$. Accordingly,
\begin{equation}
\min_{\bar{R} } \Vert U - {\bar{R}}  \Vert_F \;\; \text{s.t.} \; \Vert \bar{R} \Vert_0 \leq s \label{eq:spd_standard}
\end{equation}
is minimized for magnitude pruning. Since we train sparse, randomly initialized CNNs, our overall goal is not to \emph{mimic} a given dense CNN, but to train the sparse network to \emph{generalize} well. We consequently use \cref{eq:spd_fb,eq:spd_standard} only to find a decent subset of coefficients to be {pruned}. In contrast to SDL, deep learning methods are used to further optimize the un-pruned coefficients and additionally the FBs in the case of IP. 
In our experimental evaluation, we also test other methods than magnitude pruning, \ie \cref{eq:spd_fb,eq:spd_standard}. Still, \cref{eq:spd_fb,eq:spd_standard} measure the ability of a sparse layer to function as well as a dense layer and thus are good indicators for the general performance of IP and SP, respectively. 

Most SDL algorithms \cite{engan_1999, aharon_2006,mairal_2010} optimize $\Fb$ and $R$ alternatingly. Whereas, SP-PaI fixes the basis as $\id_{\R^m}$ and the pruning mask $\supp \bar{R}$ too. This simplifies the task, but reduces the solution space. IP overcomes the small, fixed solution space by adapting the basis during training. For IP-PaI, the pruning mask $\supp R$ is determined heuristically and also fixed which still leads to sub-optimal architectures. As shown in this work, using expensive pre-training to find a better pruning mask via LTs or adapting $\supp R$ during training via DST further improves IP's performance. 

\Cref{thm:spd} shows that a dynamic $\Fb$ leads to better approximations than using the standard basis. Consequently, the FBs' adaptivity improves performance after pruning and \cref{thm:spd} is a theoretical motivation for IP. 

Assume a convolutional layer with $\co$ output and $\ci$ input channels, kernel size $K \times K$ and $s = (1 - p) \cdot \co \cdot \ci \cdot K^2$ un-pruned coefficients. For $m = K^2 \geq 9$, the $\delta$ in \cref{eq:delta} is numerically equal to zero if $n = \co \cdot \ci \geq 100$ and $0 < s < \co \cdot \ci \cdot K^2$. Thus, for each non-trivial sparsity, the adaptivity of the FB improves results. This even holds if the pruning mask for \cref{eq:spd_fb} is fixed to be the one of the minimizer of \cref{eq:spd_standard}, \ie starting with an arbitrary pruned network and adding an adaptive FB always improves results. The proof of \cref{thm:spd} is shown in Appendix \cref{sec:proofs}. It uses the fact that \cref{eq:spd_fb} is smaller or equal to \cref{eq:spd_standard}. Equality is only possible if \cref{eq:spd_standard} has a solution such that each $K \times K$ filter is either fully pruned or dense. This is almost impossible for big layers and a non trivial sparsity. If $\supp R$ is further not fixed for \cref{eq:spd_fb}, $(\Fb, R)$ can be chosen such that \cref{eq:spd_fb} is always strictly smaller than \cref{eq:spd_standard}. 

\begin{thm}\label{thm:spd}
	Let $0 < s < m \cdot n$, $m >1$ and $U_{i,j} \sim \mathcal{N}(0,1)$ i.i.d. Let $\varepsilon_{(1)}$ be the infimum of \cref{eq:spd_fb}, and $\varepsilon_{(2)}$ the minimum of \cref{eq:spd_standard} solved by $\bar{R}^\ast$. Then, $\varepsilon_{(1)} < \varepsilon_{(2)}$ with $\mathbb{P} = 1$. If $\supp R$ is fixed for \cref{eq:spd_fb} to be $\supp \bar{R}^\ast$, $\varepsilon_{(1)} \leq \varepsilon_{(2)}$ is true and strict inequality holds with $\mathbb{P} \geq 1 - \delta$, where
	\begin{equation}\label{eq:delta}
	\delta = \begin{cases}
	\nicefrac{\binom{n}{\frac{s}{m}}}{\binom{m\cdot n}{s}} \, & \text{if} \, s \equiv 0 (\modulo m)\\
	0 \, & \text{else}
	\end{cases}\;.
	\end{equation}
	
\end{thm}

\Cref{fig:intro_FBs}\subref{fig:c10_rnd} compares SP and IP for random PaI for a VGG$16$ \cite{simonyan_2014} trained on CIFAR-$10$ \cite{krizhevsky_2012}. IP improves results tremendously compared to SP. This experimentally shows that sparse training performs better when coefficients of adaptive FBs are pruned than if spatial weights are pruned. This holds even though fixed pruning masks are used.

\subsection{Interspace representation and convolutions}\label{subsec:fb_representation}
\begin{figure}[tb!]
	\centering
	\begin{subfigure}[b]{0.225\textwidth}
		\centering
		\includegraphics[width=\textwidth]{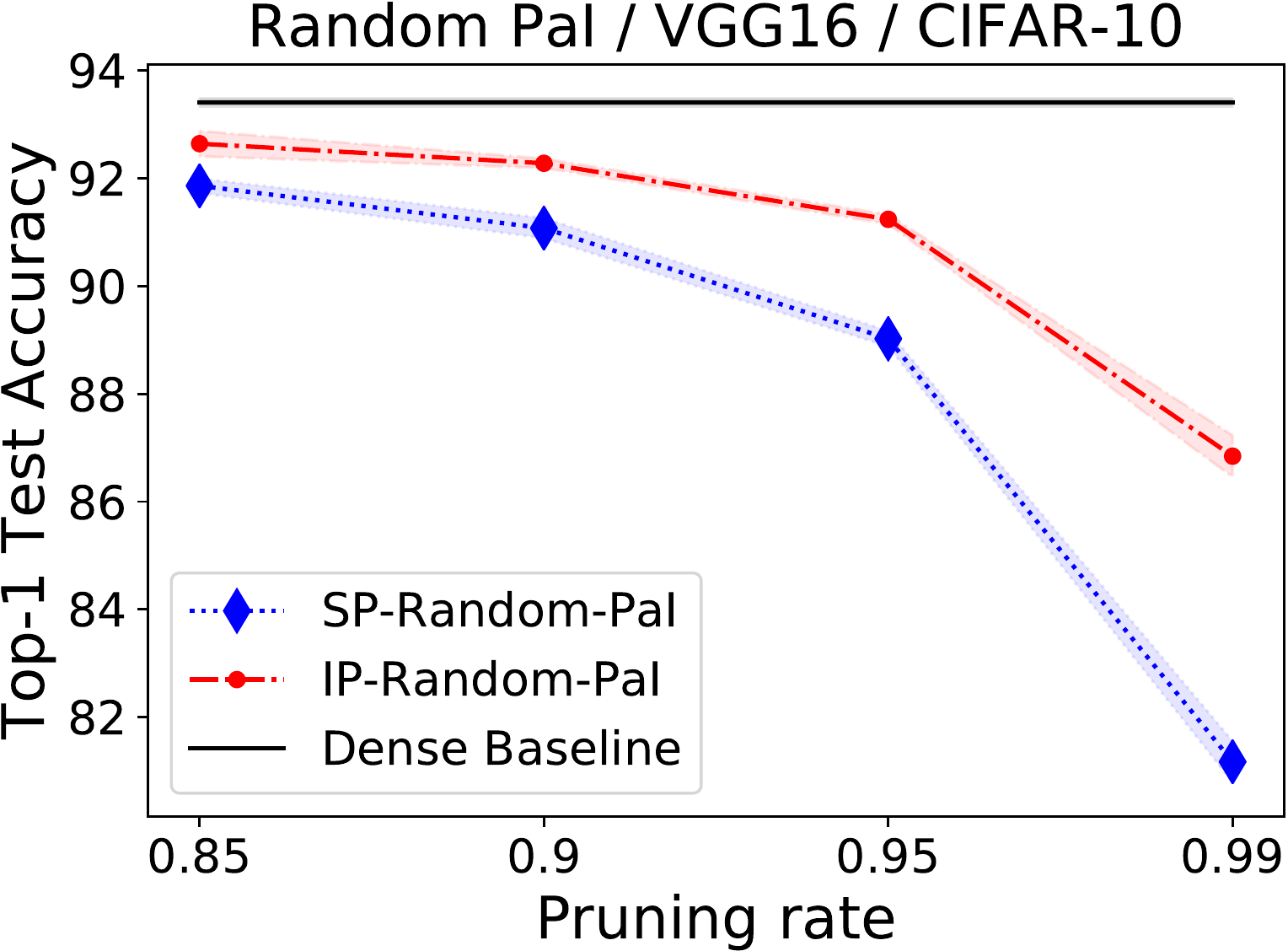}
		\caption{}			
		\label{fig:c10_rnd}
	\end{subfigure}
	\begin{subfigure}[b]{0.225\textwidth}
		\centering
		\includegraphics[width=\textwidth]{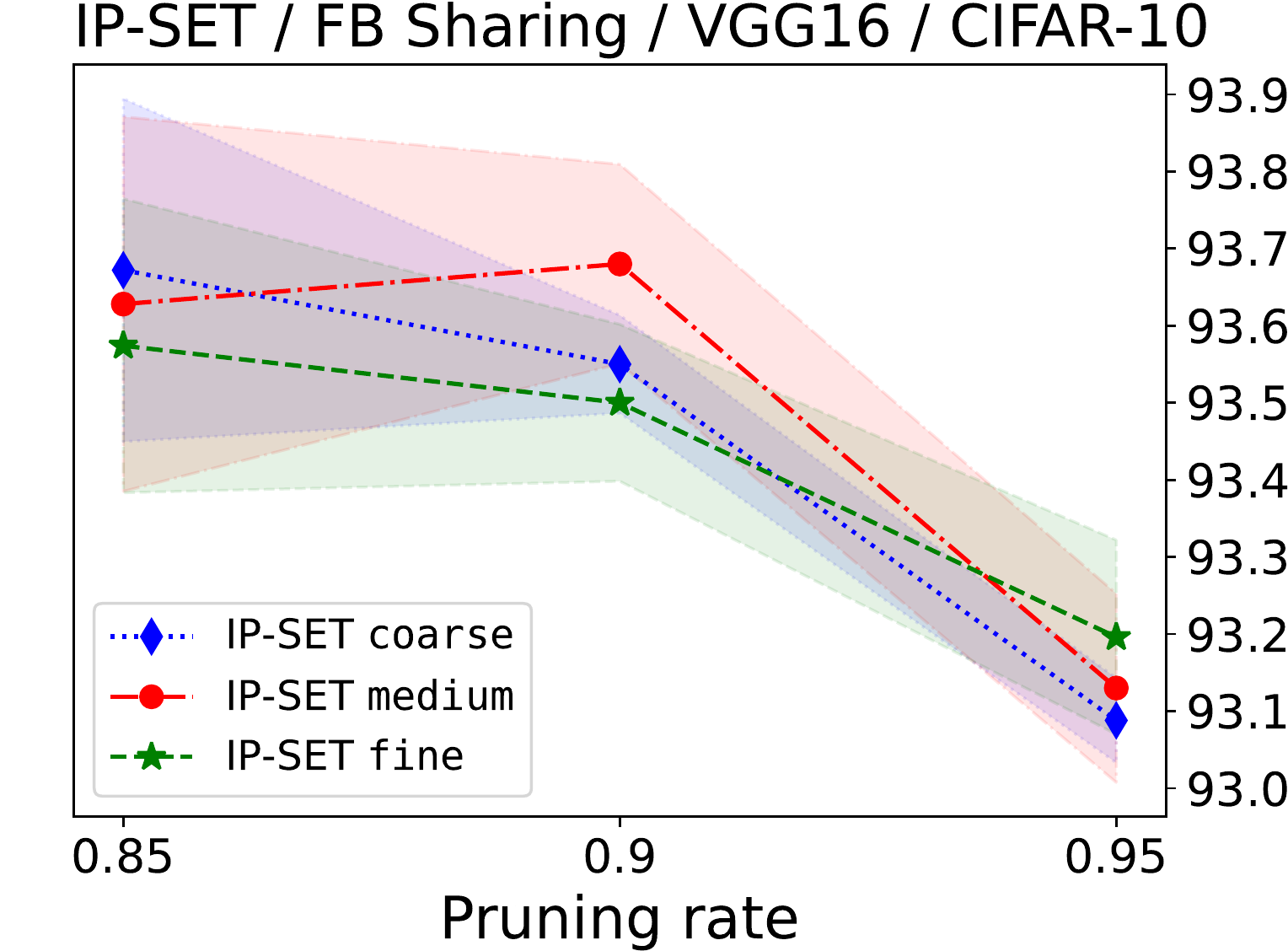}
		\caption{}
		\label{fig:c10_set_align}
	\end{subfigure}
	\caption{VGG$16$ on CIFAR-$10$: \subref{fig:c10_rnd} Random SP-PaI and random IP-PaI. \subref{fig:c10_set_align} \texttt{Coarse}, \texttt{medium} and \texttt{fine} FB sharing for IP-SET.}
	\label{fig:intro_FBs}
\end{figure}
For a convolutional layer, let $\co$ denote its number of output channels, $\ci$ its number of input channels and $K \times K$ its kernel size. To simplify formulas, we restrict the formulation to 2D convolutions with quadratic kernel, no padding, $1 \times 1$ stride and dilation. Generalizing the FB formulations to arbitrary convolutions is straightforward. 
A 2D convolution $h$ describing this layer consists of $\co \cdot \ci$ $K \times K$ filters $\hab \in \R^{K \times K}$, \ie $h = (\hab)_{\a, \b} \in \R^{\co \times \ci \times K \times K}$. Inspired by the discussion in \cref{subsec:spd}, we now represent all $\hab$ in the interspace spanned by the layer's FB $\F = \{\g{1}, \ldots, \g{K^2}\}\subset \R^{K \times K}$. The FB coefficients $\l = (\labn)_{\a,\b,n} \in \R^{\co \times \ci \times K^2}$ define the interspace representation of $\hab$, given by 
\begin{equation}
\hab = \sum_{n=1}^{K^2} \labn \cdot \g{n} \label{eq:fb_representation}\;.
\end{equation}
This is a basis transformation of the spatial representation 
\begin{equation}
\hab = \sum_{n=1}^{K^2} \habin \cdot \en \;, \; e^{(n)}_{i,j} = \delta_{i,i_n}\cdot \delta_{j,j_n} \label{eq:spatial_representation} \;.
\end{equation}
Normally, $\hab$ is defined in spatial representation. Thus, spatial coefficients are stored in $\hab \in \R^{K \times K}$. Whereas, FB coefficients are specified by vectors $ \lab \in \R^{K^2}$. By linearity, a 2D FB convolution $(Y^{(\a)})_\a = Y = h \star X$ with input feature map $X = (X^{(\b)})_\b \in \R^{\ci \times h \times w}$ can be computed for each output channel $\a \in \{1, \ldots , \co\}$ as
\begin{equation}
Y^{(\a)} =  \sum_{\b=1}^{\ci} \hab \star X^{(\b)}  \stackrel{\eqref{eq:fb_representation}}{=} \sum_{\b=1}^{\ci} \sum_{n=1}^{K^2} \labn  (\g{n} \star X^{(\b)} )\;. \label{eq:fb_conv2d}
\end{equation}
Gradients of the loss $\loss$ are needed to train the FB coefficients $\l$ and the FB $\F$. Backpropagation formulas for them are derived in Appendix \cref{subsec:comp_cost_backprop}. It holds for all $n,\a,\b$
\begin{equation}
\frac{\partial \loss}{\partial \labn} = \left \langle \gn , \frac{\partial \loss}{\partial \hab} \right \rangle ,  \frac{\partial \loss}{\partial \gn} = \sum_{\a, \b} \labn  \frac{\partial \loss}{\hab}\,.
\label{eq:backprop}
\end{equation}

\subsection{Filter basis sharing and initialization}\label{subsec:fb_init_description}
For kernel size $1 \times 1$, the FB formulation is, up to a rescaling, equivalent to the spatial representation. Thus, we assume a CNN with $L_{c}$ convolutional layers with $K > 1$ to be given and do not apply the FB formulation to $1 \times 1$ convolutions. In this work, we test three versions of FB sharing. Our FB sharing schemes differ in their \emph{granularity}. The \texttt{coarse} scheme shares one global FB $\F$ for all layers $l=1, \ldots, L_c$. Whereas, the \texttt{fine} scheme shares a FB $\Fl$ for each layer $l$, thus it uses $L_{c}$ FBs. In between lies the \texttt{medium} scheme with $5$ FBs in total. For ResNets \cite{he_2016}, one FB is shared for each of the $5$ convolutional blocks. For VGG$16$ \cite{simonyan_2014}, convolutional layers $\{1,2\}$, $\{3,4\}$, $\{5,6,7\}$, $\{8,9,10\}$ and $\{11,12,13\}$ share one FB each. The number of FBs increases from \texttt{fine} to \texttt{coarse}. The total number of FBs in the network, $J$, satisfies $J \leq L_c$. Consequently, the number of parameters in \emph{all} FBs in the network is bounded from above by $L_c \cdot K^4$. Note for the CNNs used in this work, $L_c \cdot K^4$ is at most $0.01 \%$ of all parameters in the model. Thus, the additional parameter costs for IP with our proposed sharing schemes are neglectable.

The dimension of the space spanned by each layer does not change for different FB sharing schemes and is equal to using spatial representations. However, \texttt{coarse} sharing correlates all layers in the network by using and updating the same interspace. For \texttt{fine} sharing, each layer has its own interspace which is adapted more fine-grained. For spatial representations, the basis $\B$ is fixed, not updated and does not induce correlations between weights. We found different sharing schemes to work best for varying training/\allowbreak model/\allowbreak dataset combinations. \Cref{fig:intro_FBs}\subref{fig:c10_set_align} shows our FB sharing schemes for different pruning rates. \texttt{Coarse} sharing works best for higher numbers of trained parameters. By correlating all layers through a global FB, we assume it to have a regularizing effect on training, see also \cref{subsec:generalization}. \texttt{Fine} sharing makes the network more flexible. Thus, results are the best ones for high pruning rates where the network is not able to overfit on the training data anymore. In between, \texttt{medium} sharing reaches the best results by combining the best of both worlds. 

In this work, we use a simple initialization for FBs and FB coefficients. We initialize each FB as $\B$ and the FB coefficients with a \texttt{kaiming normal} initialization \cite{He2015}. This scheme is equivalent to the \texttt{kaiming normal} initialization for standard CNNs -- which is also used for dense baselines and SP experiments. In Appendix \cref{sec:init_fbs}, we propose further initialization schemes for the interspace.

\subsection{Interspace pruning and cost comparison}\label{subsec:fb_pruning_description}
SP is modeled by superimposing \emph{pruning masks} $\bar{\mu}^{(\a,\b)} \in \{0,1\}^{K \times K}$ over filters $\hab \in \R^{K \times K}$. This results in sparse filters $\hab \odot \bar{\mu}^{(\a,\b)}$, with the Hadamard product $\odot$.
Filters represented in the interspace have coefficients $\lab \in \R^{K^2}$ \wrt a FB $\F$. Thus, interspace pruning is defined by masking FB coefficients with pruning masks $\muab \in \{0,1\}^{K^2}$ via $\lab \odot \muab$. Combined with \cref{eq:fb_conv2d}, IP yields sparse computations of convolutions:
\begin{equation}\label{eq:sparsity_ip}
Y^{(\a)} = \sum_{\b=1}^{\ci} \sum_{n \in \supp \muab} \labn \cdot \left ( \gn \star X^{(\b)} \right ) \;.
\end{equation}
The \emph{pruning rate} $p$ for SP ($p_{SP}$) and IP ($p_{IP}$) is defined as
\begin{equation}
p_{SP} = 1 -  \frac{\Vert \Lambda \Vert_0}{D}  \;, \;
p_{IP} = 1 - \frac{\Vert \Lambda \Vert_0 + \sum_{j=1}^{J} \Vert \F^{(j)} \Vert_0}{D}  \label{eq:fb_pruning_rate}\;.
\end{equation}
For SP, $\Lambda \in \R^D$ denotes the network's parameters, whereas $\Lambda \in \R^D$ contains all parameters except the FBs themselves in the IP setting. Thus, $\L$ has exactly the same number of elements for IP and SP. The pruning rates \cref{eq:fb_pruning_rate} are the fractions of parameters being equal to zero. To have a fair comparison between IP and SP, we normalize the number of non-zero parameters with the total count of coefficients in the standard dense network, \ie the dense network without FBs. The number of bias and batch normalization parameters is tiny compared to convolutional and fully connected layers. Also, all parameters of FBs together are at most $0.01 \%$ of $D$ in our experiments. Consequently, we only prune weights of fully connected layers as well as spatial- and FB coefficients of convolutional layers. FBs, bias and batch normalization parameters are all trained.

\paragraph{Computational cost comparison.}
As discussed, parameter costs for IP with our FB sharing schemes are only negligibly bigger than for SP. By the linearity of convolutions, the sparsity of filters in the interspace can be used to reduce computational costs, see \cref{eq:sparsity_ip}. In Appendix \cref{sec:comp_costs}, computational costs are calculated and compared for IP and SP. Costs are measured by the number of theoretically required \emph{floating point operations} (\mults{}) for a convolutional layer and are independent of the used FB sharing scheme. IP's overhead is composed of additional costs in the forward and backward pass. For inference, only the additional cost of the forward pass counts. Both, SP and IP, need specialized soft- and hardware that supports sparse computations to actually reduce runtime.

Assume a layer with kernel size $K \times K$, $\ci$ input and $\co$ output channels. In the forward pass, SP has $1 - p$ times the \mults{} cost of the dense layer. Due to \cref{line:overhead0}-\ref{line:overhead} in \cref{alg:fb_conv2d}, IP has a constant overhead $\nicefrac{K^2}{\co}$. In total, IP has $1-p + \nicefrac{K^2}{\co}$ times the \mults{} cost of the dense layer. 

In the backward pass, the number of \mults{} for IP is in $\mathcal{O} \left( cost \left( \frac{\partial \loss}{\partial h} \right) \right)$, \ie comparable to the cost of computing the {dense} gradient of layer $h$ in spatial representation. 

As discussed, IP needs more computations for inference than SP for equal sparsity. However, since IP finds superior sparse models, IP {actually} achieves \emph{a higher speed up} in real time measurements than SP while reaching similar or even better performance, as will be shown \cref{fig:ablation}\subref{fig:c10_lth_mults}. 
\begin{algorithm}[tb]
	\caption{FB 2D Convolution with IP}
	\label{alg:fb_conv2d}
	\small
	\begin{algorithmic}[1]
		\algrenewcommand\algorithmicprocedure{\textbf{instance variables}}
		\Procedure{}{}\Comment{of \texttt{IP\_FB\_2DConv}}
		\State \texttt{filter\_basis}: $\{g^{(1)}, \ldots , g^{(K^2)}\} \subset \R^{K \times K}$
		\State \texttt{fb\_coefficients}: $(\labn)_{\a,\b,n} \in \R^{\co \times \ci \times K^2}$
		\State \texttt{pruning\_mask}: $(\mu_n^{(\a, \b)})_{\a,\b,n} \in \{0,1\}^{\co \times \ci \times K^2}$
		\State \texttt{conv\_args} \Comment{e.g. \texttt{stride}, \texttt{padding}, \texttt{groups}, \ldots}
		\EndProcedure
		\algrenewcommand\algorithmicprocedure{\textbf{def}}
		\Procedure{forward\_pass}{$X$} \Comment{input $X \in \R^{\ci \times h \times w}$}
		\ForAll{$\b \in \{1, \ldots, \ci \}, n \in \{1, \ldots, K^2\}$} \label{line:overhead0}
		\State $Z_n^{(\b)} = \texttt{Conv2D}(g^{(n)}, X^{(\b)}, \texttt{conv\_args})$ \label{line:overhead}
		\EndFor
		\ForAll{$\a \in \{1, \ldots, \co\}$}
		\State $Y^{(\a)} = \sum_{\{(\b, n) : \mu_n^{(\a, \b)} = 1 \}} \labn \cdot Z_n^{(\b)}$
		\EndFor
		\State \Return $Y = (Y^{(\a)})_{\a=1}^{\co}$
		\EndProcedure
	\end{algorithmic}
\end{algorithm}

\paragraph{Pruning methods.}
\Cref{alg:fb_conv2d} describes sparse FB 2D convolutions with IP in pseudo code. Since automatic differentiation is standard in modern deep learning frameworks, backpropagation formulas for FB convolutions are computed automatically and are not included in \cref{alg:fb_conv2d}. The FB in \cref{alg:fb_conv2d} might be shared over several layers, see \cref{subsec:fb_init_description}. Our experiments in \cref{sec:experiments} compare SP and IP on various sparse training and other pruning methods, namely:

\underline{\textbf{DST}} randomly prunes the model at initialization. During training, unimportant coefficients are pruned based on their magnitude. In each layer, the same number of parameters is \emph{regrown} by activating their gradients. SET regrows coefficients randomly whereas RigL regrows those with high gradient magnitude. The pruning mask is updated each $1,500$ iterations for SET and $4,000$ for RigL. A cosine schedule is used to reduce the number of pruned/regrown coefficients.

\underline{\textbf{LT}} pre-trains the network for $t_0 = 500$ steps. Then, the network is trained to convergence. Now, $20\%$ of the non-zero coefficients are pruned based on their magnitude. The un-pruned part of the CNN is reset to its value at $t_0$. The whole procedure is applied $k$ times in total until the desired pruning rate $p = 1 - 0.8^{k}$ is reached. Ultimately, the final sparse network is trained, starting at $t_0$.

\underline{\textbf{PaI}} prunes the model at initialization without pre-training or changing the pruning mask during training. Random PaI prunes weights \iid with probability $p$. {SNIP} trains coefficients which have high influence on changing the loss $\loss$ when training starts. {GraSP} finds coefficients which improve the gradient flow at the beginning of training most. {SynFlow} keeps coefficients with high information throughput which is measured by their influence on the total \emph{path norm} of the sparse network.

\underline{\textbf{Gradual Magnitude Pruning}} (GMP) \cite{gale_2019} starts training with dense coefficients. During training, the CNN is gradually sparsified based on the coefficients' magnitudes. Pruned parameters are fixed at zero, thus never regrow.

\underline{\textbf{Fine-Tuning}} (FT) \cite{renda_2020} uses a pre-trained network. The $p \cdot D$ coefficients with smallest magnitude are pruned. The pre-trained coefficients of the sparse CNN are fine-tuned with the learning rate schedule of the dense training.

All these methods were developed for SP. Yet, in our experiments they are applied unchanged to the interspace setting. For more details see Appendix \cref{subsec:pruning_scores,sec:init_fbs}.

\section{Experiments and discussion}\label{sec:experiments}
\begin{figure}[tb!]
	\centering
	\begin{subfigure}[b]{0.225\textwidth}
		\centering
		\caption*{\textbf{VGG16 on CIFAR-10}}
		\includegraphics[width=\textwidth]{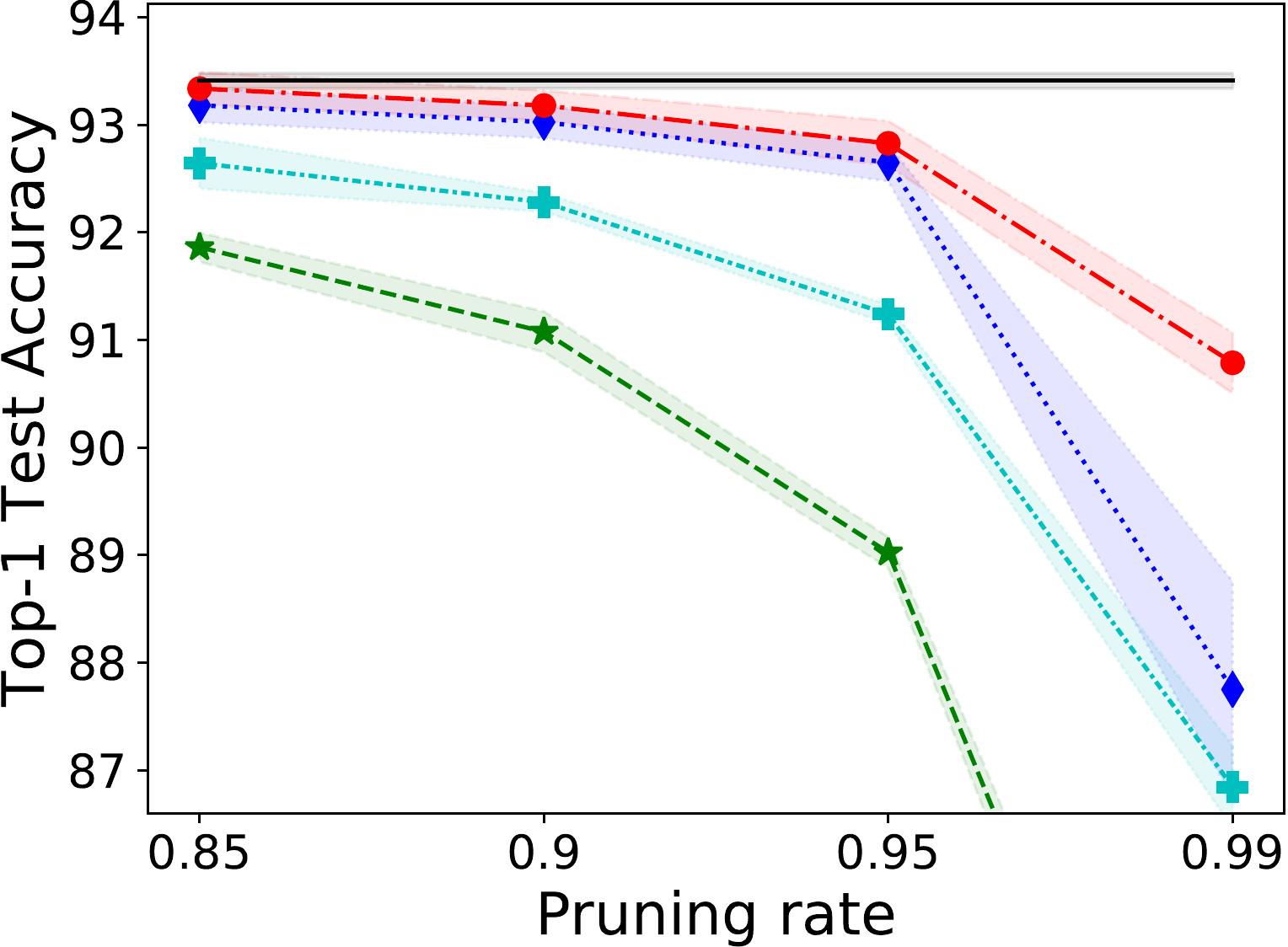}
		\caption{SNIP}
		\label{fig:c10_snip}
	\end{subfigure}
	\begin{subfigure}[b]{0.225\textwidth}
		\centering
		\caption*{\textbf{ResNet18 on ImageNet}}
		\includegraphics[width=\textwidth]{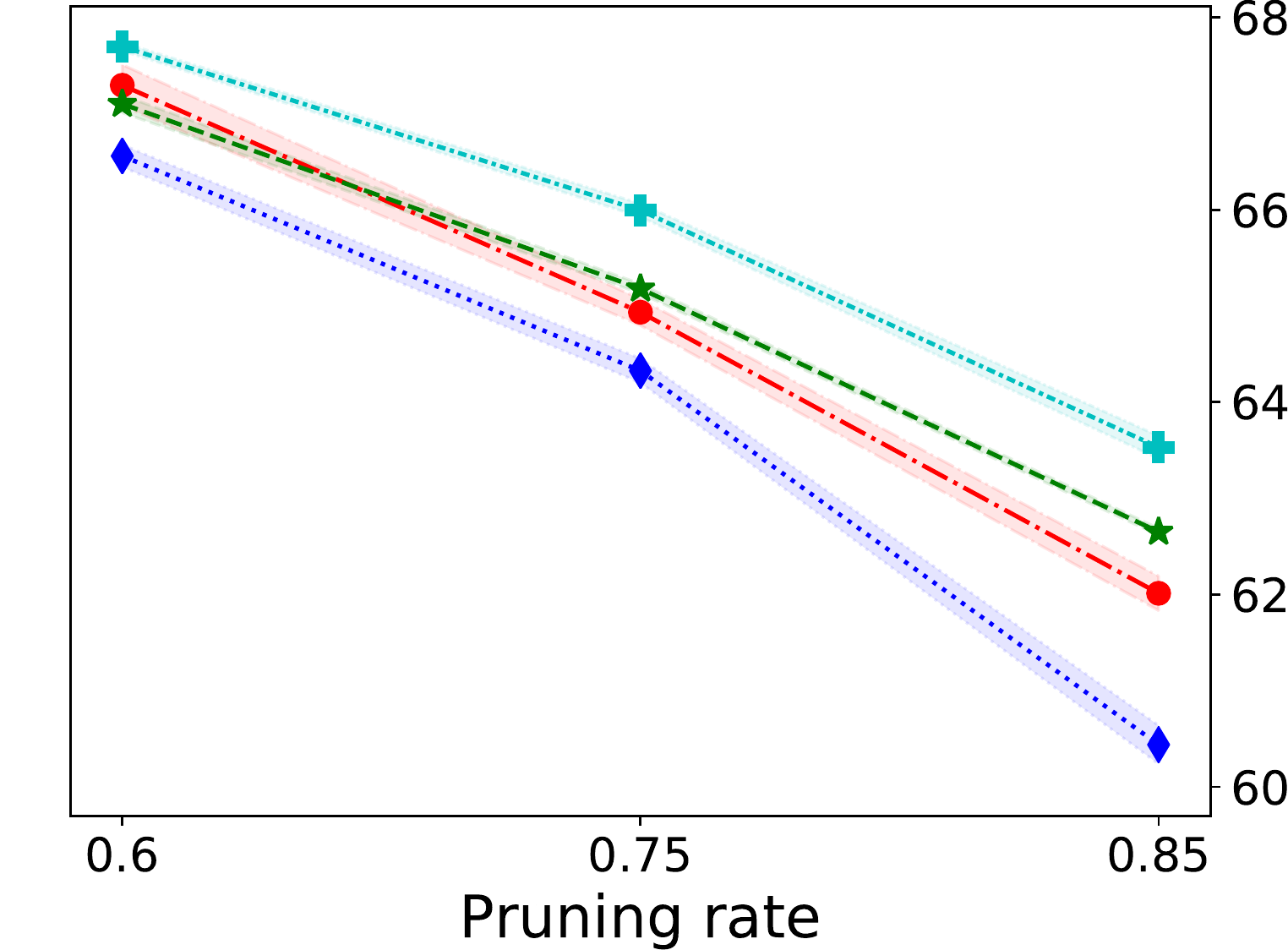}
		\caption{SNIP}	
		\label{fig:imgnet_snip}
	\end{subfigure}	
	
	\centering
	\begin{subfigure}[b]{0.225\textwidth}
		\centering
		\includegraphics[width=\textwidth]{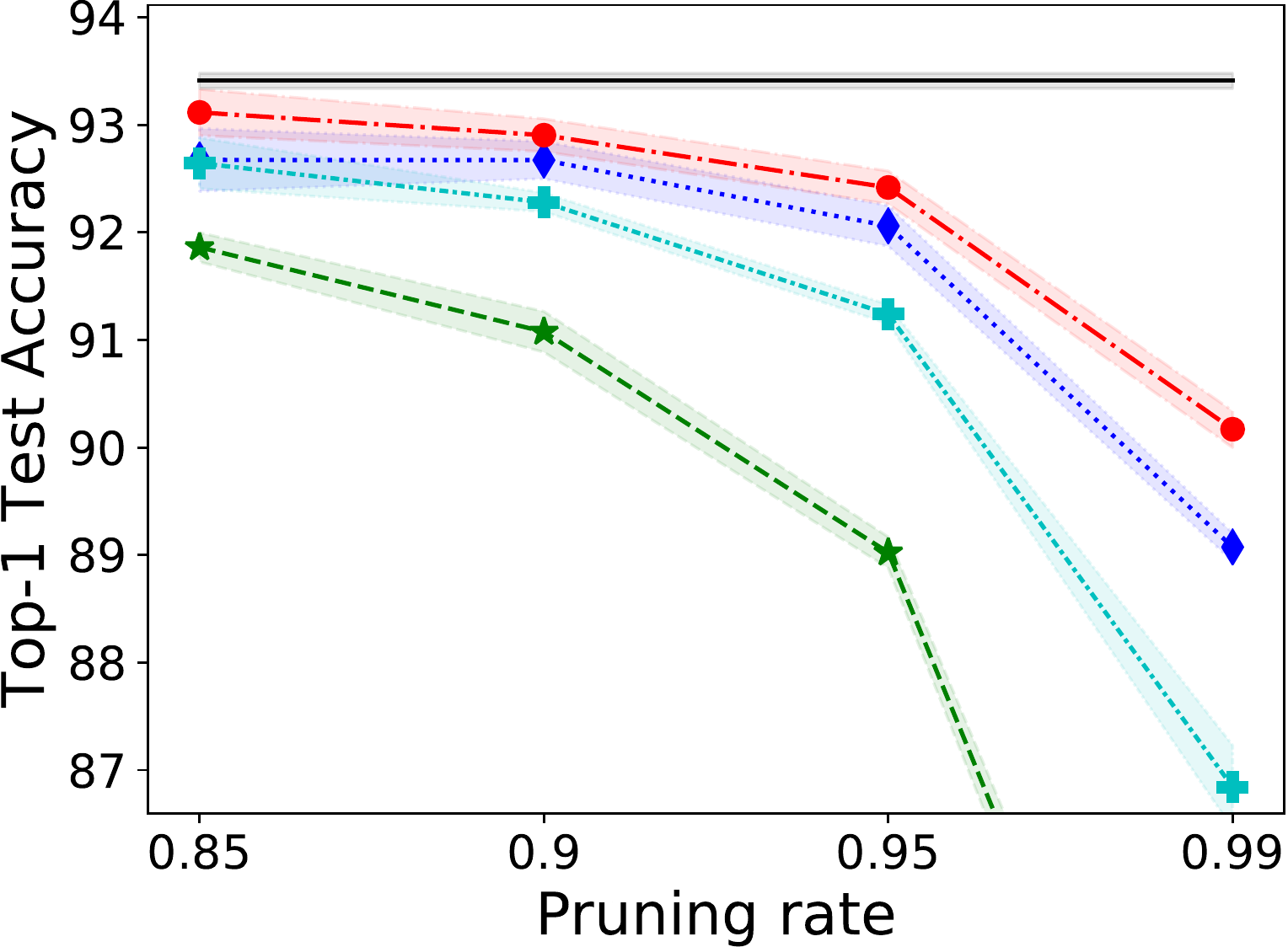}
		\caption{GraSP}
		\label{fig:c10_grasp}
	\end{subfigure}
	\begin{subfigure}[b]{0.225\textwidth}
		\centering
		\includegraphics[width=\textwidth]{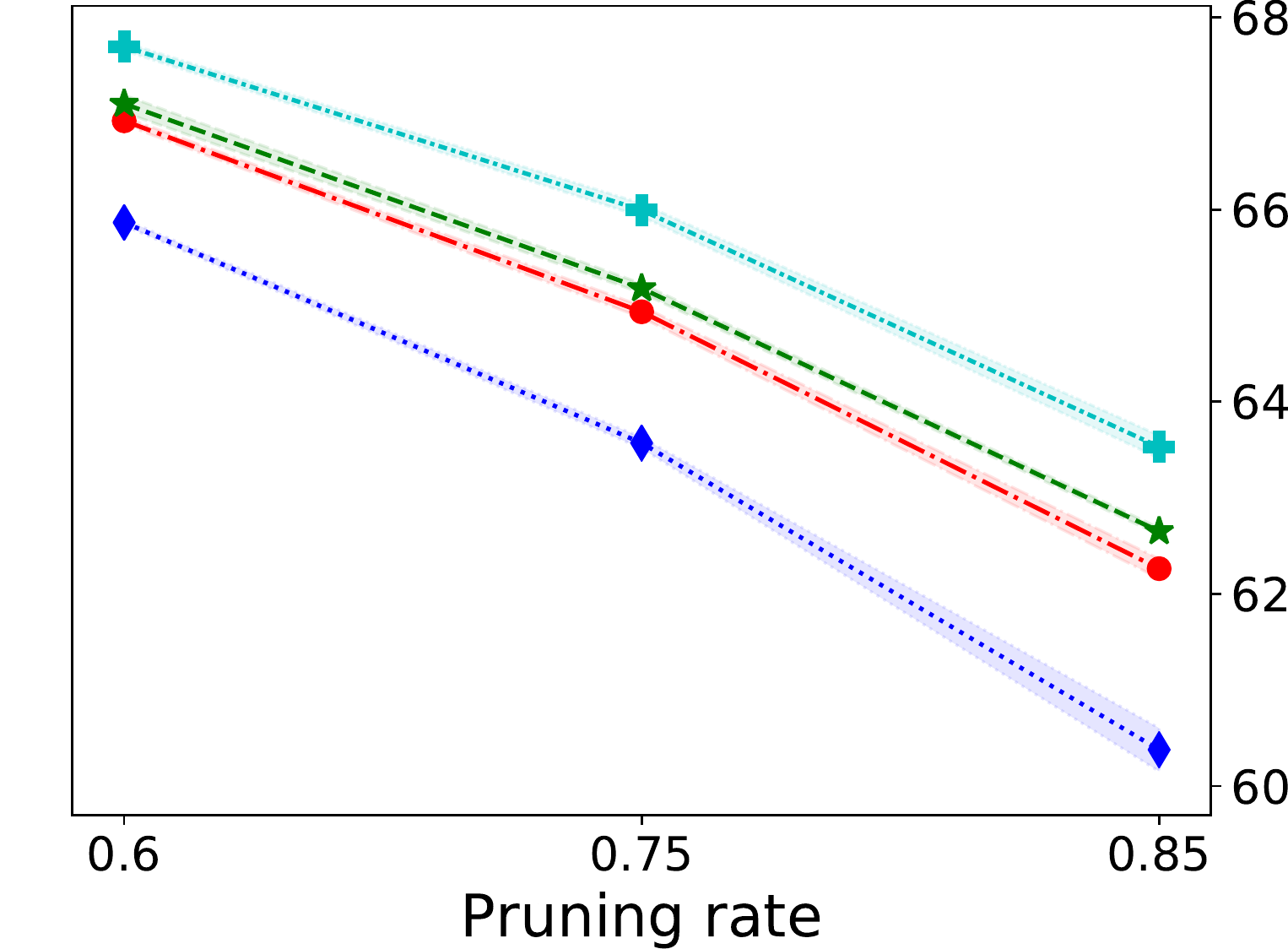}
		\caption{GraSP}	
		\label{fig:imgnet_grasp}
	\end{subfigure}

	\centering
	\begin{subfigure}[b]{0.225\textwidth}
		\centering
		\includegraphics[width=\textwidth]{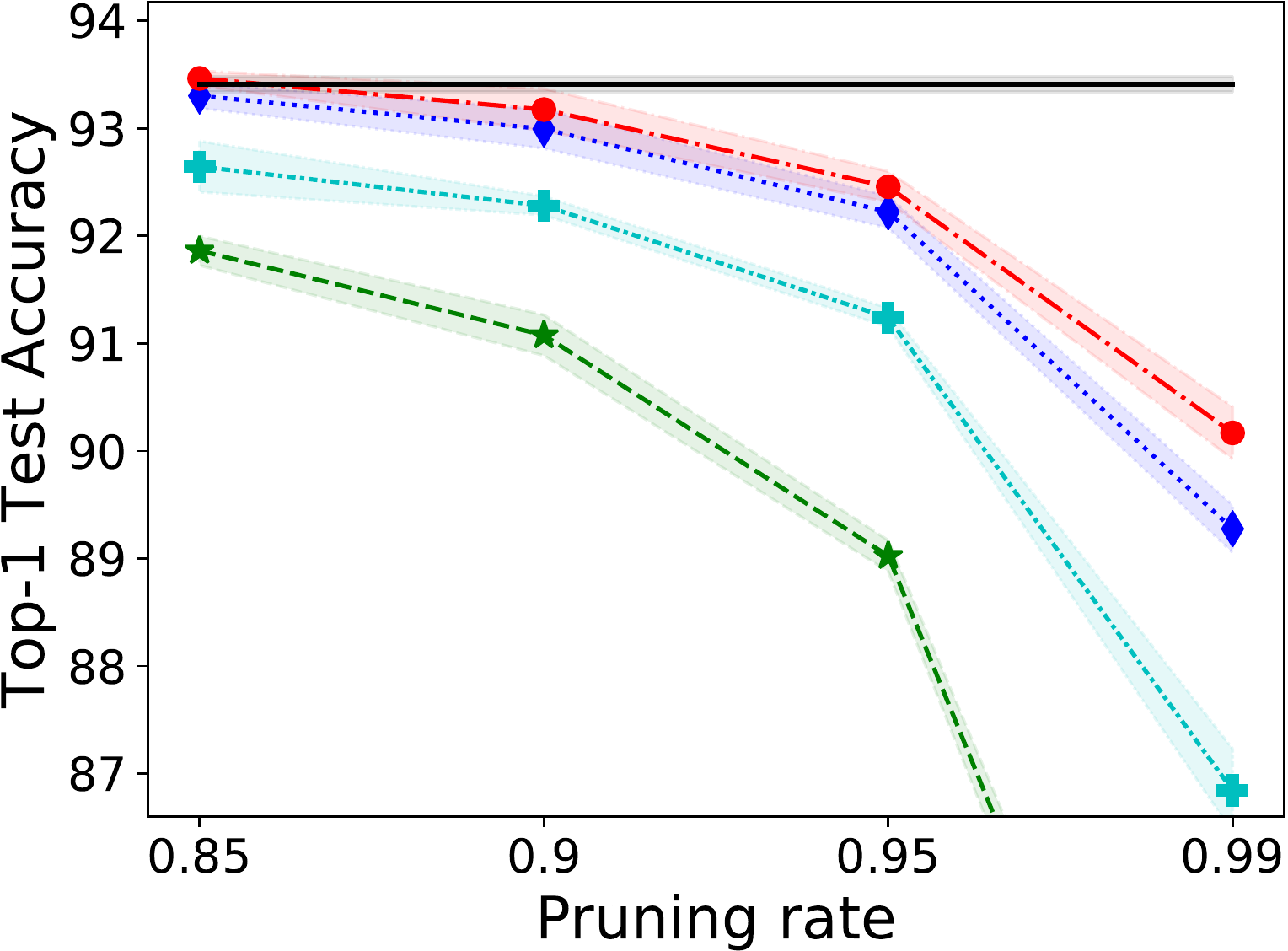}
		\caption{SynFlow}
		\label{fig:c10_synflow}
	\end{subfigure}
	\begin{subfigure}[b]{0.225\textwidth}
		\centering
		\includegraphics[width=\textwidth]{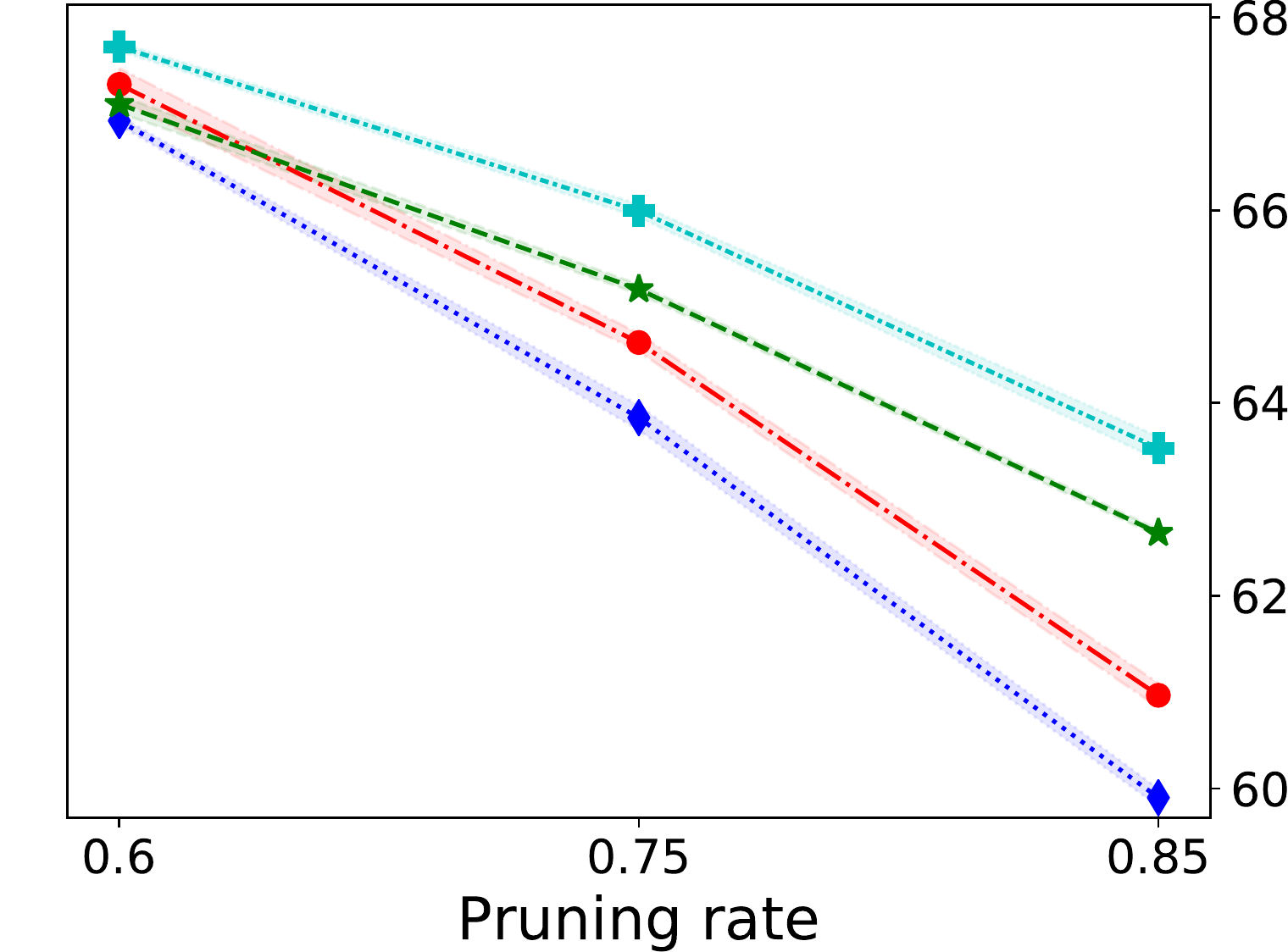}
		\caption{SynFlow}	
		\label{fig:imgnet_synflow}
	\end{subfigure}	
	\centering
	\begin{subfigure}[b]{\linewidth}
		\centering
		\includegraphics[width=\textwidth]{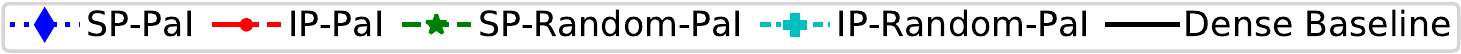} 
	\end{subfigure}
	\caption{Comparing SP and IP for PaI methods, SNIP, GraSP and SynFlow together with random PaI for CIFAR-$10$ and ImageNet.}	
	\label{fig:cifar_basic_experiment}	
\end{figure}
\Cref{subsec:exp_setup} covers the experimental setup. Next, \cref{subsec:cifar10} compares the three SOTA PaI methods \cite{lee_2018,tanaka_2020,wang_2020} for IP and SP. In \cref{subsec:dst_lth}, we discuss IP and SP for more sophisticated sparse training methods, namely LTs \cite{frankle_2020a} and the DST methods SET \cite{mocanu_2018} and RigL \cite{evci_2020}. Furthermore, we show that IP also improves SP on classical pruning methods applied during training, GMP \cite{gale_2019}, and on pre-trained models, FT \cite{renda_2020}. Improved trainability and generalization ability of IP compared to SP is shown and discussed in \cref{subsec:generalization}. 

\subsection{Experimental setup}\label{subsec:exp_setup}
We compare IP and SP for a VGG$16$ \cite{simonyan_2014} on CIFAR-$10$ \cite{krizhevsky_2012} and ResNets $18$ and $50$ \cite{he_2016} on ImageNet ILSVRC$2012$ \cite{imagenet_2012}. Models are trained with cross entropy loss. We report mean and std of five runs for CIFAR-$10$ and three for ImageNet. Weight decay is applied on coefficients but not on FBs. Coefficients of $3 \times 3$ filters and their FBs are trained jointly, whereas fixed FBs $\F = \B$ are used for $1\times 1$ filters. For ResNet$18$ we fix the FB $\F = \B$ for the $7 \times 7$ convolution whereas the $7 \times 7$ FB is trained for ResNet$50$. We use \texttt{medium} FB sharing for CIFAR-$10$ experiments, \texttt{fine} for ResNet$50$ and \texttt{coarse} sharing for all $3\times 3$ convolutions for the ResNet$18$ on ImageNet. For SP and dense baselines, standard CNNs are used. As common in the literature, we report ImageNet results on the validation set. Note, we use training schedules intended for the corresponding SP method for both, SP and IP. In particular, FBs are trained without optimized hyperparameters. Thus, they use the same learning rate as all parameters. More details on hyperparameters, evaluation and used CNN architectures are given in \cref{sec:experimental_setup,sec:architectures} in the Appendix.

\subsection{Pruning at initialization methods}\label{subsec:cifar10}
\Cref{fig:cifar_basic_experiment} compares SP and IP for PaI methods SNIP \cite{lee_2018}, GraSP \cite{wang_2020} and SynFlow \cite{tanaka_2020} together with random PaI for a VGG$16$ on CIFAR-$10$ and a ResNet$18$ on ImageNet. 

The experiments show that pruning FB coefficients instead of spatial parameters leads to significant improvements in top-$1$ test accuracy while having the same memory costs. This holds true for all PaI methods, pruning rates and for high $p$ in particular. In comparison to CIFAR-$10$, IP improves results on ImageNet even more. However, the three methods SNIP, GraSP and SynFlow are all outperformed by random PaI for ResNet$18$ on ImageNet. This demonstrates that these methods perform well for smaller datasets but show inferior results for small networks on big scale datasets like ImageNet. Still, as discussed earlier, the use of IP significantly improves all PaI methods, including random PaI. \Cref{subsec:dst_lth} shows that IP benefits from a stronger underlying pruning method to improve results further.  

Despite optimizing FBs in addition to FB coefficients, IP does not induce instability compared to SP, see \cref{fig:ablation}\subref{fig:c10_snip_gradflow} and standard deviations in \cref{fig:cifar_basic_experiment}. 
In Appendix \cref{subsec:upper_grad_bounds}, we show that the upper bounds for the gradient norms of FBs $\frac{\partial \loss}{\partial \F}$ and FB coefficients $\frac{\partial \loss}{\partial \l}$ are both determined by $\Vert \frac{\partial \loss}{\partial h} \Vert$. This boundedness of the gradients leads to stable convergence for both, $\F$ and $\l$, while the convergence behavior of $\l$ and the standard coefficients $h$ is similar, see \cref{fig:ablation}\subref{fig:c10_snip_gradflow}.

\subsection{DST, LTs and classical pruning methods}\label{subsec:dst_lth}
\begin{figure}[tb!]
	\centering
	\begin{subfigure}[b]{0.225\textwidth}
		\centering
		\includegraphics[width=\textwidth]{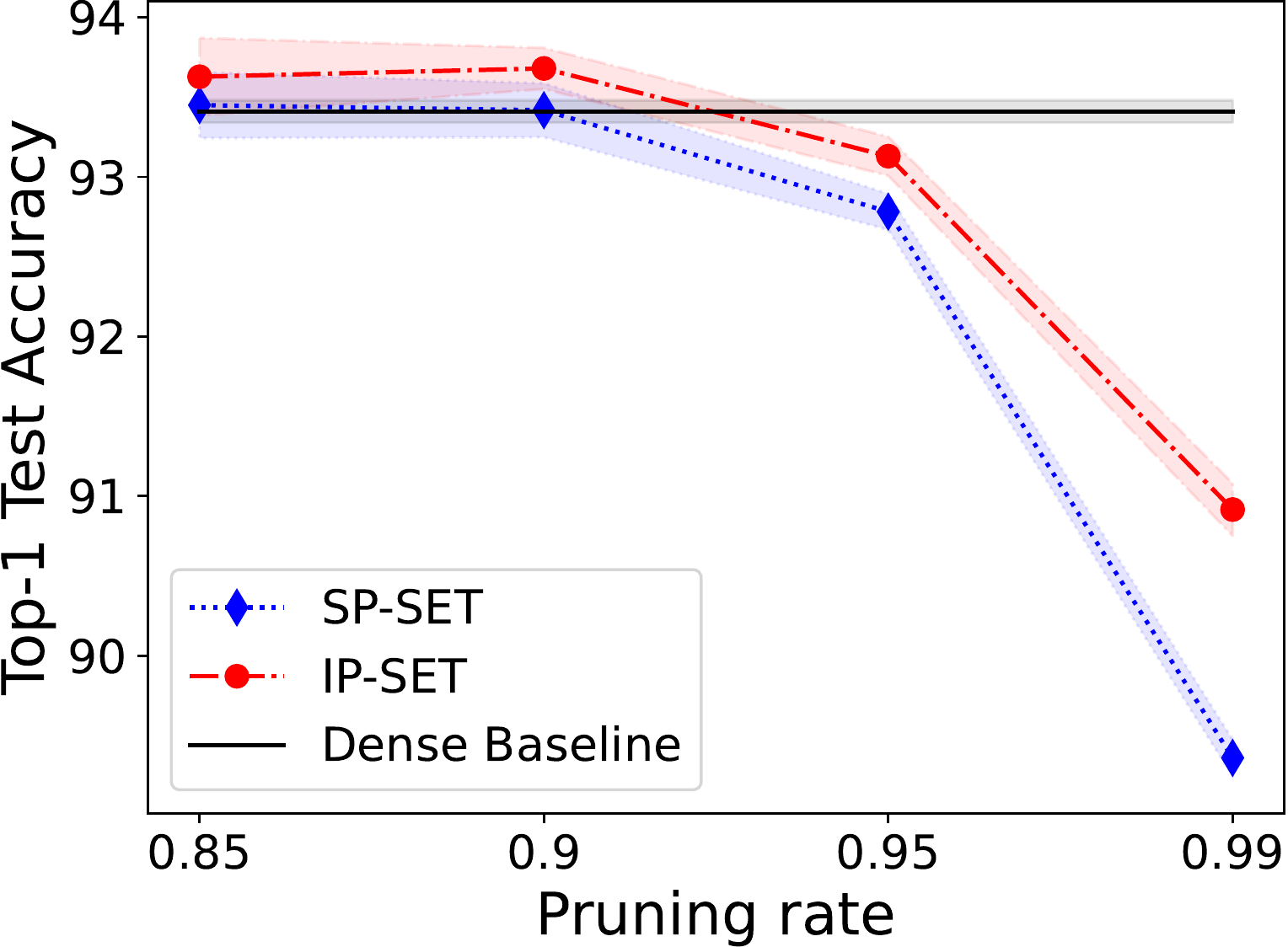}
		\caption{SET for VGG$16$ on CIFAR-$10$}
		\label{fig:c10_set}
	\end{subfigure}
	\begin{subfigure}[b]{0.225\textwidth}
		\centering
		\includegraphics[width=\textwidth]{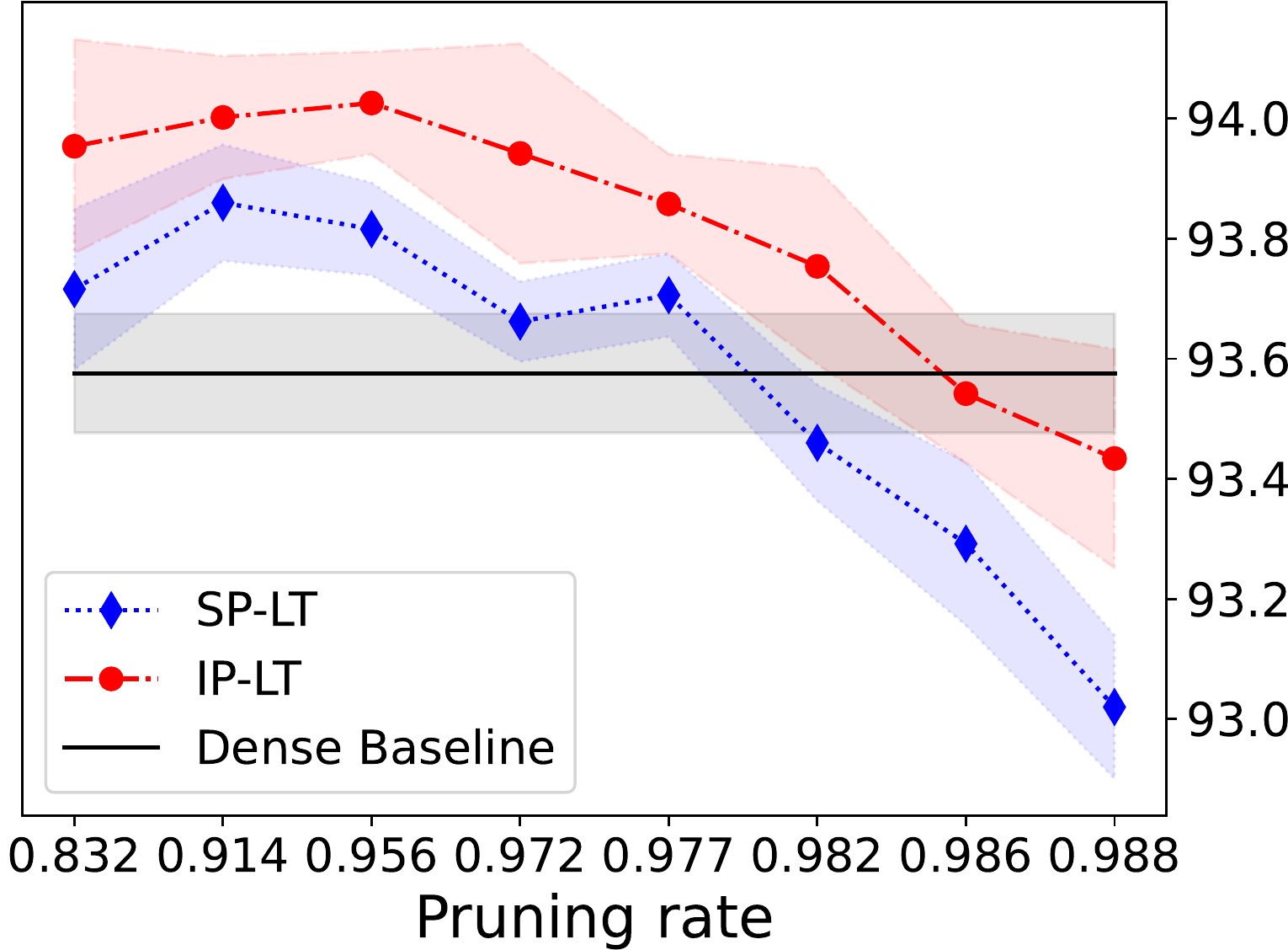}
		\caption{LT for VGG$16$ on CIFAR-$10$}	
		\label{fig:c10_lth}
	\end{subfigure}	
	\caption{Comparison between SP and IP for \subref{fig:c10_set} the DST method SET and \subref{fig:c10_lth} LT on a VGG$16$ trained on CIFAR-$10$.}	
	\label{fig:more_pruning}	
\end{figure}
For SP, more expensive or sophisticated methods like LT and DST improve sparse training results compared to PaI. We want to analyze whether this also applies to the IP setting. Furthermore, we want to check if IP boosts the SOTA methods LT and RigL as well. Finally, we benchmark IP and SP on various SOTA unstructured pruning methods for a ResNet$50$ on ImageNet.

\paragraph{DST and LT on CIFAR-10.}IP improves DST and LTs significantly, see Figs. \ref{fig:more_pruning}\subref{fig:c10_set} and \subref{fig:c10_lth}. For all $p$, IP-LT surpasses SP-LT. IP needs to train $3.7$ times less parameters ($p=0.977$) than SP to reach SP's best result for $p=0.914$. IP-LT matches the dense baseline while training only $1.4\%$ of its parameters and outperforms it for all $p \leq 0.98$. Comparable results hold for SET. IP-SET improves the dense baseline for $p\leq 0.9$, whereas SP-SET only matches it. Similar to PaI, IP-SET greatly exceeds SP-SET for high $p$. Comparing \cref{fig:cifar_basic_experiment,fig:more_pruning} shows that spending more effort in finding the sparse architecture (LT) or adapting it during training (SET) improves performance compared to PaI for both, SP and IP.

\paragraph{ResNet50 on ImageNet.}
\begin{figure}[t!]
	\centering
	\begin{subfigure}[b]{0.225\textwidth}
		\centering
		\includegraphics[width=\textwidth]{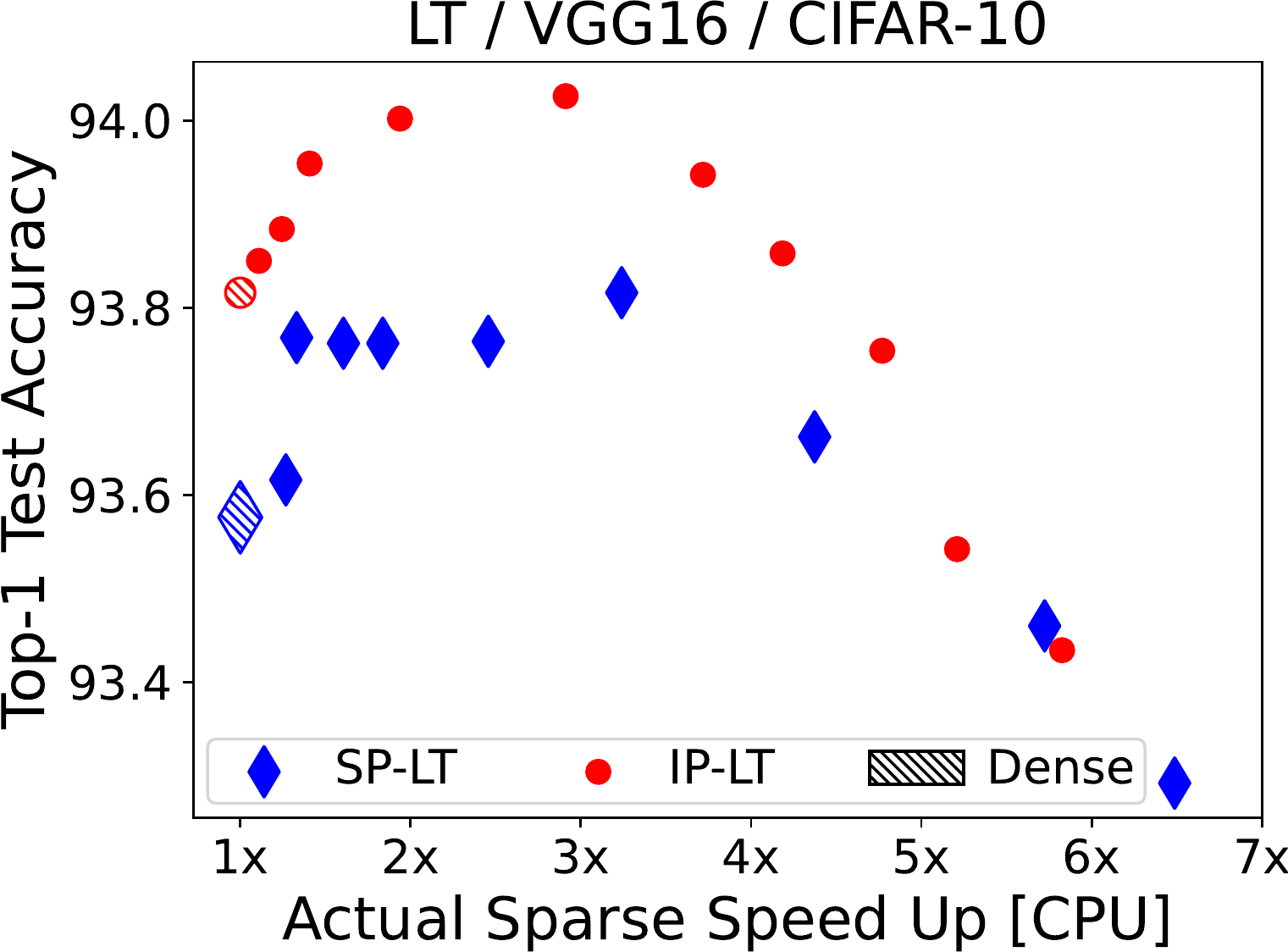}
		\caption{}
		\label{fig:c10_lth_mults}	
	\end{subfigure}
	\hfill
	\begin{subfigure}[b]{0.225\textwidth}
		\centering
		\includegraphics[width=\textwidth]{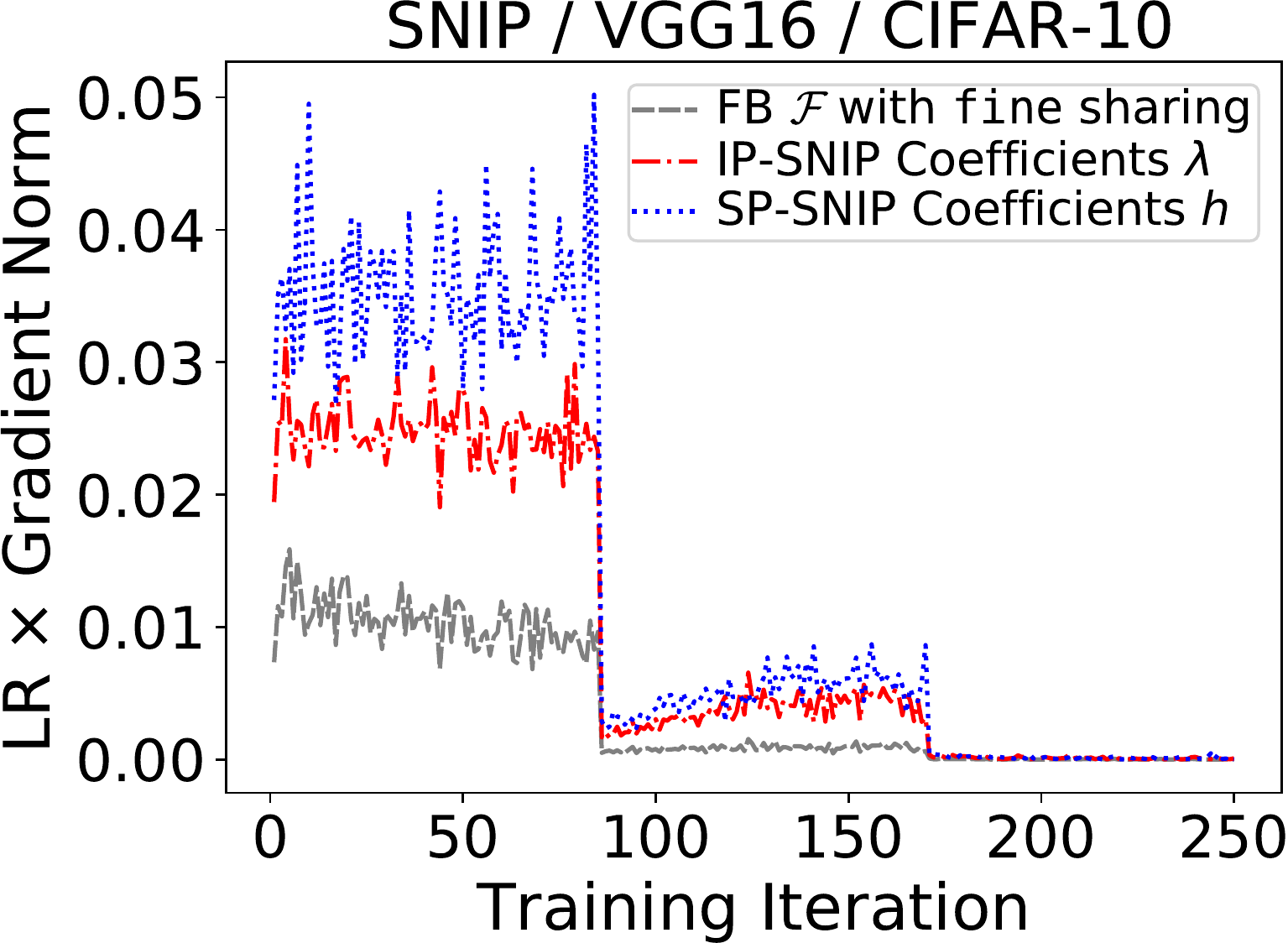}
		\caption{}
		\label{fig:c10_snip_gradflow}	
	\end{subfigure}
	\caption{VGG$16$ on CIFAR-$10$: \subref{fig:c10_lth_mults} Top-1 test accuracy over real time acceleration for IP- and SP-LT. \subref{fig:c10_snip_gradflow} Gradient $L_2$ norm $\times$ learning rate (LR) for SP- and IP-SNIP for layer $1$ and $p=0.85$.}
	\label{fig:ablation}
\end{figure}
\begin{table}[tb!]
	\newcolumntype{P}[1]{>{\centering\arraybackslash}p{#1}}
	\small
	\centering
	\begin{tabular}{@{}p{.145\linewidth}P{.225\linewidth}P{.225\linewidth}P{.225\linewidth}@{}}
		\toprule
		\multicolumn{4}{c}{\textbf{Top-1 Accuracy for ResNet50 on ImageNet}} \\
		Method &\multicolumn{1}{c}{$p=0.0$} & \multicolumn{1}{c}{$p=0.8$}  & \multicolumn{1}{c}{$p=0.9$}\\
		\midrule
		SP-FT & $77.15 \pm 0.04$ & $77.02 \pm 0.03$ & $75.67 \pm 0.09$  \\ 
		IP-FT & $\mathbf{77.30 \pm 0.04}$ & $\mathbf{77.18 \pm 0.01}$ & $\mathbf{75.89 \pm 0.09}$  \\
		\midrule
		SP-GMP & $76.64 \pm 0.06$ & $75.37 \pm 0.01$ & $73.57 \pm 0.06$  \\
		IP-GMP & $\underline{77.16 \pm 0.04}$ & $\underline{75.71 \pm 0.05}$ & $\underline{74.20 \pm 0.07}$  \\
		\midrule
		SP-RigL & $77.15 \pm 0.04$ & $75.75 \pm 0.10$ & $73.88 \pm 0.06$ \\
		IP-RigL & $\mathbf{77.30 \pm 0.04}$ & $\underline{76.03 \pm 0.08}$ & $\underline{74.32 \pm 0.11}$  \\		
		\bottomrule
	\end{tabular}	
	\caption{\label{tab:performance}ResNet$50$ trained on ImageNet for $100$ epochs.}
\end{table}
\Cref{tab:performance} compares IP and SP on the SOTA pruning methods RigL \cite{evci_2020}, GMP \cite{gale_2019} and FT \cite{renda_2020}. As shown, IP outperforms all underlying SP methods for a ResNet$50$ on ImageNet. Results are significantly improved with interspace representations even though more than $50 \%$ of the coefficients of a ResNet$50$ are $1 \times 1$ convolutions which are equivalent for IP and SP. For example, IP-FT has similar performance as a standard dense model while training only $20\%$ of its parameters. Note, using FBs does not only boost training sparse CNNs but dense training too, which will be discussed in more detail in \cref{subsec:generalization}. 

\paragraph{Computational costs.}
Up to now, IP and SP were compared for equal memory costs. As analyzed in \cref{subsec:fb_pruning_description}, IP has a small computational overhead compared to SP for equal sparsity. In applications, the actual runtime is more important than the theoretically required \mults{}. Thus, we compare the performance of IP and SP \wrt the actual acceleration on a CPU achieved by using sparse representations. Details on the implementation are provided in the Appendix \cref{subsec:real_runtime}. IP indeed has a longer runtime for equal sparsity due to the mentioned extra computations. However, by boosting performance of sparse models, IP reaches similar results than dense training with $5.2$ times speed up and better results than SP for equal runtime, see \cref{fig:ablation}\subref{fig:c10_lth_mults}.

\begin{table}[tb!]
	\newcolumntype{P}[1]{>{\centering\arraybackslash}p{#1}}
	\small
	\centering
	\begin{tabular}{@{}lcccccc@{}}
		\toprule
		\multicolumn{5}{c}{\textbf{VGG16 on CIFAR-10}} \\
		&\multicolumn{2}{c}{$p=0.85$} &  \multicolumn{2}{c}{$p=0.99$} \\
		Method & Train & Test &  Train & Test \\
		\cmidrule(r){1-1}\cmidrule(lr){2-3}\cmidrule(l){4-5}
		SP-SET & $99.85$ & $93.45$ &  $94.20$ & $89.36$  \\
		IP-SET & ${99.89}$ & $\mathbf{93.63}$  &  $\mathbf{96.89}$ & $\mathbf{90.92}$ \\
		\midrule
		SP-SNIP & $99.94$ & $93.18$ & $93.96$ & $87.75$  \\
		IP-SNIP & ${99.96}$ & $\mathbf{93.34}$  &  $\mathbf{98.38}$ & $\mathbf{90.79}$ \\
		\specialrule{.5pt}{1.5\aboverulesep}{\belowrulesep}
		\multicolumn{5}{c}{\textbf{ResNet50 on ImageNet}} \\
		&\multicolumn{2}{c}{$p=0.8$} &  \multicolumn{2}{c}{$p=0.9$} \\
		SP-RigL & $74.64$ & $75.75$ & $71.30$ & $73.88$  \\
		IP-RigL & $\mathbf{75.39}$ & $\mathbf{76.03}$  &  $\mathbf{72.08}$ & $\mathbf{74.32}$ \\
		\bottomrule
	\end{tabular}
	\caption{\label{tab:generalization} Generalization gaps for various pruning methods.}
\end{table}

\subsection{Generalization and trainability}\label{subsec:generalization}
We consider generalization as the ability to correctly classify unseen data \cite{lust_2020}. In this context a major aspect is the relationship between performance on the train and test set. Ideally, the performance on the train set should be optimal and a strong indicator for the performance on the test set. The \emph{generalization gap} is the difference between train and test accuracy. Generalization can be improved by regularizations \cite{ioffe_2015,srivastava_2014,caruana_2000,krogh_1991,zhang2016}, enabling the model to use geometrical prior knowledge about the scene \cite{cohen_2016,jaderberg_2015,rath_2020,rath_2022,coors_2018}, shifting the model back to an area where it generalizes well \cite{lust_2020b,lust_2022,ren_2019,serra_2020} but also by pruning the network \cite{bartoldson_2019,lecun_1990,hassibi_1993}.

\Cref{tab:generalization} shows training and test accuracy for the IP- and SP versions of SET and SNIP for a VGG$16$ on CIFAR-$10$ as well as RigL for a ResNet$50$ on ImageNet. IP pruned networks train better than SP pruned ones for all $p$. Note, the used ImageNet training is highly regularized. Thus, the test accuracy is \emph{higher} than the train accuracy. 
For ImageNet and $p=0.99$ on CIFAR-$10$, IP has a bigger generalization gap than SP. This is due to a much better training accuracy for IP, which in the end leads to an improved test accuracy. However, IP has a smaller generalization gap than SP for $p=0.85$ on CIFAR-$10$ where the model overfits.

\Cref{tab:lower_p} further shows that IP can generally improve results for pruning rates where training overfits. Note, $p=0$ is dense training and SP for $p=0$ is standard dense training. Improved performance in the dense setting can not be explained by IP's superior expressiveness (\cref{thm:spd}) since IP and SP can represent the same if all parameters are un-pruned. We hypothesize that correlating filters in a CNN via FB sharing regularizes training, thereby improving generalization. One indicator of this is the fact that correlating \emph{all} filters via \texttt{coarse} sharing shows the best results while \texttt{fine} sharing has comparable results to SP. Consequently, interspace representations can also be used to regularize dense training even for ResNet$50$ on ImageNet, see \cref{tab:performance}. After training, dense interspace representations can be converted to standard ones to reduce computational costs for inference. By optimizing weight decay and initialization schemes, IP's performance can be increased even further, as shown in Appendix \cref{sec:additional_ablations}. 
\begin{table}[tb!]
	\newcolumntype{P}[1]{>{\centering\arraybackslash}p{#1}}
	\small
	\setlength{\tabcolsep}{2.5pt}
	\centering
	\begin{tabular}{@{}lcccc@{}}
		\toprule
		& \multicolumn{4}{c}{Pruning rate $p$} \\
		{Method} & $0.0$ & $0.35$ & $0.6$ & $0.85$ \\
		\specialrule{.5pt}{.75\aboverulesep}{.75\belowrulesep}
		\multicolumn{5}{c}{\footnotesize \textbf{SNIP}} \\
		SP & $93.4 \pm 0.1$ & $93.4 \pm 0.1$ & $93.3 \pm 0.2$ & $93.2 \pm 0.2 $ \\
		IP-\texttt{coarse} & $\mathbf{93.9 \pm 0.2}$ & $\mathbf{93.7 \pm 0.2}$ & $\mathbf{93.8 \pm 0.1}$ & $\mathbf{93.5\pm 0.0}$ \\
		IP-\texttt{medium} & $\mathbf{93.9 \pm 0.2}$  & $93.6 \pm 0.2$ & $93.7 \pm 0.2$ & $93.3 \pm 0.2$ \\
		IP-\texttt{fine}& $93.7 \pm 0.1$ & $93.3 \pm 0.2$ & $93.3 \pm 0.2$ & $93.2 \pm 0.1$\\
		\specialrule{.5pt}{.75\aboverulesep}{.75\belowrulesep}
		\multicolumn{5}{c}{\footnotesize \textbf{SET}} \\
		SP & $93.4 \pm 0.1$ & $93.5 \pm 0.2$ & $93.3 \pm 0.2$ & $93.5 \pm 0.2$  \\
		IP-\texttt{coarse}& $\mathbf{93.9 \pm 0.1}$ & $\mathbf{93.9 \pm 0.2}$ & $\mathbf{93.8 \pm 0.1}$ & $\mathbf{93.7 \pm 0.2}$  \\			
		IP-\texttt{medium}  & $\mathbf{93.9 \pm 0.1}$ & $93.7 \pm 0.2$ & $\mathbf{93.8 \pm 0.1}$ & $93.6 \pm 0.2$  \\
		IP-\texttt{fine} & $93.7 \pm 0.1$ & $93.6 \pm 0.2$ & $93.6 \pm 0.2$ & $93.6 \pm 0.2$  \\
		\bottomrule
	\end{tabular}	
	\caption{\label{tab:lower_p} Varying FB sharing schemes for lower pruning rates $p$.}
\end{table}

\section{Conclusions and directions for future work}\label{sec:conclusions}
IP significantly improves results compared to pruning spatial coefficients. We demonstrate this by achieving SOTA results with the application of IP to SOTA standard PaI, LT, DST as well as classical pruning methods.

\Cref{thm:spd} proofs that IP leads to better sparse approximations than SP. Especially, IP generates models with \emph{higher sparsity and equal performance} than SP. Also, FB representations combined with FB sharing \emph{improve generalization} of overfitting CNNs, even for dense training. This comes with the prize of a small computational overhead for inference and additional gradient computations during training. Nevertheless, we show that sparse interspace representations accelerate dense baselines more than SP while keeping or even improving the baseline's performance.

We believe that IP can be enhanced by adapting more advanced strategies of SDL to the joint training of $\F$ and $\l$. 
Adapting IP to structured pruning is an option to maintain the network's accuracy while reducing inference time for arbitrary soft- and hardware. Combining IP with {low rank tensor approximations} lowers computational costs as well and is discussed in Appendix \cref{sec:additional_ablations,sec:comp_costs}. The interspace representation is an adaptive basis transformation of a finite dimensional vector space. Therefore, FBs $\F$ are not limited to represent convolutional filters but can express arbitrary vectors, like columns or small blocks of a matrix. This makes the concept of IP available for MLPs or self-attention modules.

\section*{Acknowledgements}
The authors would like to thank their colleagues
Julia Lust, Matthias Rath and Rinor Cakaj for their
valuable contributions and fruitful discussions.

{\small
\bibliographystyle{ieee_fullname}
\bibliography{references}

\begin{thebibliography}{10}\itemsep=-1pt

\bibitem{tensorflow}
Mart{\'{\i}}n Abadi, Ashish Agarwal, Paul Barham, Eugene Brevdo, Zhifeng Chen,
  Craig Citro, Gregory~S. Corrado, Andy Davis, Jeffrey Dean, Matthieu Devin,
  Sanjay Ghemawat, Ian~J. Goodfellow, Andrew Harp, Geoffrey Irving, Michael
  Isard, Yangqing Jia, Rafal J{\'{o}}zefowicz, Lukasz Kaiser, Manjunath Kudlur,
  Josh Levenberg, Dan Man{\'{e}}, Rajat Monga, Sherry Moore, Derek~Gordon
  Murray, Chris Olah, Mike Schuster, Jonathon Shlens, Benoit Steiner, Ilya
  Sutskever, Kunal Talwar, Paul~A. Tucker, Vincent Vanhoucke, Vijay Vasudevan,
  Fernanda~B. Vi{\'{e}}gas, Oriol Vinyals, Pete Warden, Martin Wattenberg,
  Martin Wicke, Yuan Yu, and Xiaoqiang Zheng.
\newblock Tensorflow: Large-scale machine learning on heterogeneous distributed
  systems.
\newblock {\em CoRR}, abs/1603.04467, 2016.

\bibitem{aharon_2006}
M. {Aharon}, M. {Elad}, and A. {Bruckstein}.
\newblock K-svd: An algorithm for designing overcomplete dictionaries for
  sparse representation.
\newblock {\em IEEE Transactions on Signal Processing}, 54(11):4311--4322,
  2006.

\bibitem{anwar_2017}
Sajid Anwar, Kyuyeon Hwang, and Wonyong Sung.
\newblock Structured pruning of deep convolutional neural networks.
\newblock {\em ACM Journal on Emerging Technologies in Computing Systems},
  13(3):1--18, 2017.

\bibitem{bartoldson_2019}
Brian Bartoldson, Ari Morcos, Adrian Barbu, and Gordon Erlebacher.
\newblock The generalization-stability tradeoff in neural network pruning.
\newblock In {\em Advances in Neural Information Processing Systems 33}, 2020.

\bibitem{bellec_2018}
Guillaume Bellec, David Kappel, Wolfgang Maass, and Robert Legenstein.
\newblock Deep rewiring: Training very sparse deep networks.
\newblock In {\em International Conference on Learning Representations}, 2018.

\bibitem{blalock_2020}
Davis~W. Blalock, Jose Javier~Gonzalez Ortiz, Jonathan Frankle, and John~V.
  Guttag.
\newblock What is the state of neural network pruning?
\newblock In {\em Proceedings of Machine Learning and Systems 2}, 2020.

\bibitem{cp_2018}
Miguel~A. {Carreira-Perpinan} and Yerlan {Idelbayev}.
\newblock "{Learning}-compression" algorithms for neural net pruning.
\newblock In {\em IEEE/CVF Conference on Computer Vision and Pattern
  Recognition}, 2018.

\bibitem{caruana_2000}
Rich Caruana, Steve Lawrence, and Lee Giles.
\newblock Overfitting in neural nets: Backpropagation, conjugate gradient, and
  early stopping.
\newblock In {\em Advances in Neural Information Processing Systems 13}, 2000.

\bibitem{cohen_2009}
Albert {Cohen}, Wolfgang {Dahmen}, and Ronald {Devore}.
\newblock {Compressed sensing and best k -term approximation}.
\newblock {\em Journal of the American Mathematical Society}, 22(1):211--231,
  2009.

\bibitem{cohen_2016}
Taco~S. Cohen and Max Welling.
\newblock Group equivariant convolutional networks.
\newblock In {\em Proceedings of the 33rd International Conference on Machine
  Learning}, 2016.

\bibitem{coors_2018}
Benjamin Coors, Alexandru~Paul Condurache, and Andreas Geiger.
\newblock Spherenet: Learning spherical representations for detection and
  classification in omnidirectional images.
\newblock In {\em Proceedings of the European Conference on Computer Vision},
  2018.

\bibitem{jorge_2020}
Pau de Jorge, Amartya Sanyal, Harkirat Behl, Philip Torr, Gr{\'e}gory Rogez,
  and Puneet~K. Dokania.
\newblock Progressive skeletonization: Trimming more fat from a network at
  initialization.
\newblock In {\em International Conference on Learning Representations}, 2021.

\bibitem{dettmers_2019}
Tim Dettmers and Luke Zettlemoyer.
\newblock Sparse networks from scratch: Faster training without losing
  performance.
\newblock {\em CoRR}, abs/1907.04840, 2019.

\bibitem{donoho_2006}
D.~L. {Donoho}.
\newblock Compressed sensing.
\newblock {\em IEEE Transactions on Information Theory}, 52(4):1289--1306,
  2006.

\bibitem{elsen_2020}
Erich Elsen, Marat Dukhan, Trevor Gale, and Karen Simonyan.
\newblock Fast sparse convnets.
\newblock In {\em Proceedings of the IEEE/CVF Conference on Computer Vision and
  Pattern Recognition}, 2020.

\bibitem{engan_1999}
K. Engan, S.~O. Aase, and J.~H. Hus{\o}y.
\newblock Method of optimal directions for frame design.
\newblock {\em Proceedings of the IEEE International Conference on Acoustics,
  Speech, and Signal Processing 5}, 1999.

\bibitem{evci_2020}
Utku Evci, Trevor Gale, Jacob Menick, Pablo~Samuel Castro, and Erich Elsen.
\newblock Rigging the lottery: Making all tickets winners.
\newblock In {\em Proceedings of the 37th International Conference on Machine
  Learning}, 2020.

\bibitem{frankle_2018}
Jonathan Frankle and Michael Carbin.
\newblock The lottery ticket hypothesis: Finding sparse, trainable neural
  networks.
\newblock In {\em International Conference on Learning Representations}, 2018.

\bibitem{frankle_2020a}
Jonathan Frankle, Gintare~Karolina Dziugaite, Daniel Roy, and Michael Carbin.
\newblock Linear mode connectivity and the lottery ticket hypothesis.
\newblock In {\em Proceedings of the 37th International Conference on Machine
  Learning}, 2020.

\bibitem{frankle_2021}
Jonathan Frankle, Gintare~Karolina Dziugaite, Daniel Roy, and Michael Carbin.
\newblock Pruning neural networks at initialization: Why are we missing the
  mark?
\newblock In {\em International Conference on Learning Representations}, 2021.

\bibitem{gale_2019}
Trevor Gale, Erich Elsen, and Sara Hooker.
\newblock The state of sparsity in deep neural networks.
\newblock {\em CoRR}, abs/1902.09574, 2019.

\bibitem{gale_2020}
Trevor Gale, Matei Zaharia, Cliff Young, and Erich Elsen.
\newblock Sparse gpu kernels for deep learning.
\newblock In {\em Proceedings of the International Conference for High
  Performance Computing, Networking, Storage and Analysis}, 2020.

\bibitem{gao_2021}
Shangqian Gao, Feihu Huang, Weidong Cai, and Heng Huang.
\newblock Network pruning via performance maximization.
\newblock In {\em Proceedings of the IEEE/CVF Conference on Computer Vision and
  Pattern Recognition}, 2021.

\bibitem{german_1992}
Stuart Geman, Elie Bienenstock, and Rene Doursat.
\newblock Neural networks and the bias/variance dilemma.
\newblock {\em Neural Computation}, 4(1):1–58, 1992.

\bibitem{goyal_2017}
Priya Goyal, Piotr Doll{\'{a}}r, Ross~B. Girshick, Pieter Noordhuis, Lukasz
  Wesolowski, Aapo Kyrola, Andrew Tulloch, Yangqing Jia, and Kaiming He.
\newblock Accurate, large minibatch {SGD:} training imagenet in 1 hour.
\newblock {\em CoRR}, abs/1706.02677, 2017.

\bibitem{guo_2016}
Yiwen Guo, Anbang Yao, and Yurong Chen.
\newblock Dynamic network surgery for efficient dnns.
\newblock In {\em Advances in Neural Information Processing Systems 29}. 2016.

\bibitem{han_2016}
Song Han, Xingyu Liu, Huizi Mao, Jing Pu, Ardavan Pedram, Mark~A. Horowitz, and
  William~J. Dally.
\newblock Eie: Efficient inference engine on compressed deep neural network.
\newblock {\em ACM SIGARCH Computer Architecture News}, 44(3):243--254, 2016.

\bibitem{han_2015}
Song Han, Jeff Pool, John Tran, and William Dally.
\newblock Learning both weights and connections for efficient neural network.
\newblock In {\em Advances in Neural Information Processing Systems 28}. 2015.

\bibitem{hanson_1988}
Stephen~Jose Hanson and Lorien~Y. Pratt.
\newblock Comparing biases for minimal network construction with
  back-propagation.
\newblock In {\em Advances in Neural Information Processing Systems 1}. 1989.

\bibitem{hassibi_1993}
Babak Hassibi and David Stork.
\newblock Second order derivatives for network pruning: Optimal brain surgeon.
\newblock In {\em Advances in Neural Information Processing Systems}, 1992.

\bibitem{He2015}
Kaiming {He}, Xiangyu {Zhang}, Shaoqing {Ren}, and Jian {Sun}.
\newblock Delving deep into rectifiers: Surpassing human-level performance on
  imagenet classification.
\newblock In {\em IEEE International Conference on Computer Vision}, 2015.

\bibitem{he_2016}
Kaiming He, Xiangyu Zhang, Shaoqing Ren, and Jian Sun.
\newblock Deep residual learning for image recognition.
\newblock {\em IEEE Conference on Computer Vision and Pattern Recognition},
  2016.

\bibitem{huang_2018}
Zehao Huang and Naiyan Wang.
\newblock Data-driven sparse structure selection for deep neural networks.
\newblock {\em Proceedings of the European conference on computer vision},
  2018.

\bibitem{ioffe_2015}
Sergey Ioffe and Christian Szegedy.
\newblock Batch normalization: Accelerating deep network training by reducing
  internal covariate shift.
\newblock In {\em Proceedings of the 32nd International Conference on Machine
  Learning}, 2015.

\bibitem{jaderberg_2015}
Max Jaderberg, Karen Simonyan, Andrew Zisserman, and Koray Kavukcuoglu.
\newblock Spatial transformer networks.
\newblock In {\em Advances in Neural Information Processing Systems}, 2015.

\bibitem{janowsky_1989}
Steven~A. Janowsky.
\newblock Pruning versus clipping in neural networks.
\newblock {\em Physical Review A}, 39:6600--6603, 1989.

\bibitem{karnin_1990}
Ehud~D. {Karnin}.
\newblock A simple procedure for pruning back-propagation trained neural
  networks.
\newblock {\em IEEE Transactions on Neural Networks}, 1(2):239--242, 1990.

\bibitem{krizhevsky_2012}
Alex Krizhevsky.
\newblock Learning multiple layers of features from tiny images.
\newblock {\em University of Toronto}, 2012.
\newblock \url{http://www.cs.toronto.edu/~kriz/cifar.html}.

\bibitem{krogh_1991}
Anders Krogh and John~A. Hertz.
\newblock A simple weight decay can improve generalization.
\newblock In {\em Advances in Neural Information Processing Systems 4}. 1992.

\bibitem{lecun_1990}
Yann LeCun, John~S. Denker, and Sara~A. Solla.
\newblock Optimal brain damage.
\newblock In {\em Advances in Neural Information Processing Systems 2}. 1990.

\bibitem{lee_2018}
Namhoon Lee, Thalaiyasingam Ajanthan, and Philip~H.S. Torr.
\newblock {SNIP}: Single-shot network pruning based on connection sensitivity.
\newblock In {\em International Conference on Learning Representations}, 2019.

\bibitem{li_2016}
Hao Li, Asim Kadav, Igor Durdanovic, Hanan Samet, and Hans~Peter Graf.
\newblock Pruning filters for efficient convnets.
\newblock In {\em International Conference on Learning Representations}, 2017.

\bibitem{li_2019}
Yawei Li, Shuhang Gu, Luc Van~Gool, and Radu Timofte.
\newblock Learning filter basis for convolutional neural network compression.
\newblock In {\em IEEE International Conference on Computer Vision}, 2019.

\bibitem{li_2021}
Zhengang Li, Geng Yuan, Wei Niu, Pu Zhao, Yanyu Li, Yuxuan Cai, Xuan Shen,
  Zheng Zhan, Zhenglun Kong, Qing Jin, Zhiyu Chen, Sijia Liu, Kaiyuan Yang, Bin
  Ren, Yanzhi Wang, and Xue Lin.
\newblock Npas: A compiler-aware framework of unified network pruning and
  architecture search for beyond real-time mobile acceleration.
\newblock In {\em Proceedings of the IEEE/CVF Conference on Computer Vision and
  Pattern Recognition}, 2021.

\bibitem{liu_2021b}
Shiwei Liu, Lu Yin, Decebal~Constantin Mocanu, and Mykola Pechenizkiy.
\newblock Do we actually need dense over-parameterization? {In}-time
  over-parameterization in sparse training.
\newblock In {\em Proceedings of the 38th International Conference on Machine
  Learning}, 2021.

\bibitem{liu3_2018}
Xingyu Liu, Jeff Pool, Song Han, and William~J. Dally.
\newblock Efficient sparse-winograd convolutional neural networks.
\newblock In {\em International Conference on Learning Representations}, 2018.

\bibitem{liu2_2018}
Zhenhua Liu, Jizheng Xu, Xiulian Peng, and Ruiqin Xiong.
\newblock Frequency-domain dynamic pruning for convolutional neural networks.
\newblock In {\em Advances in Neural Information Processing Systems 31}, 2018.

\bibitem{louizos_2018}
Christos Louizos, Max Welling, and Diederik~P. Kingma.
\newblock Learning sparse neural networks through l0 regularization.
\newblock In {\em International Conference on Learning Representations}, 2018.

\bibitem{lust_2020b}
Julia Lust and Alexandru~Paul Condurache.
\newblock Gran: An efficient gradient-norm based detector for adversarial and
  misclassified examples.
\newblock In {\em 28th European Symposium on Artificial Neural Networks,
  Computational Intelligence and Machine Learning}, 2020.

\bibitem{lust_2020}
Julia Lust and Alexandru~Paul Condurache.
\newblock A survey on assessing the generalization envelope of deep neural
  networks at inference time for image classification.
\newblock {\em CoRR}, abs/2008.09381, 2020.

\bibitem{lust_2022}
Julia Lust and Alexandru~Paul Condurache.
\newblock Efficient detection of adversarial, out-of-distribution and other
  misclassified samples.
\newblock {\em Neurocomputing}, 470:335--343, 2022.

\bibitem{mairal_2010}
Julien Mairal, Francis Bach, Jean Ponce, and Guillermo Sapiro.
\newblock Online learning for matrix factorization and sparse coding.
\newblock {\em Journal of Machine Learning Research}, 11:19--60, 2010.

\bibitem{malach_2020}
Eran Malach, Gilad Yehudai, Shai Shalev-Schwartz, and Ohad Shamir.
\newblock Proving the lottery ticket hypothesis: Pruning is all you need.
\newblock In {\em Proceedings of the 37th International Conference on Machine
  Learning}, 2020.

\bibitem{mao_2017}
Huizi {Mao}, Song {Han}, Jeff {Pool}, Wenshuo {Li}, Xingyu {Liu}, Yu {Wang},
  and William~J. {Dally}.
\newblock Exploring the granularity of sparsity in convolutional neural
  networks.
\newblock In {\em IEEE Conference on Computer Vision and Pattern Recognition
  Workshops}, 2017.

\bibitem{mocanu_2018}
Decebal Mocanu, Elena Mocanu, Peter Stone, Phuong Nguyen, Madeleine Gibescu,
  and Antonio Liotta.
\newblock Scalable training of artificial neural networks with adaptive sparse
  connectivity inspired by network science.
\newblock {\em Nature Communications}, 9, 2018.

\bibitem{mostafa_2019}
Hesham Mostafa and Xin Wang.
\newblock Parameter efficient training of deep convolutional neural networks by
  dynamic sparse reparameterization.
\newblock In {\em Proceedings of the 36th International Conference on Machine
  Learning}, 2019.

\bibitem{mozer_1989}
Michael~C. Mozer and Paul Smolensky.
\newblock Skeletonization: A technique for trimming the fat from a network via
  relevance assessment.
\newblock In {\em Advances in Neural Information Processing Systems 1}. 1989.

\bibitem{cuda}
NVIDIA, Péter Vingelmann, and Frank~H.P. Fitzek.
\newblock Cuda, release: 10.2.89, 2020.

\bibitem{parashar_2017}
Angshuman Parashar, Minsoo Rhu, Anurag Mukkara, Antonio Puglielli, Rangharajan
  Venkatesan, Brucek Khailany, Joel Emer, Stephen~W. Keckler, and William~J.
  Dally.
\newblock Scnn.
\newblock {\em Proceedings of the 44th Annual International Symposium on
  Computer Architecture}, 2017.

\bibitem{park_2020}
Daniel~S. Park, Yu Zhang, Chung{-}Cheng Chiu, Youzheng Chen, Bo Li, William
  Chan, Quoc~V. Le, and Yonghui Wu.
\newblock Specaugment on large scale datasets.
\newblock In {\em {IEEE} International Conference on Acoustics, Speech and
  Signal Processing}, 2020.

\bibitem{pytorch}
Adam Paszke, Sam Gross, Francisco Massa, Adam Lerer, James Bradbury, Gregory
  Chanan, Trevor Killeen, Zeming Lin, Natalia Gimelshein, Luca Antiga, Alban
  Desmaison, Andreas Kopf, Edward Yang, Zachary DeVito, Martin Raison, Alykhan
  Tejani, Sasank Chilamkurthy, Benoit Steiner, Lu Fang, Junjie Bai, and Soumith
  Chintala.
\newblock Pytorch: An imperative style, high-performance deep learning library.
\newblock In {\em Advances in Neural Information Processing Systems 32}. 2019.

\bibitem{patil_2021}
Shreyas~Malakarjun Patil and Constantine Dovrolis.
\newblock {PHEW}: Constructing sparse networks that learn fast and generalize
  well without training data.
\newblock In {\em Proceedings of the 38th International Conference on Machine
  Learning}, 2021.

\bibitem{pham_2021}
Hieu Pham, Zihang Dai, Qizhe Xie, Minh-Thang Luong, and Quoc~V. Le.
\newblock Meta pseudo labels.
\newblock In {\em IEEE Conference on Computer Vision and Pattern Recognition},
  2021.

\bibitem{ramanujan_2019}
Vivek Ramanujan, Mitchell Wortsman, Aniruddha Kembhavi, Ali Farhadi, and
  Mohammad Rastegari.
\newblock What's hidden in a randomly weighted neural network?
\newblock In {\em IEEE/CVF Conference on Computer Vision and Pattern
  Recognition}, 2020.

\bibitem{rath_2020}
Matthias Rath and Alexandru~Paul Condurache.
\newblock Invariant integration in deep convolutional feature space.
\newblock In {\em 28th European Symposium on Artificial Neural Networks,
  Computational Intelligence and Machine Learning}, 2020.

\bibitem{rath_2022}
Matthias Rath and Alexandru~Paul Condurache.
\newblock Improving the sample-complexity of deep classification networks with
  invariant integration.
\newblock In {\em Proceedings of the 17th International Joint Conference on
  Computer Vision, Imaging and Computer Graphics Theory and Applications},
  2022.

\bibitem{ren_2019}
Jie Ren, Peter~J. Liu, Emily Fertig, Jasper Snoek, Ryan Poplin, Mark Depristo,
  Joshua Dillon, and Balaji Lakshminarayanan.
\newblock Likelihood ratios for out-of-distribution detection.
\newblock In {\em Advances in Neural Information Processing Systems}, 2019.

\bibitem{renda_2020}
Alex Renda, Jonathan Frankle, and Michael Carbin.
\newblock Comparing rewinding and fine-tuning in neural network pruning.
\newblock In {\em International Conference on Learning Representations}, 2020.

\bibitem{imagenet_2012}
Olga Russakovsky, Jia Deng, Hao Su, Jonathan Krause, Sanjeev Satheesh, Sean Ma,
  Zhiheng Huang, Andrej Karpathy, Aditya Khosla, Michael Bernstein,
  Alexander~C. Berg, and Li Fei-Fei.
\newblock {ImageNet Large Scale Visual Recognition Challenge}.
\newblock {\em International Journal of Computer Vision}, 115(3):211--252,
  2015.

\bibitem{schwartz_2019}
Roy Schwartz, Jesse Dodge, Noah~A. Smith, and Oren Etzioni.
\newblock Green {AI}.
\newblock {\em Communications of the ACM}, 63(12):54--63, 2020.

\bibitem{serra_2020}
Joan Serr\`a, David Álvarez, Vicenç Gómez, Olga Slizovskaia, José~F.
  Núñez, and Jordi Luque.
\newblock Input complexity and out-of-distribution detection with
  likelihood-based generative models.
\newblock In {\em International Conference on Learning Representations}, 2020.

\bibitem{simonyan_2014}
Karen Simonyan and Andrew Zisserman.
\newblock Very deep convolutional networks for large-scale image recognition.
\newblock In {\em International Conference on Learning Representations}, 2015.

\bibitem{srivastava_2014}
Nitish Srivastava, Geoffrey Hinton, Alex Krizhevsky, Ilya Sutskever, and Ruslan
  Salakhutdinov.
\newblock Dropout: A simple way to prevent neural networks from overfitting.
\newblock {\em Journal of Machine Learning Research}, 15(56):1929--1958, 2014.

\bibitem{strubell_2020}
Emma Strubell, Ananya Ganesh, and Andrew McCallum.
\newblock Energy and policy considerations for modern deep learning research.
\newblock {\em Proceedings of the AAAI Conference on Artificial Intelligence},
  2020.

\bibitem{szegedy_2016}
Christian {Szegedy}, Vincent {Vanhoucke}, Sergey {Ioffe}, Jonathon {Shlens},
  and Zbigniew {Wojna}.
\newblock Rethinking the inception architecture for computer vision.
\newblock In {\em CVPR}, 2016.

\bibitem{tanaka_2020}
Hidenori Tanaka, Daniel Kunin, Daniel~L Yamins, and Surya Ganguli.
\newblock Pruning neural networks without any data by iteratively conserving
  synaptic flow.
\newblock In {\em Advances in Neural Information Processing Systems 33}, 2020.

\bibitem{tang_2021}
Yehui Tang, Yunhe Wang, Yixing Xu, Yiping Deng, Chao Xu, Dacheng Tao, and Chang
  Xu.
\newblock Manifold regularized dynamic network pruning.
\newblock In {\em Proceedings of the IEEE/CVF Conference on Computer Vision and
  Pattern Recognition}, 2021.

\bibitem{tinney_1967}
W.F. Tinney and J.W. Walker.
\newblock Direct solutions of sparse network equations by optimally ordered
  triangular factorization.
\newblock {\em Proceedings of the IEEE}, 55(11):1801--1809, 1967.

\bibitem{ullrich_2017}
Karen Ullrich, Edward Meeds, and Max Welling.
\newblock Soft weight-sharing for neural network compression.
\newblock In {\em International Conference on Learning Representations}, 2017.

\bibitem{unser_2003}
M. {Unser} and T. {Blu}.
\newblock Mathematical properties of the jpeg2000 wavelet filters.
\newblock {\em IEEE Transactions on Image Processing}, pages 1080--1090, 2003.

\bibitem{verdenius_2020}
Stijn Verdenius, Maarten Stol, and Patrick Forr{\'{e}}.
\newblock Pruning via iterative ranking of sensitivity statistics.
\newblock {\em CoRR}, abs/2006.00896, 2020.

\bibitem{wang_2020}
Chaoqi Wang, Guodong Zhang, and Roger Grosse.
\newblock Picking winning tickets before training by preserving gradient flow.
\newblock In {\em International Conference on Learning Representations}, 2020.

\bibitem{wang_2021}
Chien-Yao Wang, Alexey Bochkovskiy, and Hong-Yuan~Mark Liao.
\newblock Scaled-yolov4: Scaling cross stage partial network.
\newblock In {\em Proceedings of the IEEE/CVF Conference on Computer Vision and
  Pattern Recognition}, 2021.

\bibitem{wang_2021b}
Zi Wang, Chengcheng Li, and Xiangyang Wang.
\newblock Convolutional neural network pruning with structural redundancy
  reduction.
\newblock In {\em Proceedings of the IEEE/CVF Conference on Computer Vision and
  Pattern Recognition}, 2021.

\bibitem{wimmer_2020}
Paul Wimmer, Jens Mehnert, and Alexandru Condurache.
\newblock {FreezeNet}: {Full} performance by reduced storage costs.
\newblock In {\em Proceedings of the Asian Conference on Computer Vision},
  2020.

\bibitem{wimmer_2021}
Paul Wimmer, Jens Mehnert, and Alexandru Condurache.
\newblock {COPS}: {Controlled} pruning before training starts.
\newblock In {\em International Joint Conference on Neural Networks}, 2021.

\bibitem{witten_1987}
Ian~H. Witten, Radford~M. Neal, and John~G. Cleary.
\newblock Arithmetic coding for data compression.
\newblock {\em Commun. ACM}, 30(6):520–540, 1987.

\bibitem{yang_2020}
Huanrui Yang, Wei Wen, and Hai Li.
\newblock {DeepHoyer}: Learning sparser neural network with differentiable
  scale-invariant sparsity measures.
\newblock In {\em International Conference on Learning Representations}, 2020.

\bibitem{yuan_2020}
Yuhui Yuan, Xilin Chen, and Jingdong Wang.
\newblock Object-contextual representations for semantic segmentation.
\newblock In {\em Proceedings of the European conference on computer vision},
  2020.

\bibitem{zhang2016}
Chiyuan Zhang, Samy Bengio, Moritz Hardt, Benjamin Recht, and Oriol Vinyals.
\newblock Understanding deep learning requires rethinking generalization.
\newblock In {\em 5th International Conference on Learning Representations},
  2017.

\bibitem{zhuang_2020}
Tao Zhuang, Zhixuan Zhang, Yuheng Huang, Xiaoyi Zeng, Kai Shuang, and Xiang Li.
\newblock Neuron-level structured pruning using polarization regularizer.
\newblock In {\em Advances in Neural Information Processing Systems 33}, 2020.

\bibitem{zoph_2020}
Barret Zoph, Golnaz Ghiasi, Tsung-Yi Lin, Yin Cui, Hanxiao Liu, Ekin~Dogus
  Cubuk, and Quoc Le.
\newblock Rethinking pre-training and self-training.
\newblock In {\em Advances in Neural Information Processing Systems 33}, 2020.

\end{thebibliography}
}

\clearpage
\setcounter{footnote}{0} 
\setcounter{thm}{0} 
\renewcommand{\thefootnote}{A.\arabic{footnote}}
\renewcommand{\thepage}{A\arabic{page}}  
\renewcommand{\thesection}{A\arabic{section}}
\renewcommand{\thetable}{A\arabic{table}}   
\renewcommand{\thefigure}{A\arabic{figure}}
\renewcommand{\thealgorithm}{A\arabic{algorithm}}
\renewcommand{\theequation}{A.\arabic{equation}}
\renewcommand{\thethm}{A.\arabic{thm}}

\pagenumbering{roman} 

\setcounter{section}{0}
\setcounter{page}{1}
\setcounter{figure}{0}
\setcounter{algorithm}{0}
\setcounter{equation}{0}

\newcounter{bibIndex}

\begin{appendices}
\section{Structure of the Appendix}\label{sec:structural}
The Appendix is divided into the following Sections:
\begin{enumerate}[noitemsep,topsep=-5pt]
	\item[\ref{sec:structural}] Describes the structural organization of the Appendix.
	\item[\ref{sec:additional_ablations}] Contains ablation studies for IP methods which are not shown in the main body of the text.
	\item[\ref{sec:memory}] Discussion about storing unstructured sparse networks. 
	\item[\ref{sec:comp_costs}] Gives detailed information about the computations of FB-CNNs including backpropagation formulas. Especially, a comparison between the number of \mults{} required to evaluate and train a convolutional layer in the spatial and interspace representation is drawn. Finally, details on the real time measurements of sparse speed ups are given.
	\item[\ref{sec:trafo}] Computes transformation rules between the spatial and interspace coefficients, their gradients and Hessian matrices.
	\item[\ref{subsec:pruning_scores}] Here, the computations of the pruning scores used for experiments in the main body of the text are proposed. 
	\item[\ref{sec:init_fbs}] Shows three different Algorithms to initialize FB-CNNs, including the standard initialization scheme used in the main body of the text. Further, details of implementations of the pruning methods are proposed.
	\item[\ref{sec:experimental_setup}] Describes used training setups, hyperparameters, datasets and evaluation procedure for experiments in the main body of the text.
	\item[\ref{sec:architectures}] Presents network architectures, used in the experimental evaluation.
	\item[\ref{sec:proofs}] Concludes the Appendix with a mathematical proof of \cref{thm:spd}.
\end{enumerate}

\section{Additional ablations}\label{sec:additional_ablations}

\subsection{Using different initializations for the interspace}\label{subsec:init}
For SP-SNIP, the problem of vanishing gradients occurs, see \cref{fig:runtime_and_stability}\subref{fig:gradflow}. 
Filters which are spatially too sparse induce a vanishing gradient for high pruning rates. As shown in \cref{fig:pruning}, IP leads to less zeros in the spatial representation of filters than SP \emph{after} training. But, a pruned CNN has a spatially sparse topology \emph{before} training if a {standard} initialization is used. This seems not to be the optimal initial situation for training FBs jointly with their coefficients. To analyze different starting conditions for IP, we initialized the interspace with {standard}, random \texttt{ONB} and \texttt{random} initializations. For details on these different initialization schemes, see \cref{sec:init_fbs}.

Experimental results can be seen in \cref{fig:c10_base_fdnum}\subref{fig:c10_basis_change} for a VGG$16$ trained on CIFAR-$10$. For lower pruning rates, starting with $\B$ and a random ONB behaves similar. For high pruning rates, random ONBs are even better suited to be used. With them, the forward and backward dynamics of a pruned network are not impaired by spatially sparse filters at the beginning of training. Using non-orthonormal FBs leads to worse results than ONBs for lower pruning rates. Elements of a random basis are likely to be more similar to each other than those of ONBs. This redundancy worsens performance for lower pruning rates, but significantly improves results for higher sparsity. 
\begin{figure}[tb!]
	\centering
	\begin{subfigure}{0.225\textwidth}
		\centering
		\includegraphics[width=\textwidth]{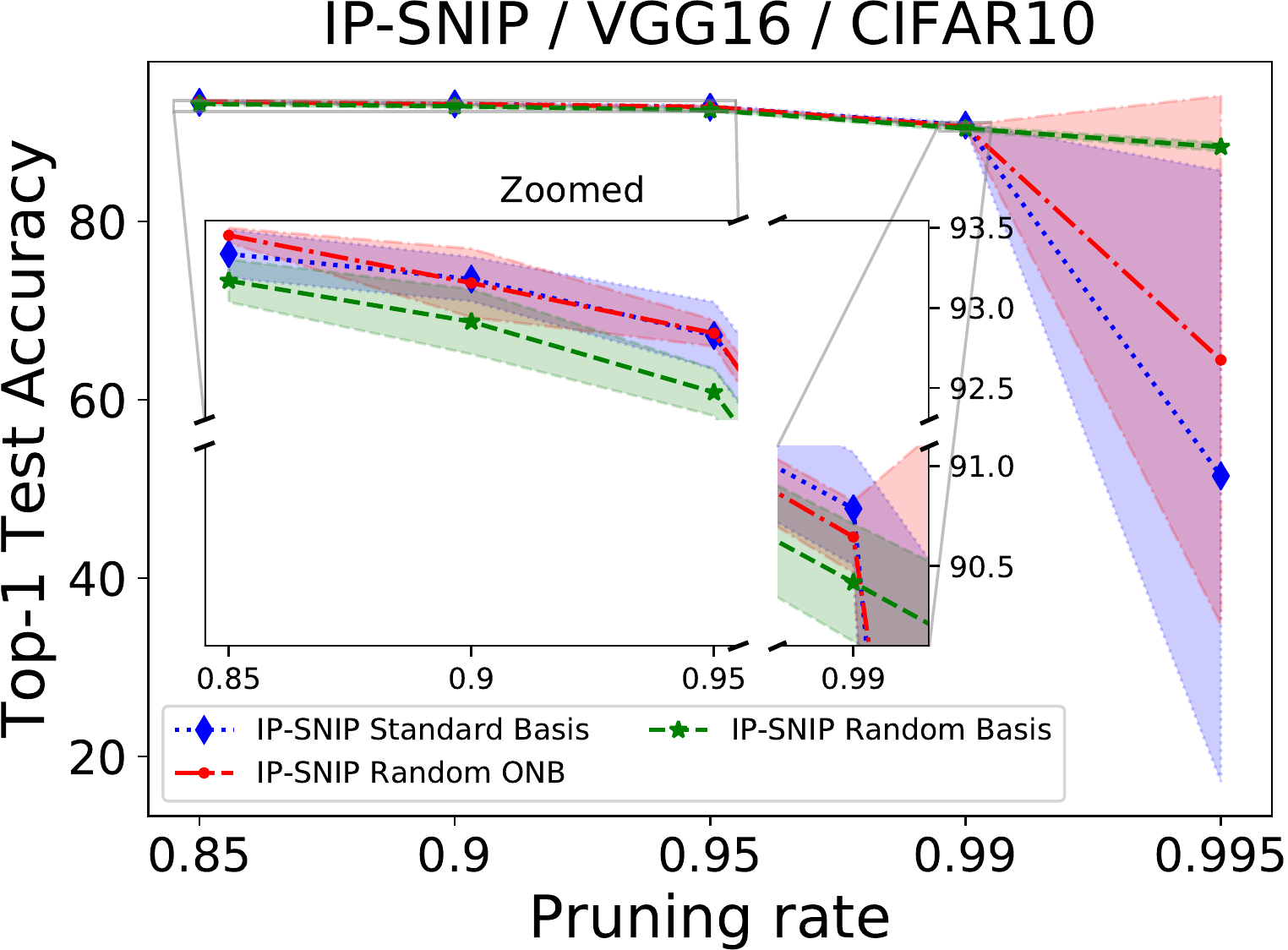}
		\caption{}	
		\label{fig:c10_basis_change}
	\end{subfigure}
	\begin{subfigure}{0.225\textwidth}
		\centering
		\includegraphics[width=\textwidth]{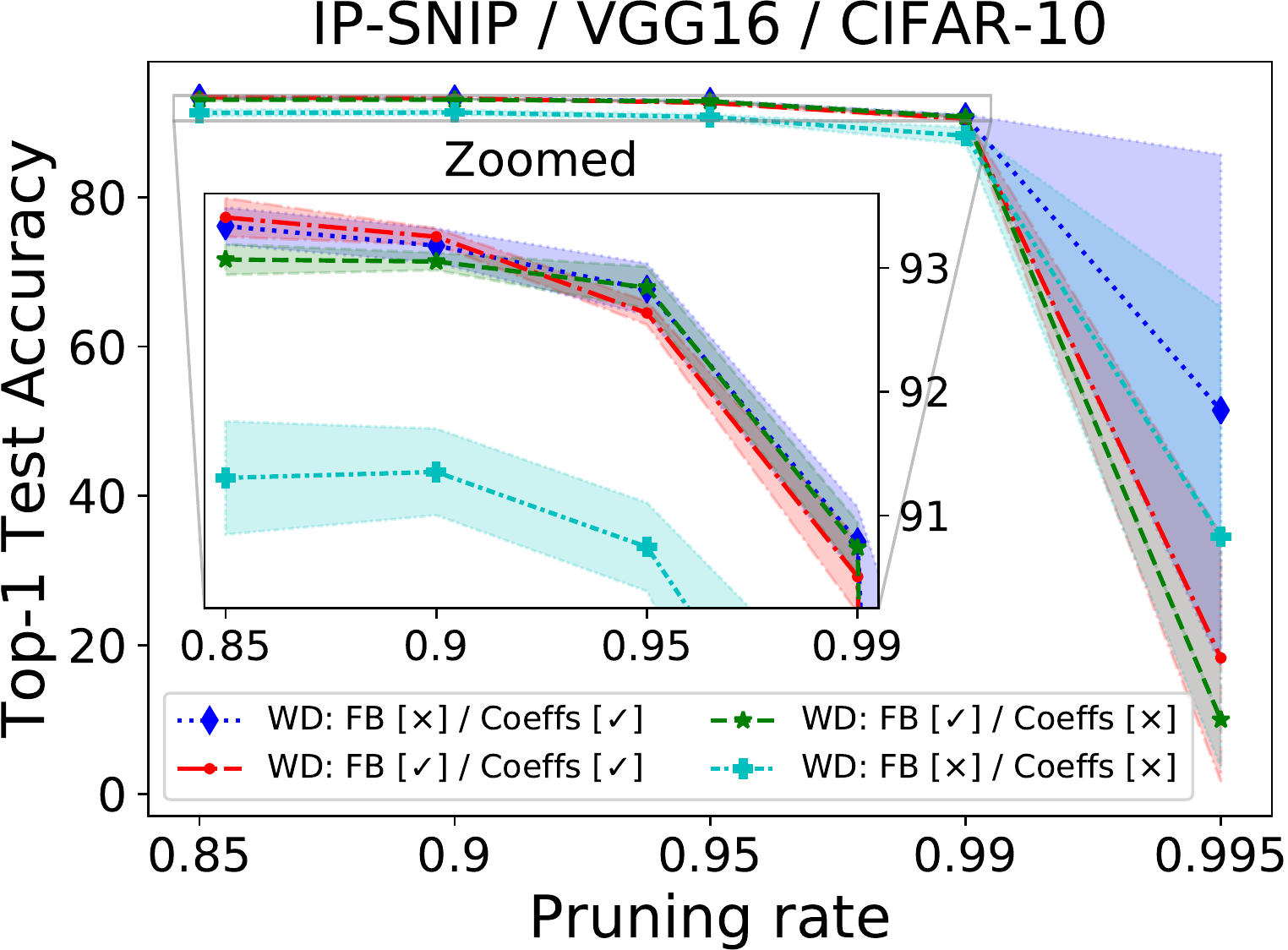}
		\caption{}
		\label{fig:c10_wd}	
	\end{subfigure}
	\caption{Both: IP-SNIP for a VGG$16$ trained on CIFAR-$10$. \subref{fig:c10_basis_change} Standard, random \texttt{ONB} and \texttt{random} initialization are compared. \subref{fig:c10_wd} Weight decay applied $[\checkmark]$ / not applied $[$\xmark$]$ on FBs and FB coefficients.}
	\label{fig:c10_base_fdnum}
\end{figure}

\subsection{Top-5 accuracy for PaI on ImageNet}
\begin{figure}[tb!]
	\centering
	\begin{subfigure}[b]{0.25\textwidth}
		\centering
		\caption*{\textbf{ResNet18 on ImageNet}}
		\includegraphics[width=\textwidth]{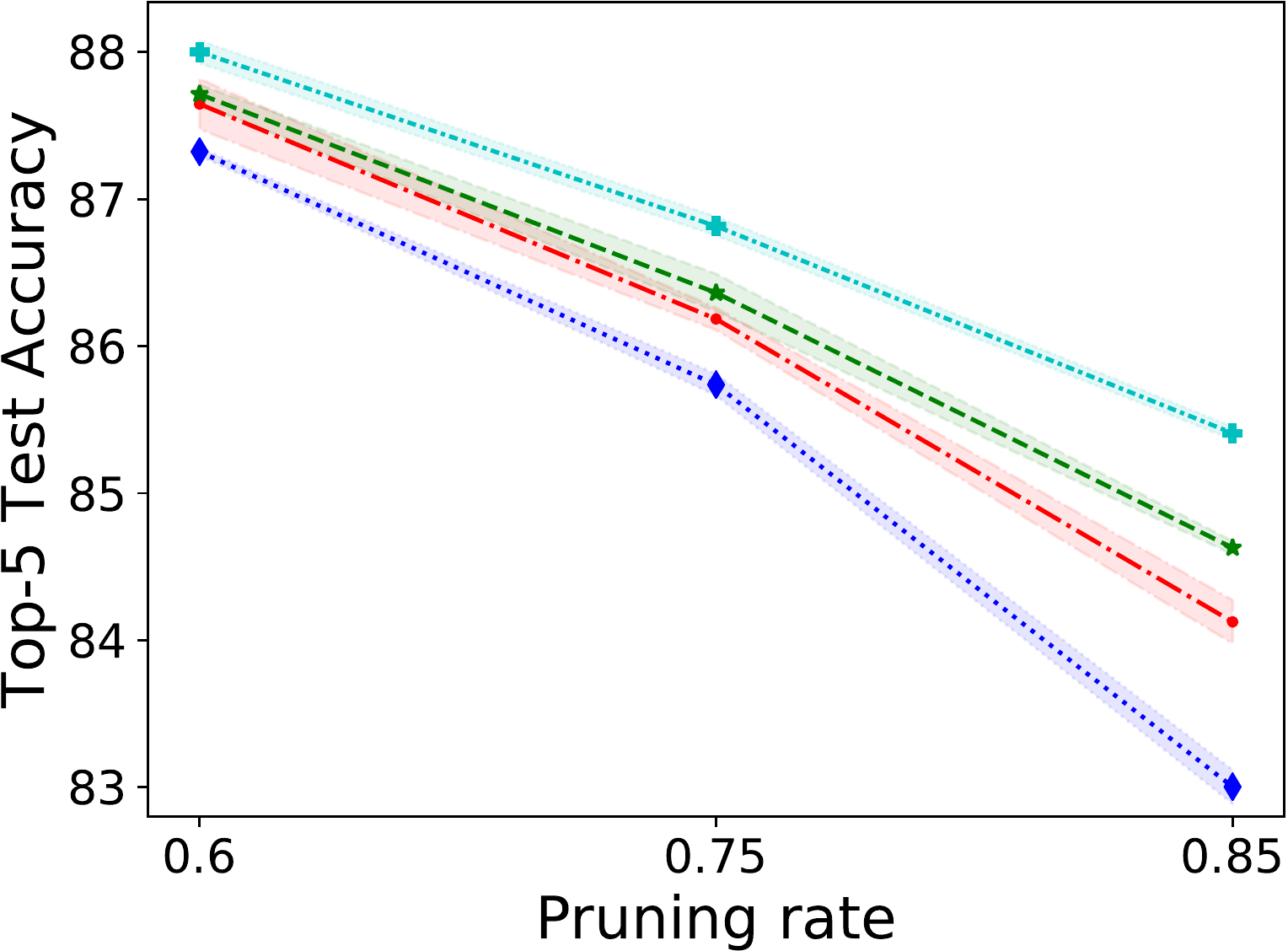}
		\caption{SNIP}	
		\label{fig:imgnet_snip_top5}
	\end{subfigure}	
	
	\begin{subfigure}[b]{0.25\textwidth}
		\centering
		\includegraphics[width=\textwidth]{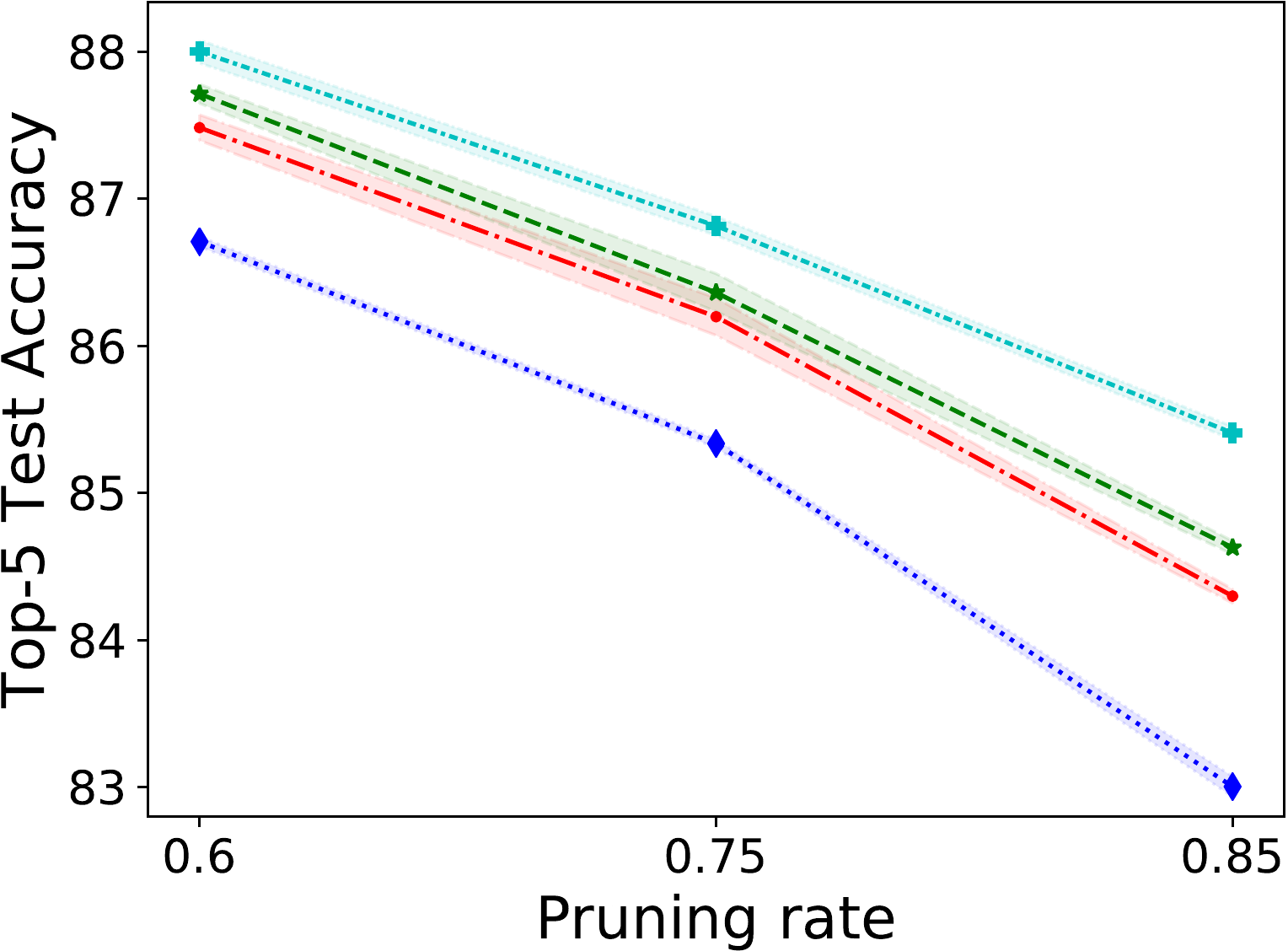}
		\caption{GraSP}	
		\label{fig:imgnet_grasp_top5}
	\end{subfigure}	
	
	\begin{subfigure}[b]{0.25\textwidth}
		\centering
		\includegraphics[width=\textwidth]{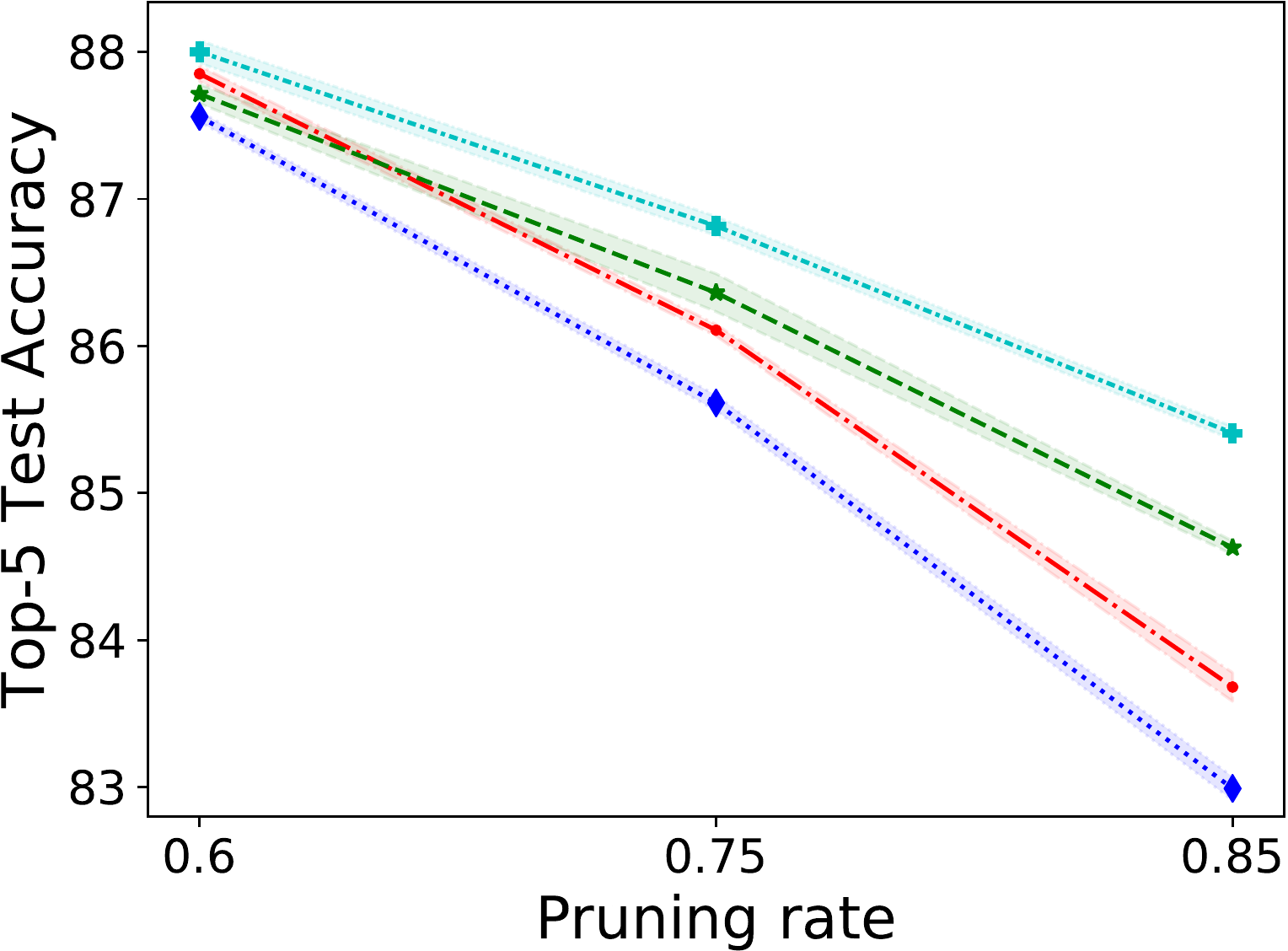}
		\caption{SynFlow}	
		\label{fig:imgnet_synflow_top5}
	\end{subfigure}	
	\begin{subfigure}[b]{\linewidth}
		\centering
		\includegraphics[width=\textwidth]{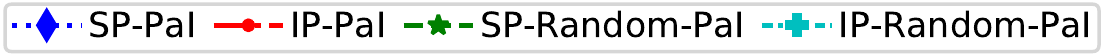} 
	\end{subfigure}
	
	\caption{Comparison between top-$5$ test accuracies for SP on SNIP \subref{fig:imgnet_snip_top5}, GraSP \subref{fig:imgnet_grasp_top5} and SynFlow \subref{fig:imgnet_synflow_top5}, and their adapted IP methods for a ResNet$18$ on ImageNet.}	
	\label{fig:top_5_imgnet}
\end{figure}
Figures \ref{fig:top_5_imgnet}\subref{fig:imgnet_snip_top5}, \subref{fig:imgnet_grasp_top5} and \subref{fig:imgnet_synflow_top5} show the top-$5$ test accuracies for the PaI ImageNet experiment with a ResNet$18$. Using IP instead of SP again improves results significantly as already shown and discussed for top-1 test accuracies in Figs \ref{fig:cifar_basic_experiment}\subref{fig:imgnet_snip}, \subref{fig:imgnet_grasp} and \subref{fig:imgnet_synflow} and \cref{subsec:cifar10}, respectively. Similar to the top-1 accuracy, random PaI reaches better top-5 results than SNIP, GraSP and SynFlow.

\subsection{Impact of weight decay} \label{subsec:ablation_wd}
Weight decay (WD) \cite{krogh_1991} reduces the network's capacity by shrinking parameters smoothly during training. Due to the bias-variance trade-off \cite{german_1992}, WD can help to increase the network's generalization ability. To find the best way to combine WD and IP, we tested all combinations of WD turned on/off for FBs and their coefficients. For this purpose, we used IP-SNIP on VGG$16$ and CIFAR-$10$, see \cref{fig:c10_base_fdnum}\subref{fig:c10_wd}. For lower $p$, not using WD at all yields the worst performance whereas the best results are obtained by applying WD on both, FBs and FB coefficients. For higher $p$, applying WD on the FBs reduces the network's capacity too much. On average, using WD on the FB coefficients but not on the FBs themselves leads to the best results.

\subsection{Similarity of filter bases}
\begin{figure}
	\centering
	\includegraphics[width=.25\textwidth]{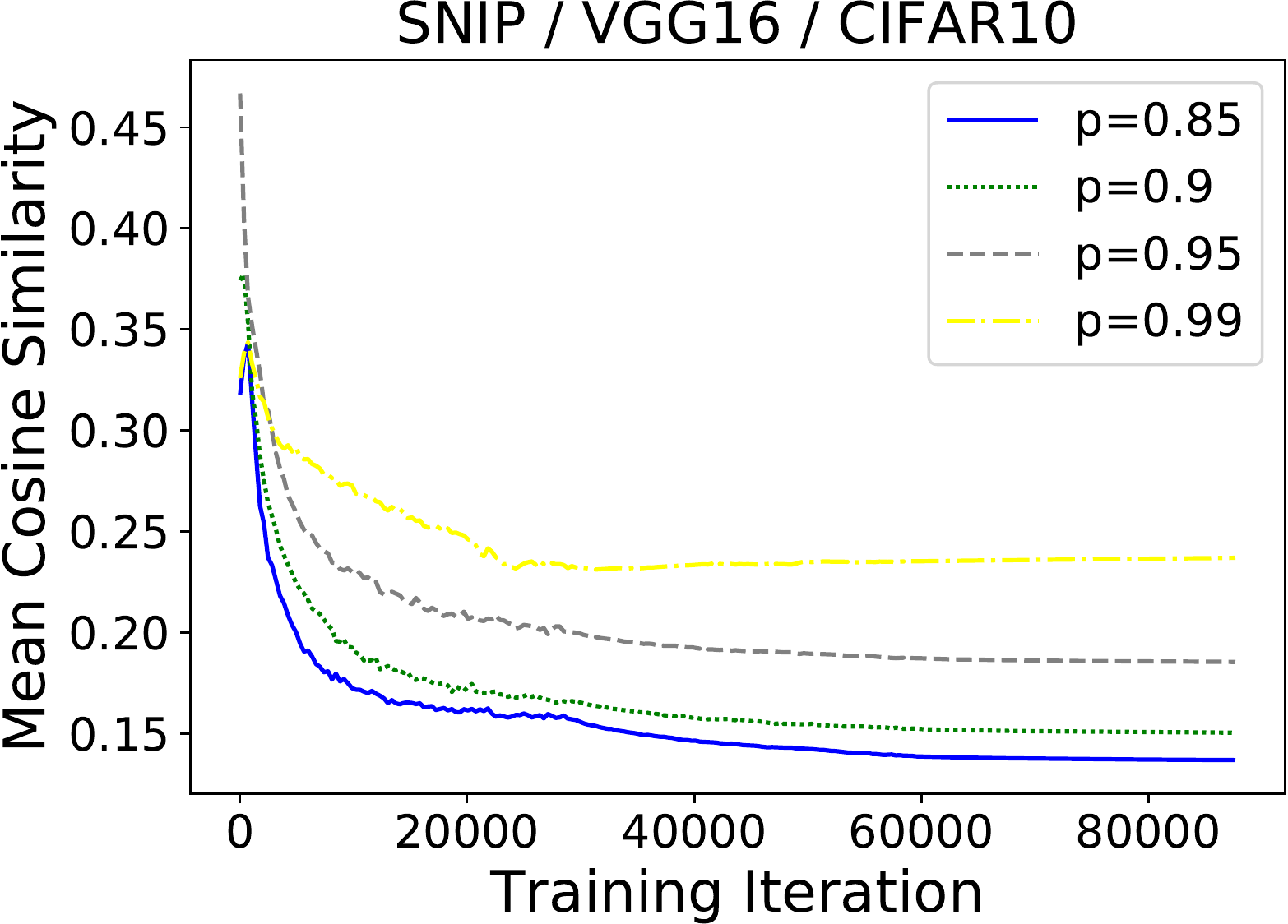}
	\caption{The mean cosine similarity of all elements of the \texttt{coarse} FB for a VGG$16$ pruned with IP-SNIP and trained on CIFAR-$10$ is tracked over training for varying pruning rates.}
	\label{fig:c10_fbnum}
\end{figure}
In \cref{fig:c10_fbnum}, the development of the $\cos$ similarity for the \texttt{coarse} FB $\F$ is tracked at training time for different pruning rates for IP-SNIP with a VGG$16$ trained on CIFAR-$10$. For $\F$, \texttt{random} initialization is used. The $\cos$ similarity of $\F$ is the sum of all absolute values of $\cos$ similarities of distinct elements in $\F$, i.e.
\begingroup\makeatletter\def\f@size{9}\check@mathfonts
\begin{equation}
\frac{2}{K^2 \cdot (K^2 - 1)}\sum_{j=1}^{K^2} \sum_{k = j +1}^{K^2} \frac{\vert \langle \g{j}, \g{k} \rangle \vert}{\Vert \g{j} \Vert_2 \cdot \Vert \g{k} \Vert_2} \;.
\end{equation}
\endgroup
It therefore measures how similar two elements in $\F$ are on average. \Cref{fig:c10_fbnum} shows that the bases have approximately the same similarity at the beginning of training for all pruning rates. For lower pruning rates, the final similarity is much smaller than for higher ones. Therefore, we assume that increasing the FB to more than $9$ filters for lower pruning rates might reduce the number of needed FB coefficients, as  there is ``enough space'' left between the $9$ basis filters. On the other hand, for high pruning rates we should be able to reduce the elements in the FB, since the basis elements tend to assimilate, \ie ``do not need the whole space''. Experimental justifications of these assumptions are shown in \cref{subsec:exp_filter_size}.

\subsection{Layerwise pruning rates for PaI}
\begin{figure}[tb!]
	\centering
	\begin{subfigure}[b]{0.225\textwidth}
		\centering
		\includegraphics[width=\textwidth]{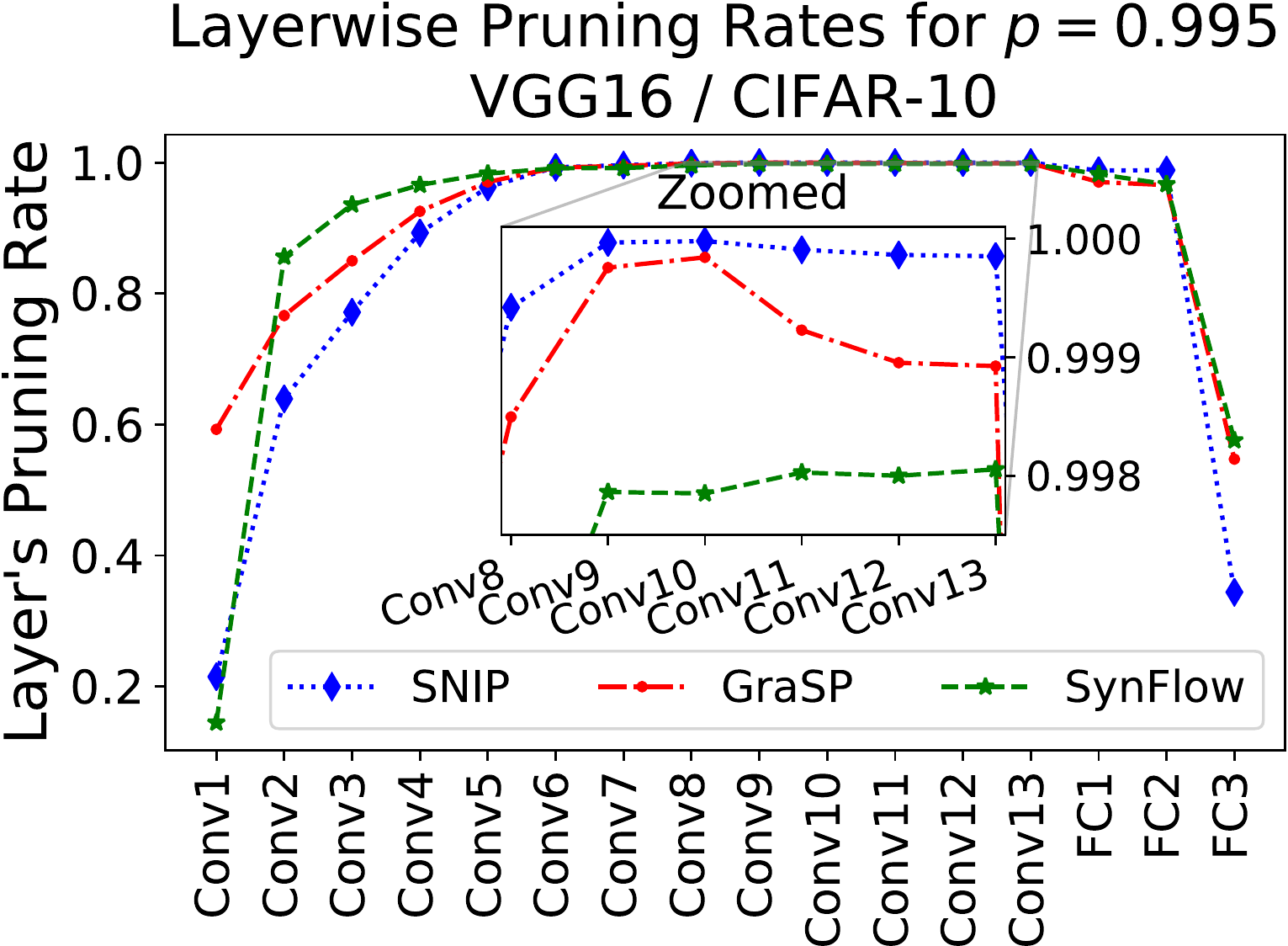}
		\caption{}	
		\label{fig:c10_p_layerwise}
	\end{subfigure}
	\begin{subfigure}[b]{0.225\textwidth}
		\centering
		\includegraphics[width=\textwidth]{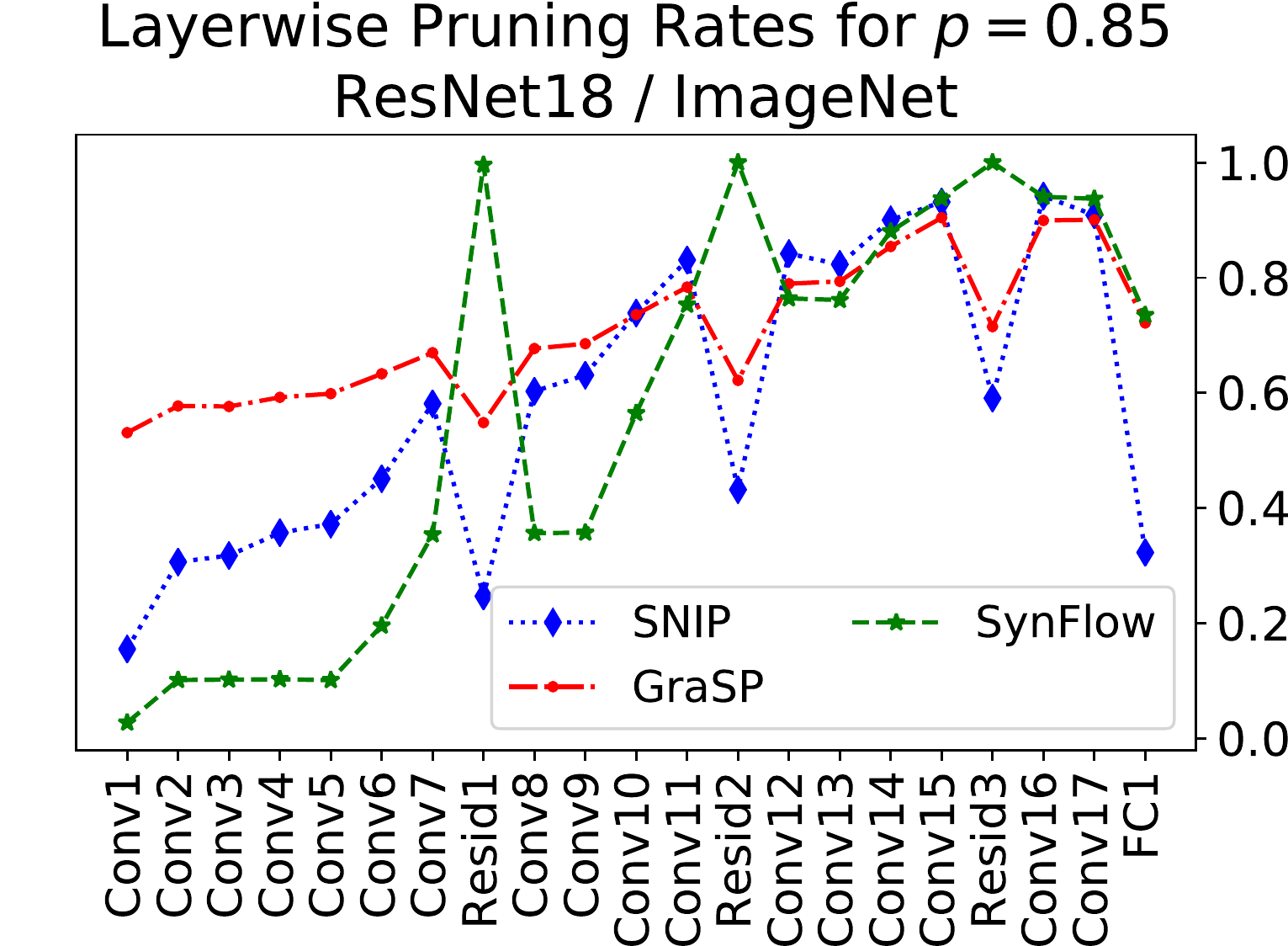}
		\caption{}
		\label{fig:imagnet_p_layerwise}	
	\end{subfigure}
	\caption{Layerwise pruning rates for PaI methods on \subref{fig:c10_p_layerwise} VGG$16$ on CIFAR-$10$ and \subref{fig:imagnet_p_layerwise} ResNet$18$ on ImageNet.}
	\label{fig:layerwise_pruning_rates}
\end{figure}
As shown in \cref{fig:layerwise_pruning_rates}\subref{fig:c10_p_layerwise}, SNIP has the problem of pruning big layers too much. For the VGG$16$, convolutional layers $9$ and $10$ are pruned almost completely. This will lead to a vanishing gradient, see \cref{fig:runtime_and_stability}\subref{fig:gradflow}. With IP, the gradient flow can be increased, but if a layer is pruned completely, even an adaptive basis can not repair the damage.

SynFlow tends to fully prune $1\times 1$ convolutional residual connections in ResNets. As shown in \cref{fig:layerwise_pruning_rates}\subref{fig:imagnet_p_layerwise}, all three residual connections are pruned completely. Consequently, IP- and SP-SynFlow show worse results than IP-/SP-SNIP and GraSP for ResNets, see for example \cref{fig:cifar_basic_experiment}\subref{fig:imgnet_synflow}. Both, SNIP and GraSP prune residual connections even less than surrounding layers.

\subsection{Generalizing filter bases}\label{subsec:exp_filter_size}
\begin{figure}[tb!]
	\centering
	\begin{subfigure}[b]{0.225\textwidth}
		\includegraphics[width=\textwidth]{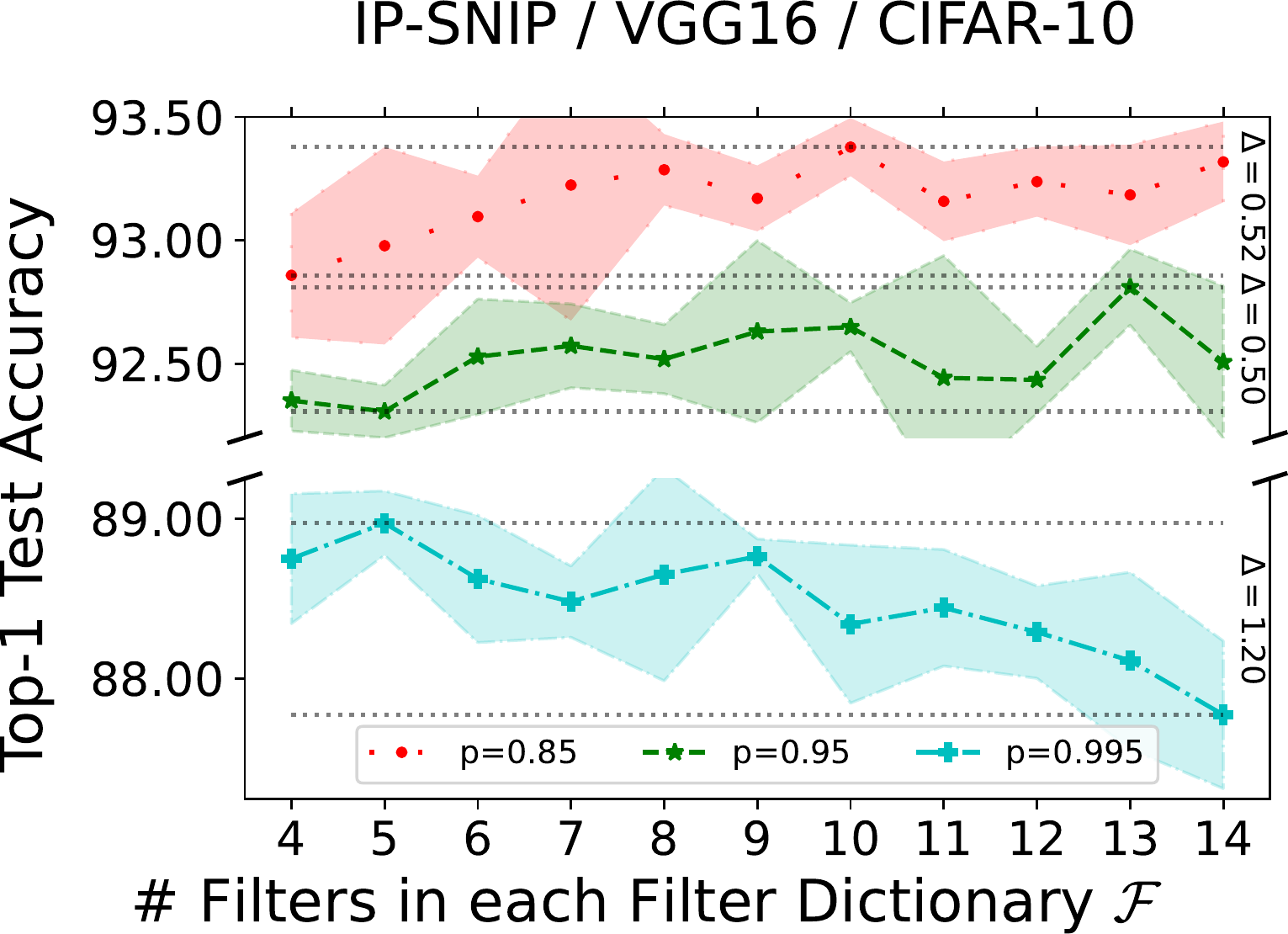}
		\caption{}			
		\label{fig:number_fds}
	\end{subfigure}
	\begin{subfigure}[b]{0.225\textwidth}
		\includegraphics[width=\textwidth]{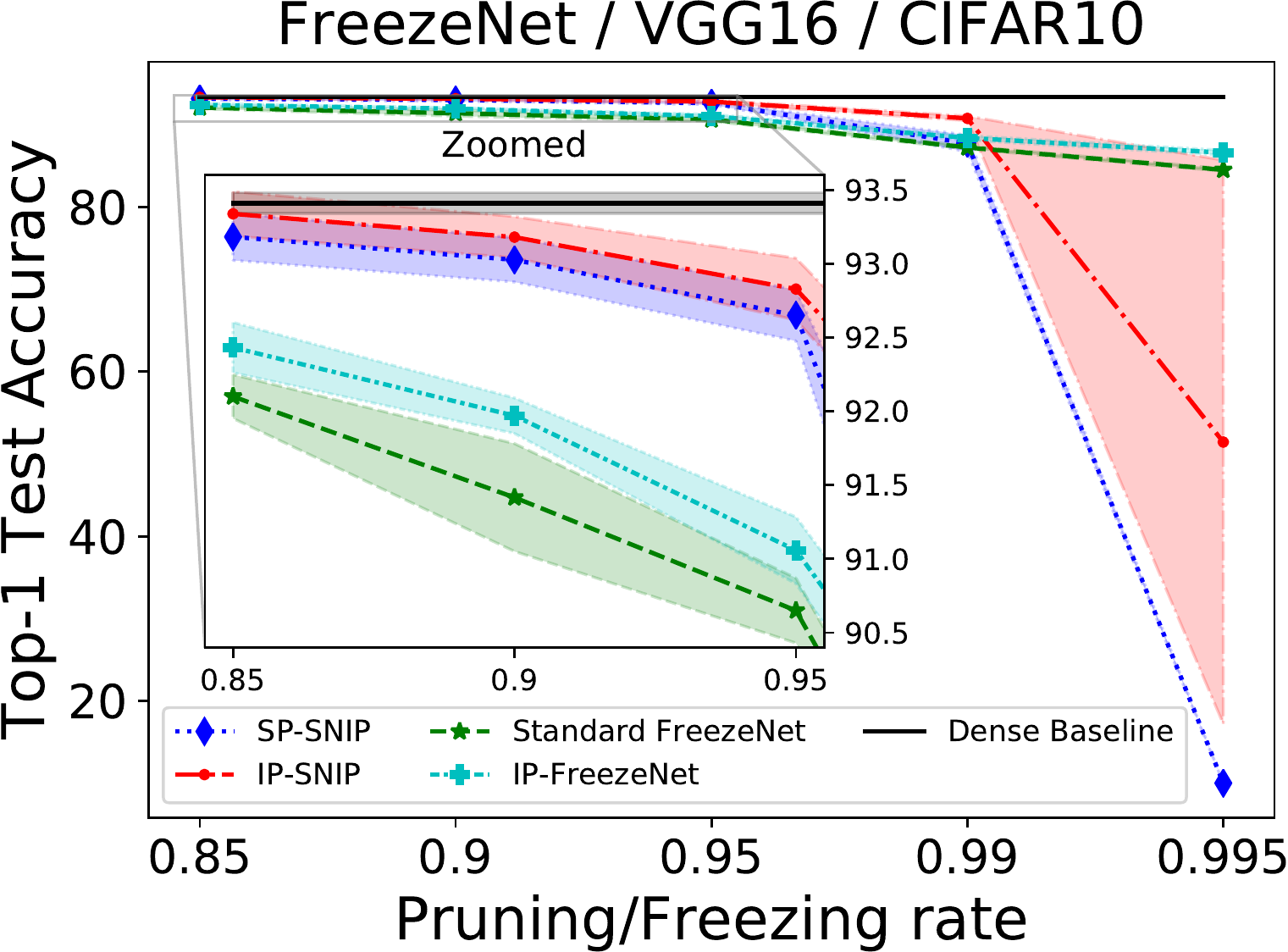}
		\caption{}			
		\label{fig:freezenet}
	\end{subfigure}
	\caption{\subref{fig:number_fds}: Different sizes of FDs for varying pruning rates $p$. \subref{fig:freezenet}: IP and SP versions of freezing parameters compared to pruning them. Frozen/pruned parameters are selected before training by the SNIP criterion.}
	\label{fig:num_freeze}
\end{figure}
Up to now, we discussed experiments where FBs $\F$ formed bases. But, the spanning system $\F \subset \R^{K \times K}$ does not need to form a basis. The interspace can also be spanned by an overcomplete $\F$, \ie $\# \F > K^2$ or an undercomplete $\F$ with $\# \F < K^2$. This leads to the more generalized formulation of \emph{filter dictionaries} (FDs) which include all sizes of $\# \F$. Of course, a FB defines a FD with $\# \F =  K^2$ elements which are additionally assumed to be linearly independent.

As discussed in \cref{sec:comp_costs}, undercomplete FDs can be used to reduce the number of computations needed for a $2$D FB convolution. However, overcomplete FDs might lead to representations of filters needing less coefficients, see \cite{unser_2003,donoho_2006,cohen_2009}. It is not clear which elements of a basis $\B$ should be removed to obtain an undercomplete FD, or added for overcomplete ones. Thus, we initialized all elements of the FDs randomly in this experiment. 

A VGG$16$ contains $3 \times 3$ filters, thus a FB has $9$ filters. \Cref{fig:num_freeze}\subref{fig:number_fds} shows IP-SNIP for a VGG$16$ trained on CIFAR-$10$. Reported results are those with the best validation accuracy from \texttt{coarse}, \texttt{medium} and \texttt{fine} FD sharing. Here, $p$ measures the pruning rate for IP with a FB, \ie $\F = 9$. For $\# \F \neq 9$, the number of non-zero FD coefficients is equal to $\# \F = 9$. Thus, the representation of a filter in the interspace spanned by its dictionary is more sparse if $\# \F > 9$ and less sparse if $\# \F < 9$ compared to $\# \F = 9$. 

More than $9$ elements in a FD improve results if coefficients are not too sparse, \eg $\# \F = 10$ for $p=0.85$ or $\# \F = 13$ for $p=0.95$. Using more sophisticated methods to determine initial FDs might help to exploit overcomplete FDs better. Since $\# \F > 9$ increases the sparsity of FD coefficients, the performance for high pruning rates drops drastically for overcomplete FDs compared to bases. 

If undercomplete FDs are used, performance worsens for lower pruning rates. Here, the capacity of the network is too low as the interspace is only $\# \F$ dimensional. Due to only few non-zero FB coefficients, this is not a limiting factor for high pruning rates anymore. The reduced dimensionality of the interspace even increases performance compared to $\# \F \geq 9$. A reason for this might be the increased information flow induced by a denser structure of the interspace. The best result for $p=0.995$, with test accuracy $88.9\%$, is achieved with $\# \F = 5$. In comparison, SP-SNIP has $10.0\%$ test accuracy for the same number of non-zero parameters. 

\subsection{Freezing coefficients}
FreezeNet \cite{wimmer_2020} is closely related to pruning before training via SNIP \cite{lee_2018}. FreezeNet trains the same parameters as SNIP but \emph{freezes} the un-trained coefficients during training instead of pruning them. By using pseudo random initializations for the network, the frozen coefficients do not have to be stored after training but can be recovered with the used random seed. By always guaranteeing a strong gradient signal, FreezeNet outperforms SNIP significantly for low numbers of trained parameters as shown in \cref{fig:num_freeze}\subref{fig:freezenet}. The opposite is true if more parameters are trained. 

We further compare freezing of spatial coefficients, \emph{standard FreezeNet}, and freezing interspace coefficients, \emph{IP-FreezeNet}. Using adaptive FBs instead of freezing the spatial coefficients again significantly improves performance. Thus, improvements induced by interspace representations are not limited to pruning but also hold for other dimensionality reductions like freezing parts of a CNN during training.

\section{Storing unstructured sparse networks}\label{sec:memory}
\begin{table}[tb!]
	\small
	\centering
	\begin{tabular}{@{}lccc@{}}
		\toprule
		Pruning rate & Size in kB & $\frac{\text{sparse size}}{\text{dense size}} \cdot 100 \%$ & Sparse \& mask \\
		\midrule
		Dense training & $53,256$ & --- & ---\\
		\midrule
		$0.2$ & $60,765$ & $114.1$ & $82.3$\\
		$0.35$ & $49,345$ & $92.7$ & $67.9$\\
		$0.6$ & $30,446$ & $57.2$  & $43.0$\\
		$0.85$ & $11,461$ & $21.5$ & $16.9$ \\
		$0.995$ & $410$ & $0.8$ & $0.6$ \\
		\bottomrule
	\end{tabular}	
	\caption{Compression for the PaI method IP-SNIP {after} training a VGG$16$ on CIFAR-$10$. All stored network parameters are in full precision, \ie $32$bit floating points. Sparse networks are stored in the CSR format whereas the dense one is stored raw. Moreover, dense and sparse networks are compressed by using \texttt{numpy.savez\_compressed}. \emph{Sparse and mask} denotes the percentage of the theoretically needed memory if only sparse parameters are stored together with the entropy encoded pruning mask.} 
	\label{tab:memory}
\end{table}
Storing sparse parameters in formats such as the \emph{compressed sparse row format} (CSR) \cite{tinney_1967} creates additional overhead. The CSR format stores all non-zero elements of a matrix together with an array that contains the column indices and an additional array with the number of elements in each row. Therefore, additional parameters have to be stored for each non-zero element to determine the corresponding column- and row index. However, the two additional arrays do not need to be stored in $32$bit full precision, but only as integers. The CSR format can be used for efficiently computing sparse matrix vector products which we also used for determining the sparse speed up for IP and SP, see \cref{subsec:real_runtime}.

We empirically tested the overhead for real memory costs of sparse networks stored in the CSR format, see \cref{tab:memory}. Note, IP or SP pruned networks have, up to some insignificant differences, equal memory costs in practice and theory. Therefore, we report IP pruned networks in \cref{tab:memory}. Training $0.5 \%$ of all parameters compressed the network to $0.8\%$ of the dense network's size for IP-SNIP with a VGG$16$ \cite{simonyan_2014} trained on CIFAR-$10$. Of course, for such a small number of non-zero elements, the overhead of the CSR format is also quite small. For pruning $85 \%$ of the parameters, $21.8\%$ of the dense memory is needed. As can be seen, additional index memory for sparse row formats increases with a decreasing pruning rate. Thus, for $p \leq 0.5$ CSR will not lead to good compression results and finally even lead to a higher memory requirement than storing the network in a dense format. As shown in \cref{fig:runtime_and_stability}\subref{fig:runtime_vs_sparsity}, using the CSR format for such low pruning rates does not significantly speed up the network inference.

Therefore, other formats for storing the sparse network can be used for lower pruning rates. By storing the pruning mask via entropy encoding, \eg \cite{witten_1987}, at most $1$bit is needed for each mask parameter. To be exact, storing the network's pruning mask for a pruning rate $p \in (0,1)$ ideally needs 
\begin{equation}
1 \geq S = - p \cdot \log_2 p - (1 - p) \cdot \log_2(1-p) \; \text{bits}
\end{equation}
for each element in the mask. If the mask is known, only the non-zero parameters have to be stored in the right order and in full precision. Thus, storing the sparse network of total size $d$ with pruning rate $p$ needs, in the ideal case, $d \cdot (S + (1 - p) \cdot 32)$ bits, compared to $d\cdot32$ bits for the dense network. In total, using entropy encoding for the pruning mask compresses the sparse network to $\nicefrac{S}{32} + (1 - p)$ of its original size. As shown in \cref{tab:memory}, storing the pruning mask together with the non-zero coefficients is cheaper than CSR for all pruning rates.

\section{Comparing computational costs for convolutions with spatial and interspace representations}\label{sec:comp_costs}
For simplicity we will do the analysis with a FB $\F$ consisting of $K^2$ elements in the following. But it is straight forward to do similar computations with an arbitrary FD $\F$ of size $N$.\footnote{Summing from $1$ to $N$ instead of $K^2$ or doing needed computations $N$ times instead of $K^2$ times.} As a results, all computational costs for the interspace setting are multiplied by a factor $\nicefrac{N}{K^2}$ to get the costs for the arbitrary FD case.\footnote{Except the costs for computing $\frac{\partial \loss}{\partial h}$ needed to update $\F$ which are equal for all sizes of $\F$.}  This shows that computations for IP are more expensive if an overcomplete FD with $\# \F > K^2$ is used. On the other hand, by reducing the size of a FD, the computations can be sped up. 

In this Section, we determine the number of \mults{} needed to evaluate a standard 2D convolutional layer and a FB 2D convolutional layer. We use \mults{} as a measure since they are easy to determine and replicable in a mathematical framework but can also be measured in real time applications. A FLOP corresponds to either a multiplication or a summation.

For the forward pass, we show that the number of required \mults{} is increased by a small, constant amount for FB-CNNs compared to standard CNNs for all pruning rates. Since the FB formulation can easily be converted to a standard representation, dense FB-CNNs therefore could be transformed into standard CNNs after training.
If a CNN is pruned, this transformation is not advisable since it usually destroys the sparsity of the network. 

In the backward pass, similar results hold. Moreover, we need to compute the gradient of the FB which of course requires additional resources in the IP setting. 

In the following, we will assume the convolutions to have quadratic $K \times K$ kernels as well as no zero padding, stride $1 \times 1$ and dilation $1 \times 1$.

\subsection{Computations in the forward pass}\label{subsec:computational_costs}
\subsubsection{Standard convolution.}
Let $h = (\hab)_{\a,\b} \in \R^{\co \times \ci \times K \times K}$ denote a convolutional layer of a CNN. Furthermore, let $X = (X^{(\b)})_\b \in \R^{\ci \times h \times w}$ be the input feature map of the corresponding layer. In the following, we determine the number of \mults{} needed to evaluate this layer. In order to do so, we first analyze the costs for one cross-correlation $\star$, used in practice to compute 2D convolutional layer \cite{tensorflow,pytorch}, \ie
\begin{align}
\hab \star X^{(\b)} & = \left ( (\hab \star X^{(\b)})_{i,j} \right)_{i,j} \\ & = \left ( \sum_{m,n=1}^{K} \hab_{m,n} \cdot X^{(\b)}_{m+i,n+j} \right)_{i,j} \in \R^{d_1 \times d_2} \;, \label{eq:cc_one_standard}
\end{align}
with $d_1 := h + 1 - K$ and $d_2 := w + 1 - K$, the dimensions of the output.
\Cref{eq:cc_one_standard} shows that the cost for one cross-correlation is given by $2 \cdot K^2 \cdot d_1\cdot d_2 $ \mults{}. The output of a 2D convolutional layer is given by
\begin{equation}
Y = \left (Y^{(\a)} \right )_\a = \left ( \sum_{\b=1}^{\ci} \hab \star X^{(\b)} \right )_\a \in \R^{\co \times d_1 \times d_2}\;,
\end{equation}
which finally leads to $\co \cdot \ci$ times the costs to compute a single cross-correlation \cref{eq:cc_one_standard}. Therefore, $2 \cdot \co \cdot \ci  \cdot K^2 \cdot d_1 \cdot d_2$ \mults{} are needed in total to compute a standard 2D convolutional layer.

\subsubsection{FB convolution.}
Let $h = (\hab)_{\a,\b} = (\sum_n \labn \cdot \gn)_{\a,\b} \in \R^{\co \times \ci \times K \times K}$ be the interspace representation of $h$, where the FB is given by $\F = \{g^{(1)}, \ldots, g^{(K^2)}\} \subset \R^{K\times K}$. The 2D convolution of this layer with input $X \in \R^{\ci \times h \times w}$ can be computed via
\begin{equation}
Y = \left (Y^{(\a)} \right )_\a =  \left (\sum_{\b=1}^{\ci} \sum_{n=1}^{K^2} \labn \cdot \left (\gn \star X^{(\b)}\right ) \right )_\a\;.\label{eq:cc_fb}
\end{equation}
Using the last equation in \cref{eq:cc_fb}, we see that $\gn \star X^{(\b)}$ has to be computed once for each combination of $\b$ and $n$, \ie $\ci \cdot K^2$ many times. The costs for computing all $\gn \star X^{(\b)}$ is therefore given by $2 \cdot \ci \cdot K^4 \cdot d_1 \cdot d_2$ \mults{}. For each combination of $\a,\b$ and $n$, $\gn \star X^{(\b)}$ has to be multiplied by the scalar $\labn$. These are $d_1 \cdot d_2$ many \mults{} for each $\a,\b$ and $n$. Summing over $\b$ and $n$ yields another $\co \cdot \ci \cdot K^2 \cdot d_1 \cdot d_2$ \mults{} in total. Thus, the total costs for computing a FB 2D convolutional layer is given by $2 \cdot \co \cdot \ci \cdot K^2 \cdot d_1 \cdot d_2 + 2\cdot \ci \cdot K^4 \cdot d_1 \cdot d_2$ \mults{}. 

By using FB convolutions, the numbers of needed \mults{} is therefore slightly increased by $2 \cdot \ci \cdot K^4 \cdot d_1 \cdot d_2$. Which is a relative increase of $\nicefrac{K^2}{\co} \cdot 100 \% $ compared to the standard case. 

\subsubsection{Pruned networks.}\label{subsec:subsec_appendix_addmult_pruned}
In the following we assume the convolutional layer $h \in \R^{\co \times \ci \times K \times K}$ to be pruned with a pruning rate of $p \in [0,1]$.\footnote{For simplicity, we assume the number of non-zero coefficients for IP and SP to be equal here. Due to extra FB parameters, the number of non-zero interspace coefficients is always slightly smaller than for the standard case in our experiments.} We suppose all zero coefficients to be known. Thus, the corresponding multiplications do not have to be computed in \cref{eq:cc_one_standard,eq:cc_fb}. 

The required number of computations for a standard 2D convolutional layer with pruning rate $p$ is therefore given by 
\begin{equation}
2 \cdot \co \cdot \ci  \cdot K^2 \cdot d_1 \cdot d_2 \cdot (1-p) \; \mults{}\;. \label{eq:forward_sp}
\end{equation}
For a pruned FB 2D convolutional layer, 
\begin{equation}
2 \cdot \co \cdot \ci \cdot K^2  \cdot d_1 \cdot d_2 \cdot (1-p) + 2 \cdot \ci  \cdot K^4 \cdot d_1 \cdot d_2 \; \mults{}\ \label{eq:forward_ip}
\end{equation}
are needed for evaluation.

The number of \mults{} for IP is increased for all pruning rates by $2 \cdot \ci  \cdot K^4 \cdot d_1 \cdot d_2$ compared to SP. These are exactly the costs for computing all combinations of $\gn \star X^{(\b)}$, needed for the forward pass for FB 2D Convolutions. These costs are independent of the pruning rate and therefore a constant overhead of IP compared to SP. Thus the additional costs for IP in the forward pass compared to SP are $\nicefrac{K^2}{\co}$ times the costs of the dense forward pass.

\subsection{Backward Pass}\label{subsec:comp_cost_backprop}
Up to now, we have computed additional \mult{} costs for IP compared to SP in the forward pass. Now we want to have a closer look at the backward pass. We note that $\frac{\partial \loss}{\partial Y}$ always has the same cost for the standard- and the FB 2D convolution layer. This holds since $\hat{X} = \sigma ({Y})$ for some activation function $\sigma$ and consequently $\frac{\partial \loss}{\partial Y} = \frac{\partial \loss}{\partial \hat{X}} \odot \sigma^\prime (Y)$. 

\subsubsection{Computing the gradient for ${X}$.}
Furthermore, it is known that
\begin{equation}
\frac{\partial \loss}{\partial X^{(\b)}} = \sum_{\a=1}^{\co}  \hab \hat{\star} \frac{\partial \loss}{Y^{(\a)}}
\end{equation}
with a strided convolution $\hat{\star}$ that corresponds to the forward pass and which needs $2 \cdot \co \cdot \ci \cdot K^2 \cdot h \cdot w$ \mults{}. By representing $\hab = \sum_{n=1}^{K^2} \labn \cdot \gn$ and using the linearity of $\hat{\star}$, we now get the computational overhead of $2 \cdot \co \cdot K^4 \cdot h \cdot w$ \mults{} which are the costs for computing 
\begin{equation}
\gn \hat{\star} \frac{\partial \loss}{Y^{(\a)}}
\end{equation}
for all $n \in \{1, \ldots, K^2\}$ and $\a \in \{1, \ldots, \co\}$. This results in an overhead of $\nicefrac{K^2}{\ci}$ compared to the costs of the standard, dense network. 

In the sparse case, again the \mult{} costs for SP are decreased by a factor $(1-p)$. Furthermore, the overhead $\nicefrac{K^2}{\ci}$ is constant since the $\gn$ are not pruned. Similar formulas to \cref{eq:forward_sp,eq:forward_ip} hold also in the backpropagation case which results in a constant overhead of IP compared to SP for computing $\frac{\partial \loss}{\partial X}$ equal to $\nicefrac{K^2}{\ci}$ times the costs of the dense computation of $\frac{\partial \loss}{\partial X}$.

\subsubsection{Gradients for coefficients.}
The backpropagation formulas for the spatial and FB coefficients are given by
\begin{equation}
\frac{\partial \loss}{\partial \hab_{i,j}} = \left (\frac{\partial \loss}{\partial Y^{(\a)}} \star X^{(\b)} \right )_{i,j} 
\end{equation}
and
\begin{equation}
\frac{\partial \loss}{\partial \labn} = \left \langle \frac{\partial \loss}{\partial Y^{(\a)}}, \gn \star X^{(\b)} \right \rangle \;, \label{eq:grad_fb_coeff1}
\end{equation}
respectively. Since $\gn \star X^{(\b)}$ is already computed in the forward pass, both computations for the standard case and the FB representation have equal \mult{} costs. In total, this equals to $2 \cdot \co \cdot \ci \cdot K^2 \cdot d_1 \cdot d_2$ \mults{} for computing $\frac{\partial \loss}{\partial h}$ or $\frac{\partial \loss}{\partial \l}$. If pruning is applied, this reduces to $2 \cdot \co \cdot \ci \cdot K^2 \cdot d_1 \cdot d_2 \cdot (1 - p)$ \mults{} for IP and SP, since gradients for pruned coefficients do not need to be computed.

Note, if the size $d_1 \cdot d_2$ of $Y \in \R^{\co \times d_1 \times d_2}$ is bigger than the kernel size $K^2$ it is even cheaper to compute the gradient of $\labn$ via
\begin{equation}
\frac{\partial \loss}{\partial \labn} = \left \langle \gn , \frac{\partial \loss}{\partial Y^{(\a)}} \star X^{(\b)} \right \rangle \;.\label{eq:grad_fb_coeff2}
\end{equation}
In \cref{eq:grad_fb_coeff1} there are $2 \cdot d_1 \cdot d_2$ \mults{} needed (if $\gn \star X^{(\b)}$ is known which we can assume due to the forward pass) whereas \cref{eq:grad_fb_coeff2} needs $2 \cdot K^2$ \mults{} \emph{if} $\frac{\partial \loss}{\partial Y^{(\a)}} \star X^{(\b)}$ is known. As we will see in the following, $\gn$ needs $\frac{\partial \loss}{\partial Y^{(\a)}} \star X^{(\b)}$ to be computed for all $\a, \b$ and consequently we can assume them to be known.
In summary we can say that the computation of the interspace coefficients $\l$ requires the same number of \mults{}, or even less, compared to the spatial coefficients.

\subsubsection{Gradient for the filter base.}
The computation of the gradients $\frac{\partial \loss}{\partial \gn}$ also generates extra costs for the backward pass of training interspace representations. It holds
\begin{equation}
\frac{\partial \loss}{\partial \gn} = \sum_{\a = 1}^{\co} \sum_{\b = 1}^{\ci} \labn \cdot \left ( \frac{\partial \loss}{\partial Y^{(\a)}} \star X^{(\b)} \right ) \;.\label{eq:grad_gn}
\end{equation}
As shown in \cref{eq:grad_gn}, $\frac{\partial \loss}{\partial \gn}$ first needs to compute all $\frac{\partial \loss}{\partial Y^{(\a)}} \star X^{(\b)} = \frac{\partial \loss}{\partial \hab}$. This is exactly the cost for computing the dense gradient $\frac{\partial \loss}{\partial h}$ which needs $2 \cdot \ci \cdot \co \cdot K^2 \cdot d_1 \cdot d_2$ \mults{}. The scaling and summation in the sum \cref{eq:grad_gn} requires $2 \cdot \ci \cdot \co \cdot K^2$ \mults{} in total. If pruning is applied, this reduces to $2 \cdot \ci \cdot \co \cdot K^2 \cdot (1-p)$. Altogether, computing the gradients of $g^{(1)}, \ldots, g^{(K^2)}$ needs $2 \cdot \co \cdot \ci \cdot K^2 \cdot (d_1 \cdot d_2 + (1 - p) \cdot K^2)$ \mults{}. In simple terms, the total computation of $\frac{\partial \loss}{\partial g}$ lies in $\mathcal{O}(costs(\frac{\partial \loss}{\partial h}))$.  

\subsubsection{Summary for the backward pass.}
In summary, the computation of $\frac{\partial \loss}{\partial X}$ of IP induces a constant overhead compared to IP. This corresponds to $\nicefrac{K^2}{\ci}$ times the costs of computing the dense gradient of $\frac{\partial \loss}{\partial X}$ by using spatial coefficients. On top of that, IP also needs to compute the gradient for the FB $\F$ which is in $\mathcal{O}(costs(\frac{\partial \loss}{\partial h}))$.

\subsection{Upper bounds for gradients}\label{subsec:upper_grad_bounds}
As \cref{eq:grad_fb_coeff2} and \cref{eq:grad_gn} show, jointly optimizing $\F$ and $\l$ leads to non trivial correlations between them. With a slight abuse of notation we assume for the following discussion $\F$ to be the $K^2 \times K^2$ matrix containing all flattened $\gn$. Further, let $h, \l \in \R^{K^2 \times \co \ci}$ contain all spatial and interspace coefficients of the layer, respectively. Therefore, it holds $h = \F \cdot \l$. By using $\frac{\partial \loss}{\partial \hab} = \frac{\partial \loss}{\partial Y^{(\a)}} \star X^{(\b)}$ and the Cauchy-Schwartz inequality, the gradients for $\F$ and $\l$ are bounded by
\begin{equation}
\left \Vert \frac{\partial \loss}{\partial \l} \right \Vert_F \leq \left \Vert \F \right \Vert_F \left \Vert \frac{\partial \loss}{\partial h} \right \Vert_F \; \text{and} \; \left \Vert \frac{\partial \loss}{\partial \F} \right \Vert_F  \leq  \Vert \l \Vert_F \left \Vert \frac{\partial \loss}{\partial h} \right \Vert_F \;.
\end{equation}
This shows that upper bounds for $\frac{\partial \loss}{\partial \F}$ and $\frac{\partial \loss}{\partial \l}$ are determined by the spatial gradient $\frac{\partial \loss}{\partial h}$. This boundedness of the gradients leads to stable convergence for both, $\F$ and $\l$, while the convergence behavior of $\l$ is similar to the standard coefficients $h$, see \cref{fig:ablation}\subref{fig:c10_snip_gradflow}. Moreover, \cref{fig:runtime_and_stability}\subref{fig:gradflow} even shows that adaptive FBs help to overcome vanishing gradients for SNIP by becoming spatially dense. IP-SNIP can use that to recover during training from a complete, PaI induced information loss, while SP-SNIP is stuck with zero gradient flow. 

\subsection{Real runtime measurements}\label{subsec:real_runtime}
\begin{figure}[tb!]
	\centering
	\begin{subfigure}{.225\textwidth}
		\centering
		\includegraphics[width=\textwidth]{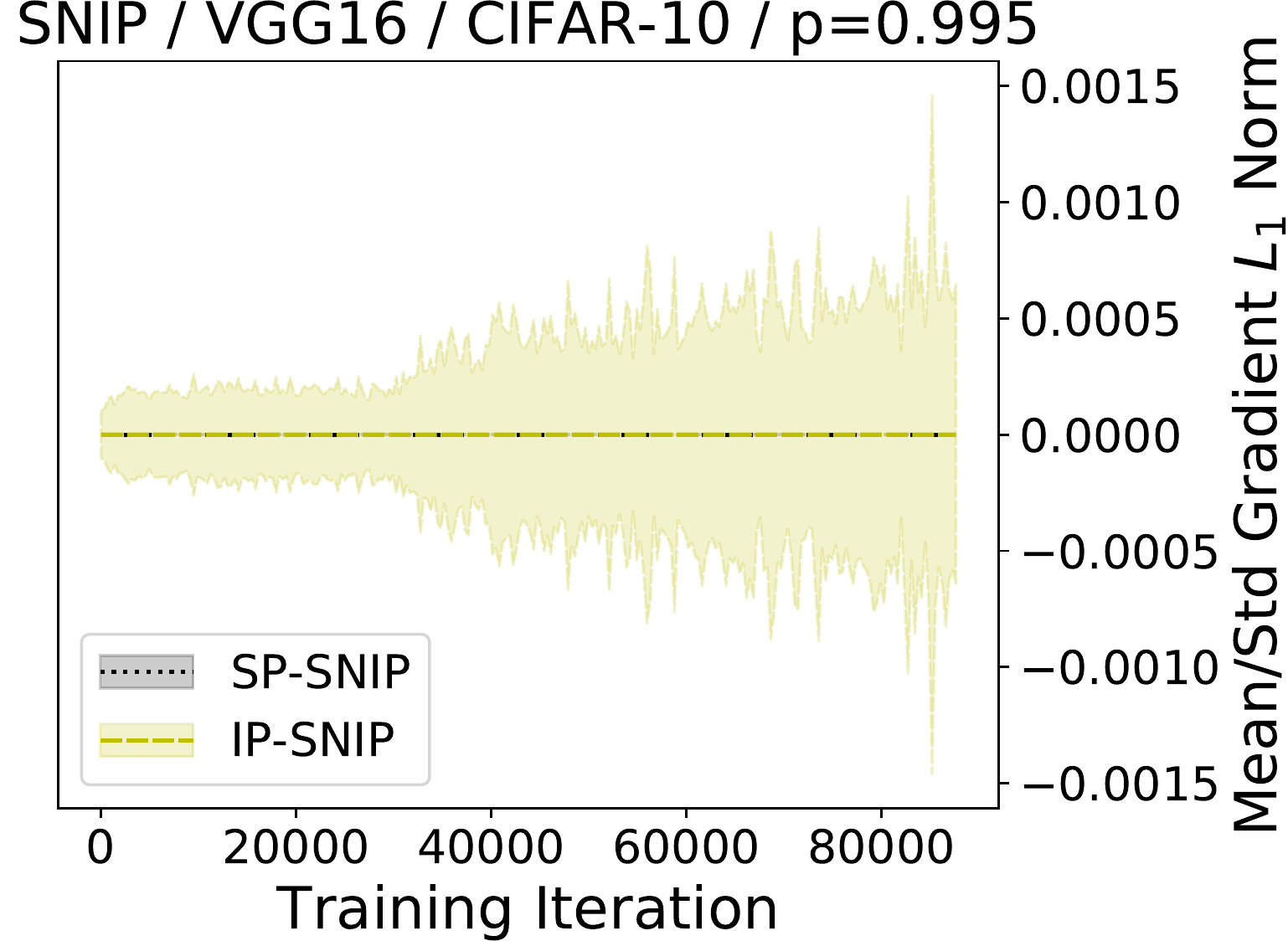}
		\caption{\label{fig:gradflow}}
	\end{subfigure}
	\hfill
	\begin{subfigure}{.225\textwidth}
		\centering
		\includegraphics[width=\textwidth]{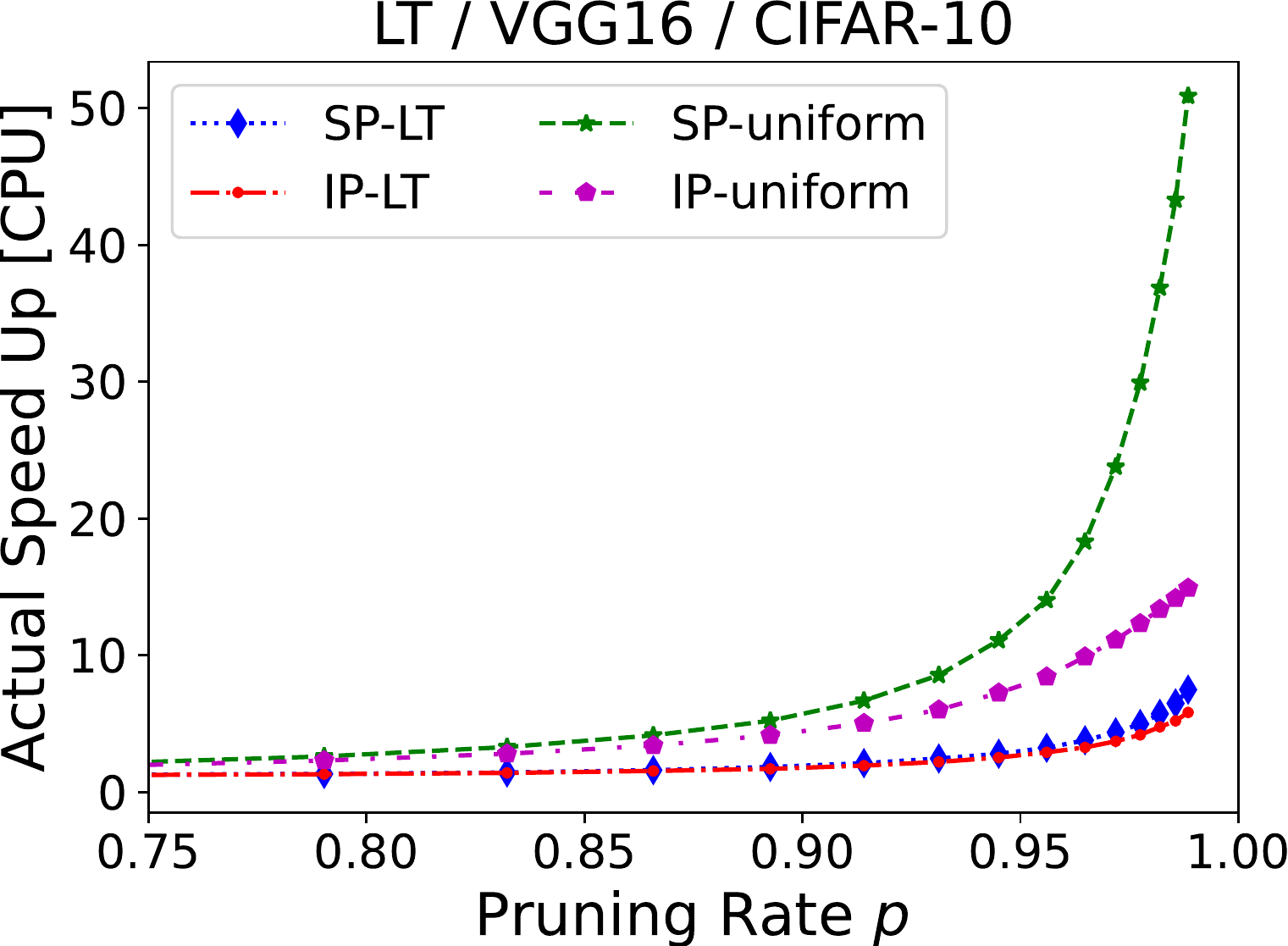}
		\caption{\label{fig:runtime_vs_sparsity}}
	\end{subfigure}
	
	\caption{\label{fig:runtime_and_stability}\subref{fig:gradflow} Mean and std of gradient $L_1$ norm for IP- and SP-SNIP for $p=0.995$. \subref{fig:runtime_vs_sparsity} Speed up on CPU for varying $p$ for SP and IP for uniform sparsity and LT's sparsity.}
\end{figure}
To measure and compare the real runtime accelerations of IP and SP for inference, we used \texttt{scipy}'s \texttt{sparse} package. To be precise, we used \texttt{scipy.sparse.csr\_matrix}, see \href{https://docs.scipy.org/doc/scipy/reference/generated/scipy.sparse.csr_matrix.html}{online documentation}. As discussed in \cref{subsec:computational_costs}, sparse FB convolutions can be computed by first convolving all $X^{(\b)}$ with all $\gn$. Afterwards, sparse matrix multiplications can be used to compute the actual output $Y$. To rule out runtime differences induced by mismatches between sparse implementations of matrix multiplications and convolutions, we simulated sparse convolutions with sparse matrix multiplications of matching dimensions. Therefore, a sparse convolution $h \star X$ with $h \in \R^{\co \times \ci \times K \times K}$ and $X \in \R^{\ci \times H \times W}$ corresponds to a matrix-vector multiplication $\hat{h} \cdot \hat{X}$ with $\hat{h} \in \R^{\co \cdot H \cdot W \times \ci \cdot K^2}$ and $\hat{X} \in \R^{\ci \cdot K^2}$.

We measured the runtime of a VGG$16$ on input images $X \in \R^{3 \times 32 \times 32}$ (\ie CIFAR-$10$) with two different sparsity configurations, the sparsity distribution found by pruning with LT, see \cref{subsec:dst_lth}, and uniform sparsity for each layer. For simplicity, we omit the batch normalization layers and non-linearities. Runtime is measured on one core of an Intel XEON E5-2680 v4 2.4 GHz CPU where we used batch size $1$ and the mean runtime of $25$ runs.  

Figure \cref{fig:runtime_and_stability}\subref{fig:runtime_vs_sparsity} shows the comparison between model sparsity and the actual runtime speed up on a CPU. Since the used CSR \cite{tinney_1967} format for sparse coefficients adds additional overhead to the actually executed computations, runtime is sped up significantly only for $p\geq 0.75$. Note, different sparsity distributions can lead to varying accelerations for a similar global pruning rate $p$. IP indeed has a longer runtime due to the mentioned extra computations. But by boosting performance of sparse models, IP reaches similar results than dense training with $5.2$ times speed up and better results than SP for equal runtime, as shown in \cref{fig:ablation}\subref{fig:c10_lth_mults}.

\section{Transformation rules in the interspace}\label{sec:trafo}
Since the interspace representation is obtained by a linear transformation of the standard, spatial representation, we will derive formulas for this transformation. By knowing them, it will be straight forward to also determine transformation rules for the corresponding gradients and higher derivatives. Those transformation rules might be useful if pruning methods that need first or second order information are used. We test three methods in our work that need the information of the gradient, SNIP \cite{lee_2018}, GraSP \cite{wang_2020} and SynFlow \cite{tanaka_2020}. Moreover, GraSP needs second order information as well. Since all these methods are applied at initialization and we use an initialization equivalent to the standard network in the main part of this work, the gradient and Hessian are equivalent at that time. Still, if such pruning methods are applied with different initializations, the knowledge of these transformation rules might be helpful to overcome scaling problems. Furthermore, we believe the transformation rules to be fruitful for analyzing the information flow in FB-CNNs which we think is an interesting direction for future work.

Again we will assume the special case considered in the paper, \ie $\F$ forming a basis. 

\subsection{Transformation rules for filters}
For a given layer in a CNN, let $\a$ denote the output channel, $\b$ the corresponding input channel and $K \times K$ be the kernel size of a filter $\hab$. For the layer's FB $\F  = \{\g{1}, \ldots, \g{K^2}\}$, the filter's interspace representation is given by
\begin{equation}
\hab = \sum_{n=1}^{K^2} \labn \cdot \g{n} \label{eq:fb_representation_appendix}\;.
\end{equation}
Here, the FB coefficients of $\hab$ are given by $\lab = (\labn)_n \in \R^{K^2}$. Let $\B$ be the standard basis for $\R^{K \times K}$. Then, the spatial representation of filter $\hab$ is given by
\begin{equation}
\hab = \sum_{n=1}^{K^2} \hab_{i_n,j_n} \cdot \en = \sum_{n=1}^{K^2}  \fabn \cdot \en \label{eq:spatial_representation_appendix} \;,
\end{equation}
with spatial coefficients $\fab := (\fabn)_n = ( \langle \hab, \en \rangle )_n \in \R^{K^2}$ and standard basis $\B$ given by $\B = \{e^{(1)}, \ldots, e^{(K^2)}\}$ and 
\begin{equation}
\begin{aligned}
e^{(n)}_{i,j} = \delta_{i,i_n} \cdot \delta_{j, j_n}  \; , \; &(i_n,j_n) \in \{1,\ldots, K\}^2\;, \\ \; &(i_n,j_n) \neq (i_m, j_m) \; \text{for} \; n\neq m\;.
\end{aligned}
\end{equation}
Consequently, 
\begin{equation}
\fab = \Psi \cdot \lab \;,\; \Psi = \left ( \langle \g{m}, \en \rangle \right )_{n,m} \in \R^{K^2 \times K^2}   \label{eq:trafo_spatial}
\end{equation}
holds. Note, since FBs are shared for at least one layer, the basis transformation matrix $\Psi$ is not labeled with the input- and output channels $\a$ and $\b$, respectively. But of course, formulas can be adapted to the case of more than one FB per layer. Since we assume $\F$ to form a basis, the reverse is given by
\begin{equation}
\lab = \Psi^{-1} \cdot \fab \label{eq:trafo_fb}\;.   
\end{equation}
Note, if we use $\F$ as a general dictionary, and not a basis anymore, a reverse can still be computed by the Moore-Penrose pseudo inverse $\Psi^\dagger = \Psi^T\cdot (\Psi \cdot \Psi^T)^{-1}$ if $\F$ forms a \emph{generating system} for $\R^{K \times K}$. If $\F$ forms a linear independent, undercomplete dictionary, we can express $\lab = \hat{\Psi} \fab$ for a suitable $\hat{\Psi} \in \R^{\# \F \times K^2}$. A reverse of this is given by $\fab = \hat{\Psi}^\dagger \lab$, where again $\hat{\Psi}^\dagger = \hat{\Psi}^T\cdot (\hat{\Psi} \cdot \hat{\Psi}^T)^{-1}$ forms the Moore-Penrose pseudo inverse. 

\subsection{Transformation rules for gradients}
Let $\loss$ denote the loss function used to train the CNN. Assuming $\hab$ to be given in the interspace representation \cref{eq:fb_representation_appendix},
\begin{equation}
\frac{\partial \loss}{\partial \lab} = \left ( \frac{\partial \fab}{\partial \lab}\right )^T \cdot \frac{\partial \loss}{\partial \fab}  = \Psi^T \cdot \frac{\partial \loss}{\partial \fab} \label{eq:trafo_fb_grad}
\end{equation}
holds by the chain rule and \cref{eq:trafo_spatial}. Consequently,
\begin{equation}
\frac{\partial \loss}{\partial \fab} =  (\Psi^{-1})^T \cdot \frac{\partial \loss }{\partial \lab} \label{eq:trafo_spatial_grad}
\end{equation}
is true for the gradient as well. 

By comparing \cref{eq:trafo_fb} with \cref{eq:trafo_fb_grad} (or \cref{eq:trafo_spatial} with \cref{eq:trafo_spatial_grad}) we see that the coefficients and their corresponding gradients transform complementary to each other if $\Psi^T \neq \Psi^{-1}$. Note, $\Psi^{-1} = \Psi^T$ if and only if $\Psi$ is orthogonal which is equivalent to $\F$ forming an orthonormal basis (ONB). 

\subsection{Transformation rules for Hessian}
In order to compute the Hessian of the loss function, we have to index all possible filters in a CNN. Let $\labl$ and $\fabl$ denote the interspace and spatial coefficients of a filter in layer $l$ corresponding to input channel $\b$ and output channel $\a$. Here, the basis transformation in layer $l$ is given by $\Psi^{(l)}$. If a FB is shared for layers $l$ and $l+1$, then $\Psi^{(l)} = \Psi^{(l+1)}$ would hold. Furthermore, $\kl$ is the filter size in layer $l$ and $\col$ and $\cil$ denote the number of output and input channels, respectively. We assume the CNN to have $L_c$ convolutional layers in total. Let $H^\B$ be the Hessian matrix of $\loss$ \wrt to coefficients of $\B$. The corresponding values of the Hessian are given by
\begin{equation}
H^\B((\a,\b,n;l),(\a^\prime,\b^\prime,n^\prime;l^\prime) ) = \frac{\partial^2 \loss}{\partial \fabln \partial \fpabln} \;.
\end{equation}
Equivalently, the Hessian \wrt $\F$ is given by $H^\F$ with values
\begin{equation}
H^\F((\a,\b,n;l),(\a^\prime,\b^\prime,n^\prime;l^\prime) ) = \frac{\partial^2 \loss}{\partial \labln \partial \lpabln} \;.
\end{equation}
Using multi-index notation, we can describe the transformation of the Hessian matrix compactly. For $\m:=(\a,\b,n;l)$ and $\mp := (\a^\prime,\b^\prime,n^\prime;l^\prime)$, define 
\begin{align}
\fm  &:= \fabln \;,\; \f  :=  (\fm)_{\m\in\mathcal{M}}  \label{eq:mult_f} \\
\lm &:= \labln  \;,\; \l :=  (\lm)_{\m\in\mathcal{M}} \label{eq:mult_l} \\
\Psi_{\m, \mp} & :=  \delta_{\a, \a^\prime} \cdot \delta_{\b, \b^\prime} \cdot \delta_{l, l^\prime} \cdot \Psi^{(l)}_{n, n^\prime}\;,\\ \Psi & :=   (\Psi_{\m,\mp})_{\m,\mp \in\mathcal{M}} \label{eq:mult_phi}\;,
\end{align}
where all possible multi-indices are given by
\begin{equation}
\mathcal{M} := \bigcup_{l=1}^{L_c} \bigcup_{\a = 1}^{\co^{(l)}} \bigcup_{\b=1}^{\ci^{(l)}} \bigcup_{n=1}^{K_{(l)}^2} \{(\a,\b,n;l) \}\;.
\end{equation}
Let $d := \# \mathcal{M}$ be the dimension of the CNN. For all matrices/vectors $A \in \R^{d_1 \times d}$ and $B \in \R^{d\times d_2}$ with $d_1, d_2 \in \{1, d\}$, indexed with multi-indices, we define multi-index multiplication via
\begin{equation}
(A\cdot B)_{\m, \mp} := \sum_{\md \in \mathcal{M}} A_{\m, \md} \cdot B_{\md, \mp}\;.
\end{equation}
Using the multi-index notation, together with \cref{eq:mult_f,eq:mult_l,eq:mult_phi}, leads to simple transformations of coefficients and their gradients for the whole CNN, given by
\begin{equation}
\f = \Psi \cdot \l \;, \qquad \frac{\partial \loss}{\partial \f} = (\Psi^{-1})^T \cdot \frac{\partial \loss}{\partial \l}\;
\end{equation}
with the transpose of a multi-index matrix $A$ defined via
\begin{equation}
A^T_{\m,\mp} := A_{\mp, \m}\;
\end{equation}
and
\begin{align}
\Psi^{-1} & :=   (\Psi^{-1}_{\m,\mp})_{\m,\mp \in\mathcal{M}}\; \\  \Psi^{-1}_{\m, \mp} & :=  \delta_{\a, \a^\prime} \cdot \delta_{\b, \b^\prime} \cdot \delta_{l, l^\prime} \cdot (\Psi^{(l)})^{-1}_{n, n^\prime}\; .
\end{align}
It holds
\begin{align}
H^\B_{\m, \mp} &  =  \frac{\partial^2 \loss}{\partial \fm \partial \fpm} \\
& =  \frac{\partial}{\partial \fm} \left ( \sum_{\md \in \mathcal{M}} \Psi^{-1}_{\md, \mp} \cdot \frac{\partial \loss}{\partial \l_{\md}}\right) \\
& =  \sum_{\md, \mt \in \mathcal{M}} \Psi^{-1}_{\mt, \m} \cdot H^\F_{\mt, \md} \cdot \Psi^{-1}_{\md, \mp}  \\
& =  \left ( (\Psi^{-1})^T \cdot H^\F \cdot \Psi^{-1} \right )_{\m, \mp}\;.
\end{align}

\section{Computation of pruning scores}\label{subsec:pruning_scores}
In the following, we will derive the computations of the pruning scores used in the experimental evaluation in \cref{sec:experiments} in the main paper. We will present the original scores for SP and their corresponding IP version. In this Section, we assume all FBs $\F$ to form bases and fully connected layers to be described by $1 \times 1$  convolutions. 

\subsection{Pruning scores in general}
In Secs. \ref{subsec:random_score} - \ref{subsec:grasp_score}, five different methods for computing a pruning score vector $S \in \R^d$ are presented. These are the pruning scores used in our experimental evaluation. 

A global pruning score means that the whole network is pruned altogether based on this score vector. Here, $d$ denotes the number of all prunable parameters -- for simplicity pooled in a big vector $\l \in \R^d$. For each parameter $\l_j$, there exists exactly one corresponding pruning score $S_j$. The higher a pruning score, the more important the corresponding parameter is. Thus, for pruning a network with prunable parameters $\l$ to pruning rate $p \in [0,1]$, only the $k$ biggest entries in $S$ are not pruned, where
\begin{equation}
k := \lfloor (1 - p) \cdot d  \rfloor\;.
\end{equation}
Consequently, the global pruning mask $\mu \in \{0,1\}^d$ is defined via
\begin{equation}
\mu_j = \begin{cases} 1 \;, & S_j \; \text{belongs to the} \; k \; \text{biggest entries in} \; S   \\ 0 \;, & \; \text{else} \end{cases} \;.
\end{equation}
After the pruning mask is computed, the network's prunable parameters are masked with the pruning mask via $\l \odot \mu$.

Note, for DST methods, we use \emph{layerwise} pruning. The pruning score is computed equivalent to the global case but each layer is pruned with an individual pruning rate. Consequently, the number of pruned parameters is computed for each layer individually.

\subsection{Random pruning score}\label{subsec:random_score}
Random pruning scores are easy to obtain. For each coefficient $\fabln$ in the SP case or $\labln$ in the IP case, a corresponding random number is drawn \iid from a $\mathcal{N}(0,1)$ distribution.

\subsection{Magnitude pruning score}\label{subsec:magnitude_score}
Magnitudes are used as pruning criterion for LTs \cite{frankle_2020a}, the DST methods SET \cite{mocanu_2018} and RigL \cite{evci_2020}, FT \cite{renda_2020} and GMP \cite{gale_2019}. Using magnitudes as pruning criterion assumes that big coefficients are more likely to significantly influence the network's output than smaller ones. The corresponding formula is straight forward and given by
\begin{equation}
S_{Mag}(\a,\b, n ;l, \B) := \left \vert \fabln \right \vert \;. \label{eq:magnitude_standard}
\end{equation}
The corresponding formula for the FB representation is given by
\begin{equation}
S_{Mag}(\a,\b, n ;l, \F) := \left \vert \labln \right \vert \;. \label{eq:magnitude_fb}
\end{equation}
By the transformation rules for the coefficients \cref{eq:trafo_spatial}, it holds
\begin{equation}
\left \Vert \fabl \right \Vert = \left \Vert \Psi^{(l)} \labl \right \Vert \;. \label{eq:magnitude_comparison}
\end{equation}
Consequently, $K \times K$ filters $\habl$ do not need to have the same total pruning score in the spatial and FB representation, but might be scaled differently. We used magnitude pruning during or after training, where usually $\fabl \neq \labl$. Even though magnitude pruning is normally used for the spatial representation, we did not have any scaling issues in the IP setting. This again indicates that jointly optimizing the FBs and their coefficients is stable.

\subsection{SynFlow pruning score}\label{subsec:synflow_score}
SynFlow \cite{tanaka_2020} is a pruning score, calculating the contribution of a parameter to the CNN's overall information flow. This is done, by differentiating the so called $L_1$ \emph{path norm} of the network. The formula is given by
\begin{equation}
S_{SynFlow}(\a,\b, n ;l, \B) := \frac{\partial \mathcal{R} 
}{\partial \fabln} \cdot \fabln \;,\label{eq:synflow_standard}
\end{equation}
with 
\begin{equation}
\mathcal{R} := \sum_{p \in \mathcal{P}} \prod_{\fabln \in p} \vert \fabln \vert \;.\label{eq:synflow_path_norm}
\end{equation}
Here, 
\begin{equation}
\mathcal{P} = \{\{\varphi_{n_1}^{(\a_1,\b;1)}, \varphi_{n_2}^{(\a_2,\a_1;2)}, \ldots, \varphi_{n_{L_c}}^{(\a_{L_c},\a_{L_{c-1}};L_c)}\}\}
\end{equation}
describes all existing paths in a CNN which start in the input layer and end in the output layer.

Consequently, the corresponding SynFlow score \wrt coefficients for the FBs is given by
\begin{equation}
S_{SynFlow}(\a,\b, n ;l,\F) := \frac{\partial \mathcal{R} 
}{\partial \labln} \cdot \labln \;.\label{eq:synflow_fb}
\end{equation}
The basis transformation between $\B$ and $\F$ does not change the total pruning score of a filter $\habl$. This can be seen by
\begin{align}
& \sum_{n=1}^{K^2} S_{SynFlow}(\a,\b, n;l,\B) \\ = & \left \langle \frac{\partial \mathcal{R} 
}{\partial \fabl}, \fabl \right \rangle \\
=  &\left \langle \left ( {\Psi^{(l)}}^{-1} \right )^T \frac{\partial \mathcal{R} 
}{\partial \labl}, \Psi^{(l)} \labl \right \rangle \\
= & \left \langle \frac{\partial \mathcal{R} }{\partial \labl}, {\Psi^{(l)}}^{-1} \Psi^{(l)} \labl \right \rangle \\
=  &\left \langle \frac{\partial \mathcal{R}}{\partial \labl}, \labl \right \rangle \\
=  &\sum_{n=1}^{K^2} S_{SynFlow}(\a,\b, n;l,\F)
\;.\label{eq:}
\end{align}
The second equality is induced by the transformation formulas \cref{eq:trafo_spatial,eq:trafo_spatial_grad}. By having the same total pruning score for a filter for coefficients \wrt $\F$ and $\B$, we do not need to worry about possible scaling issues for the SynFlow score.

\subsection{SNIP pruning score}\label{subsec:snip_score}
SNIP \cite{lee_2018} computes a so called \emph{saliency score} for each parameter of a CNN before training. The idea is to measure the effect of changing the activation of a coefficient on the loss function. If this effect is big, the corresponding coefficient is trained, otherwise it is pruned. Let $\fabl \in \R^{\kl^2}$ be the vector consisting of all spatial coefficients in the $l$-th layer of a CNN with input channel $\b$ and output channel $\a$. Its saliency score is then computed as 
\begin{align}
S_{SNIP}(\a,\b,n;l,\B)  :=& \left \vert  \left. \frac{\partial \loss (m \cdot \fabln)}{\partial m} \right \vert_{m=1} \right \vert \\ = & \left \vert \frac{\partial \loss}{\partial \fabln} \cdot \fabln \right \vert \;,\label{eq:snip_standard}
\end{align}
where $m\in \R$ models the activation of the filter value and $\loss$ is the used loss function. The second equality is induced by using the chain rule \cite{wang_2020}. 

The corresponding SNIP score w.r.t. $\F$ is given by 
\begin{align}
S_{SNIP}(\a,\b,n;l,\F) :=  &\left \vert \left. \frac{\partial \loss (m\cdot \labln)}{\partial m} \right \vert_{m=1} \right \vert \\  = & \left \vert \frac{\partial \loss }{\partial \labln} \cdot \labln \right \vert \;.\label{eq:snip_fbpruning}
\end{align}
By inserting the transformation formulas \eqref{eq:trafo_spatial} and \eqref{eq:trafo_spatial_grad} into \cref{eq:snip_standard}, we get the relationship for the SNIP score of a filter $\habl$ as
\begin{align}
& S_{SNIP}(\a,\b;l,\B) \\ :=  & (S_{SNIP}(\a,\b,n;l,\B))_{n=1}^{K^2} \\  = &  \left \vert ({\Psi^{(l)}}^{-1})^T \frac{\partial \loss}{\partial \labl} \right \vert \odot \left \vert \Psi^{(l)} \labl \right \vert \;.\label{eq:snip_general_comparison}
\end{align}
By comparing \cref{eq:snip_general_comparison,eq:snip_fbpruning}, we see that changing the basis from $\B$ to $\F$ leads to different transformations of the gradient and the basis coefficient for a non-orthonormal FB $\F$. 

In our experiments in the main body of the work, we computed the SNIP score with $\F = \B$, thus spatial and FB SNIP scores are equivalent. But, if arbitrary FBs are used, the scaling \cref{eq:snip_general_comparison} might cause problems and should be taken into account. 

\subsection{GraSP pruning score}\label{subsec:grasp_score}
The GraSP score \cite{wang_2020} approximates the influence of the removal of a spatial coefficient onto the network's gradient flow before training starts, the so called \emph{importance score}. Using multi-index notation, it is computed as
\begin{equation}
S_{GraSP}(\m; \B) := - \left (H^\B \cdot \frac{\partial \loss}{\partial \varphi} \right )_{\m} \cdot \varphi_{\m} \;.\label{eq:grasp_standard}
\end{equation}
The corresponding score w.r.t. to basis coefficients $\F$ is given by
\begin{equation}
S_{GraSP}(\m; \F) := - \left (H^\F \cdot \frac{\partial \loss}{\partial \l} \right )_{\m} \cdot \l_{\m} \;.\label{eq:grasp_fb}
\end{equation}
By inserting the transformation rules for the coefficient, gradient and Hessian matrix, we derive
\begin{align}
&- \left (H^\B \cdot \frac{\partial \loss}{\partial \varphi} \right )_{\m}\cdot \fm \\
= &- \left ((\Psi^{-1})^T \cdot H^\F \cdot \Psi^{-1} \cdot  (\Psi^{-1})^T \cdot \frac{\partial \loss}{\partial \l} \right )_{\m} \cdot (\Psi \cdot \lambda)_{\m} \;.\label{eq:grasp_trafo}
\end{align}
Therefore, the GraSP score is scaled differently for varying layers. Similar to the SNIP score, scaling issues might needed to be handled if FBs do not form ONBs. 

On the other hand, if all FBs form ONBs, \cref{eq:grasp_trafo} reduces to 
\begin{equation}
S_{GraSP}(\m; \B) = - \left (\Psi \cdot H^\F \cdot \frac{\partial \loss}{\partial \l} \right )_{\m} \cdot (\Psi \cdot \lambda)_{\m} .
\end{equation}
Similar to SynFlow, it therefore holds
\begin{equation}
\sum_{{\m} \in \mathcal{M}_{\a,\b;l}} S_{GraSP}({\m}; \B) = \sum_{ {\m}\in \mathcal{M}_{\a,\b;l}} S_{GraSP}({\m}; \F)
\end{equation}
for $\mathcal{M}_{\a,\b;l} := \{(\a, \b, n;l) : n = 1,\ldots, K_{(l)}^2 \}$, the multi-indices corresponding to an arbitrary filter $\habl$. In this case, the total pruning score of a filter $\habl$ does not depend on the representation.

\section{Pruning methods and initialization of the interspace}\label{sec:init_fbs}
\begin{algorithm}[tb!]
	\caption{{Standard} interspace initialization for a FB 2D convolutional layer}
	\label{alg:standard_init}
	\begin{algorithmic}[1]
		\Require{Filter size $K \times K$, number of output channels $\co$, number of input channels $\ci$}
		\Statex
		\State $\mu_h \leftarrow 0$ mean of spatial coefficients
		\State $\s_h \leftarrow \sqrt{\frac{2}{\ci \cdot K^2}}$ variance of spatial coefficients 
		\State Initialize $g^{(1)}, \ldots, g^{(K^2)}$ via $\gn = \en$ for all $n = 1, \ldots, K^2$
		\State Initialize $\labn \sim \mathcal{N}(\mu_h, \s_h^2)$ \iid for all $\a \in \{1, \ldots, \co\}, \b \in \{1, \ldots, \ci\}, n \in \{1, \ldots, K^2\}$
		\State \Return FB $\F = \{g^{(1)}, \ldots, g^{(K^2)}\}$, FB coefficients $\l = (\labn)_{\a,\b,n}$
	\end{algorithmic}
\end{algorithm}
\begin{algorithm}[tb!]
	\caption{Random \texttt{ONB} interspace initialization for a FB 2D convolutional layer}
	\label{alg:onb_init}
	\begin{algorithmic}[1]
		\Require{Filter size $K \times K$, number of output channels $\co$, number of input channels $\ci$}
		\Statex
		\State $\mu_h \leftarrow 0$ mean of spatial coefficients 
		\State $\s_h \leftarrow \sqrt{\frac{2}{\ci \cdot K^2}}$ variance of spatial coefficients 
		\State Initialize $\tilde{g}^{(1)}, \ldots, \tilde{g}^{(K^2)} \in \R^{K\times K}$ with $\tilde{g}^{(n)}_{i,j} \sim \mathcal{N}(0,1)$ \iid  \Comment{\parbox[t]{.6\linewidth}{$\{\tilde{g}^{(1)}, \ldots, \tilde{g}^{(K^2)}\}$ with $\mathbb{P}=1$ lin. independent}}
		\State Apply Gram-Schmidt on $\{\tilde{g}^{(1)}, \ldots, \tilde{g}^{(K^2)}\}$ to obtain ONB $\F = \{\g{1}, \ldots, \g{K^2}\}$
		\State Initialize spatial coefficients $\fabn \sim \mathcal{N}(\mu_h, \s_h^2)$ \iid for all $\a \in \{1, \ldots, \co\}, \b \in \{1, \ldots, \ci\}, n \in \{1, \ldots, K^2\}$
		\State Compute basis transformation matrix $\Psi$ according to \cref{eq:trafo_spatial}
		\State $\lab \leftarrow \Psi^T \cdot \fab$ FB coefficients
		\State \Return FB $\F = \{\g{1}, \ldots, \g{K^2}\}$, FB coefficients $\l = (\labn)_{\a,\b,n}$
	\end{algorithmic}
\end{algorithm}
\begin{algorithm}[tb!]
	\caption{\texttt{Random} interspace initialization for a FD 2D convolutional layer with $\# \F = N$ arbitrary}
	\label{alg:rnd_init}
	\begin{algorithmic}[1]
		\Require{Size of filter dictionary $N$, filter size $K \times K$, number of output channels $\co$, number of input channels $\ci$}
		\Statex
		\State $\mu_h \leftarrow 0$ mean of spatial coefficients
		\State $\s_h \leftarrow \sqrt{\frac{2}{\ci \cdot K^2}}$ variance of spatial coefficients 
		\State Initialize $\tilde{g}^{(1)}, \ldots, \tilde{g}^{(N)} \in \R^{K\times K}$ with $\tilde{g}^{(n)}_{i,j} \sim \mathcal{N}(0,1)$ \iid \Comment{\parbox[t]{.8\linewidth}{$\{\tilde{g}^{(i_1)}, \ldots, \tilde{g}^{(i_m)}\}$ with $i_1 \neq \ldots \neq i_m$ and $m \leq K^2$ with $\mathbb{P}=1$ lin. independent}}
		\State Compute pixelwise sample mean $\tilde{\mu}_{i,j}$ and sample variance $\tilde{\s}_{i,j}$ according to \cref{eq:mean_variance_eyes}
		\State $\g{n}_{i,j} \leftarrow \sqrt{\frac{1}{N} - \frac{1}{N^2}} \cdot \frac{\tilde{g}^{(n)}_{i,j} - \tilde{\mu}_{i,j}}{\tilde{\s}_{i,j}} + \frac{1}{N}$ \Comment{rescale FD} 
		\State Initialize $\labn \sim \mathcal{N}(\mu_h, \s_h^2)$ \iid for all $\a \in \{1, \ldots, \co\}, \b \in \{1, \ldots, \ci\}, n \in \{1, \ldots, N\}$
		\State \Return FD $\F = \{\g{1}, \ldots, \g{N}\}$, FD coefficients $\l = (\labn)_{\a,\b,n}$
	\end{algorithmic}
\end{algorithm}
In our experiments, we used the so called \texttt{kaiming normal} initialization \cite{He2015} for the standard CNNs. Meaning that $\hab_{i,j} \sim \mathcal{N}(\mu_h,\s_h^2)$ \iid with
\begin{equation}\label{eq:kaiming}
\mu_h = 0 \; \text{and} \; \s_h = \sqrt{\frac{2}{\ci \cdot K^2}} \;.
\end{equation}
We initialized all FB-CNNs such that their \emph{spatial representations} follow a \texttt{kaiming normal} initialization, see Algs. \ref{alg:standard_init} - \ref{alg:rnd_init}. For simplicity, we propose the initialization of FB coefficients together with the FB. Of course, if a FB is shared for more than one layer, it has to be initialized just once.

\paragraph{Derivation of rescaling in Algorithm \ref{alg:rnd_init}.}
In \cref{alg:rnd_init}, $\F$ may contain $N \neq K^2$ elements. Thus, obtaining an equivalent initialization to the spatial \texttt{kaiming normal} initialization can not always be obtained by a simple basis transformation. Consequently, we rescale $\F$ in order to mimic a \texttt{kaiming normal} initialization of spatial coefficients if FB coefficients are initialized with a \texttt{kaiming normal} initialization as well.

Let $\mu_{i,j}$ and $\sigma^2_{i,j}$ be the pixel wise sample mean and sample variance of the FD $\F$ with arbitrary size $N \geq 1$, \ie
\begin{equation}
\mu_{i,j} := \frac{1}{N} \sum_{n=1}^{N} \g{n}_{i,j}  \; \text{and} \; \s_{i,j}^2 := \frac{1}{N} \sum_{n=1}^{N} (\g{n}_{i,j} - \mu_{i,j})^2 \;. \label{eq:mean_variance_eyes}
\end{equation}
By using \cref{eq:mean_variance_eyes} it holds for an arbitrary \iid initialization of $\l$ with mean $\mu_\l$ and variance $\sigma_\l^2$ 
\begin{align}
\mathbb{E} [h_{i,j}] = \mathbb{E} [\sum_{n=1}^N \l_n \g{n}_{i,j}] = \sum_{n=1}^N \mathbb{E} [\l_n] \g{n}_{i,j} = \mu_\l  N  \mu_{i,j} \label{eq:mean_fb_init} 
\end{align}
and
\begin{align}
\mathbb{E}[h_{i,j}^2] & = \sum_{n,m} \mathbb{E} [\l_n \l_m] \g{n}_{i,j} \g{m}_{i,j}  \\ &= \sum_{n} \mathbb{E} [\l_n^2]  {\g{n}_{i,j}}^2 + \sum_{n} \sum_{n\neq m} \mathbb{E} [\l_n]^2  \g{n}_{i,j}  \g{m}_{i,j} \\
& = (\s_\l^2 + \mu_\l^2) \sum_{n} {\g{n}_{i,j}}^2 +  \mu_\l^2 \sum_{n} \sum_{n\neq m} \g{n}_{i,j} \g{m}_{i,j} \\
& = \s_\l^2 \sum_{n} {\g{n}_{i,j}}^2 +  \mu_\l^2 \sum_{n, m} \g{n}_{i,j} \g{m}_{i,j} \\ 
& = N \s_\l^2 (\s_{i,j}^2 + \mu_{i,j}^2 ) + \mu_\l^2 N^2 \mu_{i,j}^2 \label{eq:variance_fb_init}\;.
\end{align}
We now want to determine $\mu_{i,j}$ and $\s_{i,j}$ such that $\mathbb{E}[h_{i,j}] = \mu_\l$ and $\mathbb{E}[h_{i,j}^2] = \s_\l^2 + \mu_\l^2$ holds, \ie the distribution of $\l$ and $h$ have the same mean and variance. By \cref{eq:mean_fb_init}, setting $\mu_{i,j} = \nicefrac{1}{N}$ guarantees $\mathbb{E}[h_{i,j}] = \mu_\l$. By inserting $\mu_{i,j} = \nicefrac{1}{N}$ into \cref{eq:variance_fb_init}, we see that $\sigma_{i,j}^2 = \nicefrac{1}{N} - \nicefrac{1}{N^2}$ implies $\mathbb{E}[h_{i,j}^2] = \s_\l^2 + \mu_\l^2$. Consequently, FDs are rescaled in \cref{alg:rnd_init} to have pixelwise sample mean $\mu_{i,j} = \nicefrac{1}{N}$ and sample variance $\s^2_{i,j} = \nicefrac{1}{N} - \nicefrac{1}{N^2}$. 

\subsection{General setup for all pruning methods}
For SP and the dense baselines, we initialized the standard CNN with the \texttt{kaiming normal} initialization. For IP, we initialize networks according to \cref{alg:standard_init} for all experiments in the main body of the text. In \cref{sec:additional_ablations}, also \cref{alg:onb_init,alg:rnd_init} are used as initializations for the interspace.

Pruning masks are computed according to the formulas described in \cref{subsec:pruning_scores} for SP and IP. In the following we will describe the used pruning methods in detail.

\subsection{Lottery tickets with resetting coefficients}
LTs with resetting coefficients to an early training iteration \cite{frankle_2020a} are obtained as follows. We first train the network to epoch $t_0 = 500$ and store the corresponding model, optimizer, \etc. Afterwards, the network is trained to convergence. Then, $20 \%$ of the coefficients are pruned, based on their magnitudes. All non-zero parameters are reset to their values at training time $t_0=500$ and pruned ones are fixed at zero from now on. Thus, contrarily to DST, a pruned coefficient will never be able to recover. The training schedule parameters, like learning rate or moving averages for batch normalization and SGD with momentum, are reset to their corresponding value at $t_0$ as well. For IP, also the FBs are reset to step $t_0$. Then, the network is again trained to convergence, $20 \%$ of the non-zero coefficients are pruned and the remaining non-zero parameters are reset again. This is done, until the desired pruning rate is reached. Note, for the last pruning step also $< 20\%$ of the non-zero parameters might be pruned to exactly match the desired pruning rate. If the final pruning rates is reached, the network with desired sparsity is trained for a last, final time. All in all, 
\begin{equation}
k = 1 + \left \lceil \frac{\log (1 - p)}{\log 4 - \log 5} \right \rceil
\end{equation}
trainings are needed to obtain and train a network with sparsity $p$ using this iterative approach.

Following \cite{frankle_2020a}, we do not prune the fully connected layer for LTs but keep it dense.

\subsection{Dynamic sparse training}
Dynamic sparse training methods adapt the network's pruning mask during training \cite{bellec_2018,dettmers_2019,evci_2020,liu_2021b,mocanu_2018,mostafa_2019}. In this work we use SET \cite{mocanu_2018} which is based on estimating the importance of coefficients via magnitude pruning and regrowing coefficients due to a random selection. RigL \cite{evci_2020} improves this approach by regrowing coefficients which have the biggest gradient magnitudes. 

Before training, the networks are pruned randomly. It was shown in \cite{evci_2020} that using layerwise sparsity corresponding to an Erd\H{o}s-R\'enyi-kernel leads to good results. This means that each layer has sparsity depending on its size, \ie
\begin{equation}
1 - \varepsilon \cdot \frac{\co^{(l)} + \ci^{(l)} + 2 \cdot K_{(l)}}{\co^{(l)} \cdot \ci^{(l)} \cdot K^2_{(l)}}\;,
\end{equation}
and $\varepsilon$ is a global parameter, tuned such that a global sparsity of $p$ is obtained.

During training, the pruning mask is frequently updated. For this, parameters are pruned for each layer with rate $p_t$. To be precise, all \emph{non-zero} parameters in a layer are pruned with the rate $p_t$. The pruning rate $p_t$ depends on the training step $t$ and decays with a cosine schedule in order to improve convergence \cite{evci_2020}. It holds
\begin{equation}
p_t = p_{min} + \frac{1}{2} \left ( p_{init} - p_{min} \right) \cdot \left ( 1 + \cos \left( \frac{t \pi}{T} \right) \right )
\end{equation}
where $T$ is the number of total training steps, $p_{min} = 0.005$ the minimal pruning rate and $p_{init} = 0.5$ the initial rate used for updating the pruning mask. Of course, in each layer an equal number of non trained coefficients are regrown after pruning. Following \cite{mocanu_2018} and \cite{evci_2020}, regrown coefficients are initialized with value $0$ but are updated via SGD from this moment on.

SET and RigL work optimal for different pruning mask update frequencies \cite{liu_2021b}. According to \cite{liu_2021b}, we update pruning masks each $1,500$ training steps for SET and each $4,000$ steps for RigL.

\subsection{Pruning at initialization}
We test PaI methods, SNIP \cite{lee_2018}, GraSP \cite{wang_2020} and SynFlow \cite{tanaka_2020} together with random PaI. In contrast to LTs and DST, PaI is quite simple. For SNIP and GraSP we compute the pruning scores described in \cref{subsec:snip_score,subsec:grasp_score} with the help of $100$ batches of training data for the CIFAR-$10$ experiments and $15$ for ImageNet. As proposed by \cite{tanaka_2020}, we compute the pruning scores for SNIP and GraSP with all batch normalization layers \cite{ioffe_2015} set to PyTorch's \texttt{train} mode. Afterwards, all coefficients are pruned one-shot. 

The GraSP scores for SP, \cref{eq:grasp_standard}, and IP, \cref{eq:grasp_fb}, require the computation of a Hessian vector product $H \cdot g$. Fortunately, not the whole Hessian $H$ needs to be computed to evaluate such products. For this, we use the linearity of the derivative. For $H \in \R^{d \times d}$ and an arbitrary vector $v \in \R^d$, it holds
\begin{align}
(H\cdot v)_i & = \sum_{j} \frac{\partial^2 \loss}{\partial \lambda_i \partial \lambda_j} \cdot v_j \\
& = \frac{\partial}{\partial \lambda_i} \left ( \sum_{j} \frac{\partial \loss}{\partial \l_j} v_j \right) \\
& = \frac{\partial}{\partial \lambda_i} \left \langle \frac{\partial \loss}{\partial \l }, v \right \rangle \;,
\end{align}
and consequently $H \cdot v = \frac{\partial \langle \frac{\partial \loss}{\partial \l}, v \rangle}{\partial \l}$. By using $v$ as the fixed gradient $g_v = \frac{\partial \loss}{\partial \l}$, we can compute $H \cdot g_v$ with only three backward passes. The first one is needed to compute the fixed gradient $g_v$. The second and third are required to compute $H \cdot g_v = \frac{\partial \langle \frac{\partial \loss}{\partial \l}, g_v \rangle}{\partial \l}$. An implementation of this can be found in the official code base for GraSP, \href{https://github.com/alecwangcq/GraSP}{see this link} (MIT license).

In contrast to these one-shot methods, pruning masks for SynFlow are computed in an iterative fashion. For this purpose, we compute the pruning score proposed in \cref{subsec:synflow_score} with one forward- and backward pass, prune a small fraction of elements and repeat it $100$ times. The pruning rate grows with an exponential schedule \cite{tanaka_2020} which gives the pruning rate
\begin{equation}
p_k = 1 - (1-p)^{\nicefrac{k}{100}}
\end{equation}
after the $k$\textsuperscript{th} pruning iteration. For computing the SynFlow score, we set all batch normalization layers to \texttt{eval} mode, as suggested by \cite{tanaka_2020}.

\subsection{Gradual magnitude pruning}
Following \cite{gale_2019} we gradually sparsify the model based on the coefficients' magnitudes. After each $N$ training iterations, the pruning rate is increased and new weights are pruned. The pruning rate at iteration $k \cdot N$ is given by
\begin{equation}
p(k\cdot N) = \begin{cases}
0, &  k\cdot N < t_0 \\
p\cdot \left(1 - \left( 1 - \frac{k \cdot N - t_0}{t_1 - t_0} \right)^3 \right) , & k\cdot N \in [t_0, t_1] \\
1, &  k\cdot N > t_1
\end{cases}\;.
\end{equation} 
Here, $p$ denotes the final pruning rate and $t_0, t_1$ are the training iterations where the gradual pruning begins and ends, respectively. For each iteration $k\cdot N$ with $k \in \mathbb{N}$, the $p(k \cdot N) \cdot d$ weights with smallest magnitude are pruned. Since pruned weights are frozen at $0$, those weights will be pruned again and they will therefore never recover.

We follow the suggestions in \cite{gale_2019} for choosing $t_0, t_1$ and $N$ for the final pruning rates $p \in \{0.8, 0.9\}$ for the ResNet$50$ on ImageNet (summarized in \href{https://github.com/google-research/google-research/blob/master/state_of_sparsity/results/sparse_rn50/technique_comparison/rn50_magnitude_pruning.csv}{this table}). Since we use a different batch size than \cite{gale_2019} in our experiments ($256$ compared to $1,024$), we adapt their choices for $t_0, t_1$ and $N$ to our batch size by multiplying them by $4$. However, these choices are not optimized for batch size $256$ and we therefore report slightly worse results than \cite{gale_2019}. For our experiments, we use $N = 8,000, t_0 = 160,000 $ and $t_1 = 400,000$ for $p=0.8$. For $p=0.9$ we set $N = 8,000, t_0 = 160,000 $ and $t_1 = 304,000$.

\subsection{Fine tuning}
We also compare IP and SP for magnitude pruning applied on a pre-trained, dense network while the sparse network is fine-tuned afterwards. As suggested by \cite{renda_2020}, we do not use a classical fine-tuning setup beginning with a small learning rate, but use the setup of the dense pre-training also for fine-tuning. Thus, the sparse fine-tuning starts with a high learning rate. Naturally, for IP we use the pre-trained FBs as starting point for the fine-tuning. For FT, we use the same training setup as for RigL.

\section{Experimental setup}\label{sec:experimental_setup}
\begin{table*}[tb]
	\small
	\centering
	\makebox[\textwidth]{
		\begin{tabular}{@{}lcccccccccc@{}}
			\hline
			{Dataset/Model} & {Mean $\pm$ Std} & {$\#$ Params} & \begin{tabular}[x]{@{}c@{}}$\#$ FB Params\\\texttt{fine} sharing\end{tabular}  & \begin{tabular}[x]{@{}c@{}}Time $1$ \\ epoch [s]\end{tabular} & $\#$GPUs & \begin{tabular}[x]{@{}c@{}}GPU\\memory\end{tabular} & \begin{tabular}[x]{@{}c@{}}$\#$CPU\\cores\end{tabular} & RAM \tabularnewline
			\hline 
			{CIFAR-$10$/VGG$16$} & {$93.41 \pm 0.07\%$} & {$15.3$mio} & $1,053$ & {$23.7$} & $1$ & $11$ GB & $1$ & $12$ GB \tabularnewline
			{CIFAR-$10$/VGG$16$-LT} & {$93.58 \pm 0.10\%$} & {$14.7$mio} & $1,053$ & {$21.9$} & $1$ & $11$ GB & $1$ & $12$ GB\tabularnewline
			{ImageNet/ResNet$18$} & {$69.77 \pm 0.06\%$} & {$11.7$mio}& $1,296$ & {$2,770.5$} & $8$ & $8 \times 11$ GB & $8$ & $8 \times 12$ GB \tabularnewline
			{ImageNet/ResNet$50$ (RigL \& FT)} & {$77.15 \pm 0.04\%$} & {$25.6$mio}& $3,697$ & {$4,354.2$} & $2$ & $2 \times 11$ GB & $4$ & $4 \times 12$ GB \tabularnewline
			{ImageNet/ResNet$50$ (GMP)} & {$76.64 \pm 0.06\%$} & {$25.6$mio}& $3,697$ & {$2,622.8$} & $4$ & $4 \times 11$ GB & $8$ & $8 \times 12$ GB \tabularnewline
			\hline
	\end{tabular}	}
	\caption{Top-$1$ test accuracies for densely trained models with additional information, including the hardware setup. Note, $\#$ FB parameters for \texttt{fine} sharing is the \emph{most} number of extra parameters induced by the interspace representation. We suggest to use at least $2$ times the number of CPU cores than GPUs, otherwise data loading becomes a bottleneck (compare ResNet$18$ and ResNet$50$ for GMP which have approximately the same runtime despite different network sizes and $\#$ GPUs).}
	\label{tab:dense_models}
\end{table*}
\begin{table*}[tb]
	\centering
	\begin{tabular}{@{}lcccccc@{}}
		\hline
		{\small{}Experiment} & {\small{}CIFAR-$10$ (no LTs)} & {\small{}CIFAR-$10$ (LTs)}  
		& {\small{}ImageNet (PaI)} & {\small{}ImageNet (RigL \& FT)} & {\small{}ImageNet (GMP)}\tabularnewline
		\hline 
		{\small{}Network} & {\small{}VGG$16$}  & {\small{}VGG$16$-LT} 
		& {\small{}ResNet$18$} & {\small{}ResNet$50$} & {\small{}ResNet$50$} \tabularnewline
		{\small{}$\#$ Trainings} & {\small{}$5$} & {\small{}$5$}  
		& {\small{}$3$} & {\small{}$3$} & {\small{}$3$} \tabularnewline
		{\small{}$\#$ Epochs} & {\small{}$250$} & {\small{}$160$} 
		& {\small{}$90$} & {\small{}$100$} & {\small{}$100$} \tabularnewline
		{\small{}Batch Size} & {\small{}$128$} & {\small{}$128$}
		& {\small{}$512$} & {\small{}$128$} & {\small{}$256$} \tabularnewline
		{\small{}$\#$ GPUs} & {\small{}$1$}  & {\small{}$1$} 
		& {\small{}$8$} & {\small{}$2$} & {\small{}$4$}\tabularnewline
		{\small{}Optimizer: SGD-} & {\small{}Momentum} & {\small{}Momentum} 
		& {\small{}Momentum} & {\small{}Momentum} & {\small{}Momentum}\tabularnewline
		{\small{}Momentum} & {\small{}$0.9$} & {\small{}$0.9$} & {\small{}$0.9$} 
		& {\small{}$0.9$} & {\small{}$0.9$} \tabularnewline
		{\small{}Learning Rate} & {\small{}$0.1$} & {\small{}$0.1$} & {\small{}$0.1$} 
		& {\small{}$0.1$} & {\small{}$0.1$} \tabularnewline
		{\small{}LR Decay} & {\small{}$\times0.1$}  & {\small{}$\times0.1$}	& {\small{}$\times0.1$} 
		& {\small{}$\times0.1$} & {\small{}$\times0.1$}\tabularnewline
		& {\small{}every $30$k iterations} & {\small{}epochs $80/120$} 
		& {\small{}epochs $30/60$} & {\small{}epochs $30/60/90$}  & {\small{}epochs $30/60/80$}\tabularnewline
		{\small{}LR Warm-up} & {\small{}\xmark} & {\small{}\xmark}
		& {\small{}\xmark} & {\small{}$5$ epochs, linear}  & {\small{}$5$ epochs, linear}\tabularnewline
		{\small{}Weight Decay} & {\small{}$5\cdot10^{-4}$} & {\small{}$10^{-4}$}
		& {\small{}$10^{-4}$} & {\small{}$10^{-4}$} & {\small{}$10^{-4}$}\tabularnewline
		{\small{}Label smoothing} & {\small{}\xmark} & {\small{}\xmark} 
		& {\small{}\xmark} & {\small{}$\varepsilon = 0.1$}  & {\small{}$\varepsilon = 0.1$}\tabularnewline
		
		\hline
	\end{tabular}
	\caption{Setups for experiments
		in the main body of the paper and \cref{sec:additional_ablations}. Each run of the $\#$ trainings was executed with a different random seed.}
	\label{tab:exp_setups}	
\end{table*}
In this Section, we describe the setup for the experiments discussed in the main body of the paper and in \cref{sec:additional_ablations}. 
Our experiments were conducted on an internal cluster with CentOS Linux release 7.9.2009 (Core). We used Python $3.8$ with the deep learning framework PyTorch$1.9$ \cite{pytorch} (BSD license) together with cudatoolkit $10.2$ \cite{cuda}. As hardware we had an Intel XEON E$5$-$2680$ v$4$ à $2.4$ GHz CPU and $n$ NVIDIA GeForce $1080$ti GPUs, where $n=1$ for the CIFAR-$10$ experiments, $n\in \{2, 4\}$ for ResNet$50$ on ImageNet and $n=8$ for ResNet$18$ on ImageNet. Further details on the training time for one epoch, the number of used CPU cores, used RAM and GPU memory are given in \cref{tab:dense_models}.

Since we compare and adapt different pruning methods, we used the publicly available codes for reproducing the results and modifying them for the IP setting. These are:
\begin{itemize}[noitemsep,topsep=0pt]
	\item \href{https://github.com/ganguli-lab/Synaptic-Flow}{Link to code} for SynFlow \cite{tanaka_2020} (unknown license),
	\item \href{https://github.com/alecwangcq/GraSP}{Link to code} for GraSP \cite{wang_2020} (MIT license),
	\item PyTorch adaption of \href{https://github.com/namhoonlee/snip-public}{link to code} for SNIP \cite{lee_2018} (MIT license),
	\item \href{https://github.com/facebookresearch/open_lth}{Link to code} as the official PyTorch version for Lottery Tickets \cite{frankle_2020a} (MIT license),
	\item \href{https://github.com/Shiweiliuiiiiiii/In-Time-Over-Parameterization}{Link to code} as a base for DST methods which is the official code for \cite{liu_2021b} (unknown license). In \cite{liu_2021b}, training schedules for SET \cite{mocanu_2018} and RigL \cite{evci_2020} are improved. This code base is also used for FT \cite{renda_2020} and GMP \cite{gale_2019}. 
\end{itemize}
For the IP versions of these pruning methods, we additionally updated the code to use \cref{alg:fb_conv2d} as a 2D convolution.

The initializations of the used CNNs and FB-CNNs are described in \cref{sec:init_fbs}. 
Used hyperparameters are summarized in \cref{tab:exp_setups}. We also trained the dense standard networks with the same training schedules as the pruned ones. Results for the dense models can be found in \cref{tab:dense_models}.

\paragraph{Training schedule.}
As common in the literature of sparse training, see for example \cite{tanaka_2020,lee_2018,wang_2020}, no hyperparameter tuning is done in this work. We want to highlight, that all used training setups are chosen from one of the adapted pruning methods. Especially, FBs and FB coefficients are trained with the standard learning rates, optimized for training spatial coefficients. 

For PaI on CIFAR-$10$, we used the setup from \cite{lee_2018}, whereas SET uses the training schedule from \cite{liu_2021b}. The CIFAR-$10$ experiment for LTs is equal to the one used in \cite{frankle_2020a}. For ImageNet on ResNet$18$, we used the standard PyTorch ImageNet training \href{https://github.com/pytorch/examples/tree/master/imagenet}{(see this link)} (BSD 3-clause license) as baseline for our training -- the same as used in \cite{wang_2020}. Results can be further improved by adding learning rate warm-up for $5$ epochs \cite{goyal_2017} and label smoothing \cite{szegedy_2016} with smoothing parameter $\varepsilon = 0.1$. This improvement is inherited from \cite{liu_2021b} and applied to train RigL on ResNet$50$. For FT, we use the same training hyperparameters as for RigL whereas we use the suggested one from the original paper \cite{gale_2019} for GMP.

\paragraph{CIFAR-10.}
CIFAR-$10$ \cite{krizhevsky_2012} consists of  $60,000$ $32 \times 32$ RGB images. CIFAR-$10$ is a publicly available dataset with, best to our knowledge, no existing licenses. Authors are allowed to use the datasets for publications if the tech report \cite{krizhevsky_2012} is referenced. CIFAR-$10$ has $10$ classes with $6,000$ images per class. The data is split into $50,000$ training and $10,000$ test images. For each training, we randomly split the training images into two parts, $45,000$ images for training and $5,000$ images for validation. All images for training, validation and testing are normalized by their channel wise mean and standard deviation. Furthermore, we additionally use the standard data augmentation for CIFAR on the training images. This is given by cropping and random horizontal flipping of the images. Test results are reported for the early
stopping epoch, the epoch with the highest validation accuracy. Used hyperparameters are summarized in \cref{tab:exp_setups}. 

\paragraph{ImageNet.}
The ImageNet ILSVRC$2012$ \cite{imagenet_2012} dataset is an image classification dataset, containing approximately $1.2$ million RGB images for training and $150,000$ RGB images for validation, divided into $1,000$ classes. ImageNet has a custom license allowing non-commercial research. To be allowed to use ImageNet for non-commercial research, access to the image database has to be requested -- which we did. Full terms for the usage of the ImageNet database can be found in \href{https://www.image-net.org/download}{this link}.

Again, all images are normalized for each channel. Training images are randomly cropped to size $224 \times 224$ and randomly flipped in the horizontal direction. The validation images are resized to size $256 \times 256$ and their central $224 \times 224$ pixels are used for validation. For the ResNet$50$ experiment, we also add label smoothing on the training loss with a smoothing factor $0.1$. As common in the literature, we report results on the validation set, since labels for the test set are publicly not available. Hyperparameters are provided in \cref{tab:exp_setups}.

\section{Network architectures}\label{sec:architectures}
In this Section we provide the used network architectures VGG$16$ \cite{simonyan_2014} and the adapted version VGG$16$-LT for CIFAR-$10$, 
as well as ResNet$18$ and ResNet$50$ \cite{he_2016} for ImageNet. 

The architectures are shown in \cref{tab:VGG,tab:ResNetImageNet,tab:ResNet50ImageNet}. Note, we use two different versions of a VGG$16$, a small one for the LT experiments and a bigger one for the remaining experiments. A graphical description of the residual block and the bottleneck block used for ResNets, is shown in \cref{fig:res_block,fig:res_block_bottleneck}, respectively. Furthermore, all architectures are used in their standard form or with FB convolutions. Therefore, (FB) always indicates, that a FB version of the filter is used for the FB-CNN. Not all convolutional layers are marked with a (FB), since we only apply the FB formulation on convolutional layers with kernel size $K > 1$. Additionally, we indicate the layers which share one FB for \texttt{coarse}, \texttt{medium} and \texttt{fine} FB sharing in \cref{tab:VGG,tab:ResNetImageNet,tab:ResNet50ImageNet}.

All tensor dimensions of standard 2D convolutional filters
are given as $\co \times \ci \times K \times K$,
where $\ci$ equals the number of input channels, $\co$ the
number of output channels and $K \times K$ the size of each
convolutional kernel. FB convolutions have coefficients represented by a tensor of size $\co \times \ci \times K^2$. For pooling layers, $K \times K$ denotes the tiling size. In the case of linear layers, the tensor size is
given as $\co \times \ci$, where $\ci$ is the number of
incoming neurons and $\co$ the number of outgoing neurons. For
training the networks, we used the cross entropy loss function. 

\begin{table*}[tb]
	\def\arrvline{\kern0pt \hfill{\vrule width 5pt}\hfill \kern0pt}
	\def\smallline{\kern0pt \hfill{\vrule width 1pt}\hfill \kern0pt}
	\def\halfline{\rule[-1pt]{5pt}{7pt}}
	\def\almostline{\rule[-1pt]{5pt}{9pt}}
	\def\almosttopline{\rule[-4pt]{5pt}{10pt}}
	\centering
	\makebox[\textwidth]{
		{\scriptsize{}}%
		\begin{tabular}{@{}cccccccccccccl@{}}
			\hline
			{\scriptsize{}Module} & {\scriptsize{}Output Size} & {\scriptsize{}$\ci$} & {\scriptsize{}$\co$} & {\scriptsize{}$K$} & {\scriptsize{}Repeat} & {\scriptsize{}Stride} & {\scriptsize{}Padding} & {\scriptsize{}Bias} & {\scriptsize{}BatchNorm} & {\scriptsize{}ReLU} & {\scriptsize{}\texttt{Coarse}} & {\scriptsize{}\texttt{Medium}} & {\scriptsize{}\texttt{Fine}}\tabularnewline
			\hline 
			{\scriptsize{}\texttt{(FB) Conv2D}} & {\scriptsize{}$32\times32$} & {\scriptsize{}$3$} & {\scriptsize{}$64$} & {\scriptsize{}$3\times3$} & {\scriptsize{}$\times1$} & {\scriptsize{}$1\times1$} & {\scriptsize{}$1\times1$} & {\scriptsize{}\cmark} & {\scriptsize{}\cmark} & {\scriptsize{}\cmark} & \almosttopline & \almosttopline & \halfline\tabularnewline
			{\scriptsize{}\texttt{(FB) Conv2D}} & {\scriptsize{}$32\times32$} & {\scriptsize{}$64$} & {\scriptsize{}$64$} & {\scriptsize{}$3\times3$} & {\scriptsize{}$\times1$} & {\scriptsize{}$1\times1$} & {\scriptsize{}$1\times1$} & {\scriptsize{}\cmark} & {\scriptsize{}\cmark} & {\scriptsize{}\cmark} & \arrvline & \arrvline & \halfline \tabularnewline
			{\scriptsize{}\texttt{MaxPool2D}} & {\scriptsize{}$16\times16$} & {\scriptsize{}$64$} & {\scriptsize{}$64$} & {\scriptsize{}$2\times2$} & {\scriptsize{}$\times1$} & {\scriptsize{}$2\times2$} & {\scriptsize{}$0\times0$} & {\scriptsize{}\xmark} & {\scriptsize{}\xmark} & {\scriptsize{}\xmark} & \smallline & &\tabularnewline
			{\scriptsize{}\texttt{(FB) Conv2D}} & {\scriptsize{}$16\times16$} & {\scriptsize{}$64$} & {\scriptsize{}$128$} & {\scriptsize{}$3\times3$} & {\scriptsize{}$\times1$} & {\scriptsize{}$1\times1$} & {\scriptsize{}$1\times1$} & {\scriptsize{}\cmark} & {\scriptsize{}\cmark} & {\scriptsize{}\cmark}& \arrvline & \arrvline & \halfline \tabularnewline
			{\scriptsize{}\texttt{(FB) Conv2D}} & {\scriptsize{}$16\times16$} & {\scriptsize{}$128$} & {\scriptsize{}$128$} & {\scriptsize{}$3\times3$} & {\scriptsize{}$\times1$} & {\scriptsize{}$1\times1$} & {\scriptsize{}$1\times1$} & {\scriptsize{}\cmark} & {\scriptsize{}\cmark} & {\scriptsize{}\cmark}& \arrvline & \arrvline& \halfline\tabularnewline
			{\scriptsize{}\texttt{MaxPool2D}} & {\scriptsize{}$8\times8$} & {\scriptsize{}$128$} & {\scriptsize{}$128$} & {\scriptsize{}$2\times2$} & {\scriptsize{}$\times1$} & {\scriptsize{}$2\times2$} & {\scriptsize{}$0\times0$} & {\scriptsize{}\xmark} & {\scriptsize{}\xmark} & {\scriptsize{}\xmark}& \smallline & &\tabularnewline
			{\scriptsize{}\texttt{(FB) Conv2D}} & {\scriptsize{}$8\times8$} & {\scriptsize{}$128$} & {\scriptsize{}$256$} & {\scriptsize{}$3\times3$} & {\scriptsize{}$\times1$} & {\scriptsize{}$1\times1$} & {\scriptsize{}$1\times1$} & {\scriptsize{}\cmark} & {\scriptsize{}\cmark} & {\scriptsize{}\cmark}& \arrvline & \arrvline& \halfline\tabularnewline
			{\scriptsize{}\texttt{(FB) Conv2D}} & {\scriptsize{}$8\times8$} & {\scriptsize{}$256$} & {\scriptsize{}$256$} & {\scriptsize{}$3\times3$} & {\scriptsize{}$\times2$} & {\scriptsize{}$1\times1$} & {\scriptsize{}$1\times1$} & {\scriptsize{}\cmark} & {\scriptsize{}\cmark} & {\scriptsize{}\cmark}& \arrvline &\arrvline & \halfline {\scriptsize{}$\times 2$} \tabularnewline
			{\scriptsize{}\texttt{MaxPool2D}} & {\scriptsize{}$4\times4$} & {\scriptsize{}$256$} & {\scriptsize{}$256$} & {\scriptsize{}$2\times2$} & {\scriptsize{}$\times1$} & {\scriptsize{}$2\times2$} & {\scriptsize{}$0\times0$} & {\scriptsize{}\xmark} & {\scriptsize{}\xmark} & {\scriptsize{}\xmark}& \smallline & &\tabularnewline
			{\scriptsize{}\texttt{(FB) Conv2D}} & {\scriptsize{}$4\times4$} & {\scriptsize{}$256$} & {\scriptsize{}$512$} & {\scriptsize{}$3\times3$} & {\scriptsize{}$\times1$} & {\scriptsize{}$1\times1$} & {\scriptsize{}$1\times1$} & {\scriptsize{}\cmark} & {\scriptsize{}\cmark} & {\scriptsize{}\cmark}& \arrvline &\arrvline & \halfline\tabularnewline
			{\scriptsize{}\texttt{(FB) Conv2D}} & {\scriptsize{}$4\times4$} & {\scriptsize{}$512$} & {\scriptsize{}$512$} & {\scriptsize{}$3\times3$} & {\scriptsize{}$\times2$} & {\scriptsize{}$1\times1$} & {\scriptsize{}$1\times1$} & {\scriptsize{}\cmark} & {\scriptsize{}\cmark} & {\scriptsize{}\cmark}& \arrvline & \arrvline& \halfline {\scriptsize{}$\times 2$}\tabularnewline
			{\scriptsize{}\texttt{MaxPool2D}} & {\scriptsize{}$2\times2$} & {\scriptsize{}$512$} & {\scriptsize{}$512$} & {\scriptsize{}$2\times2$} & {\scriptsize{}$\times1$} & {\scriptsize{}$2\times2$} & {\scriptsize{}$0\times0$} & {\scriptsize{}\xmark} & {\scriptsize{}\xmark} & {\scriptsize{}\xmark}& \smallline & &\tabularnewline
			{\scriptsize{}\texttt{(FB) Conv2D}} & {\scriptsize{}$2\times2$} & {\scriptsize{}$512$} & {\scriptsize{}$512$} & {\scriptsize{}$3\times3$} & {\scriptsize{}$\times3$} & {\scriptsize{}$1\times1$} & {\scriptsize{}$1\times1$} & {\scriptsize{}\cmark} & {\scriptsize{}\cmark} & {\scriptsize{}\cmark}& \almostline & \halfline & \halfline {\scriptsize{}$\times 3$}\tabularnewline
			{\scriptsize{}\texttt{MaxPool2D}} & {\scriptsize{}$1\times1$} & {\scriptsize{}$512$} & {\scriptsize{}$512$} & {\scriptsize{}$2\times2$} & {\scriptsize{}$\times1$} & {\scriptsize{}$2\times2$} & {\scriptsize{}$0\times0$} & {\scriptsize{}\xmark} & {\scriptsize{}\xmark} & {\scriptsize{}\xmark}& & &\tabularnewline
			{\scriptsize{}\texttt{Linear}} & {\scriptsize{}$512$} & {\scriptsize{}$512$} & {\scriptsize{}$512$} & {\scriptsize{}---} & {\scriptsize{}$\times2$} & {\scriptsize{}---} & {\scriptsize{}---} & {\scriptsize{}\cmark} & {\scriptsize{}\cmark} & {\scriptsize{}\cmark}& \multicolumn{3}{c}{{\scriptsize{}\bfseries {\color{red}Removed for VGG16-LT}}}\tabularnewline
			{\scriptsize{}\texttt{Linear}} & {\scriptsize{}$10$} & {\scriptsize{}$512$} & {\scriptsize{}$10$} & {\scriptsize{}---} & {\scriptsize{}$\times1$} & {\scriptsize{}---} & {\scriptsize{}---} & {\scriptsize{}\cmark} & {\scriptsize{}\xmark} & {\scriptsize{}\xmark}& & &\tabularnewline
			\hline
		\end{tabular}
	}
	\caption{VGG$16$ and VGG$16$-LT for CIFAR-$10$ with \texttt{coarse}, \texttt{medium} and \texttt{fine} FB sharing schemes specified in the last three columns. Each \halfline{}, connected by either another \halfline{} or {\rule{1pt}{7pt}}, corresponds to exactly one FB $\F$ shared for all filters in the corresponding layers. The thin connection {\rule{1pt}{7pt}} corresponds to \texttt{MaxPool2D} layers which do not use the FBs themselves. Note, for VGG$16$-LT, the first and second \texttt{Linear} layers are removed. 
	}
	\label{tab:VGG}
\end{table*}

\begin{table*}[tb]
	\def\arrvline{\kern0pt \hfill{\vrule width 5pt}\hfill \kern0pt}
	\def\smallline{\kern0pt \hfill{\vrule width 1pt}\hfill \kern0pt}
	\def\halfline{\rule[-1pt]{5pt}{7pt}}
	\def\almostline{\rule[-1pt]{5pt}{9.5pt}}
	\def\almosttopline{\rule[-4pt]{5pt}{10pt}}
	\centering
	\makebox[\textwidth]{
		{\scriptsize{}}%
		\begin{tabular}{@{}ccccccccccccll@{}}
			\hline
			{\scriptsize{}Module} & {\scriptsize{}Output Size} & {\scriptsize{}$\ci$} & {\scriptsize{}$\co$} & {\scriptsize{}$K$} & {\scriptsize{}Stride} & {\scriptsize{}Padding} & {\scriptsize{}Bias} & {\scriptsize{}BatchNorm} & {\scriptsize{}ReLU} & {\scriptsize{}\texttt{Coarse}} & {\scriptsize{}\texttt{Medium}}  & {\scriptsize{}\texttt{Fine}} \tabularnewline
			\hline 
			{\scriptsize{}\texttt{Conv2D}} & {\scriptsize{}$112 \times 112$} & {\scriptsize{}$3$} & {\scriptsize{}$64$} & {\scriptsize{}$7 \times 7$} & {\scriptsize{}$2 \times 2 $} & {\scriptsize{}$3 \times 3$} & {\scriptsize{}\xmark} & {\scriptsize{}\cmark} & {\scriptsize{}\cmark}&  &  &    \tabularnewline
			{\scriptsize{}\texttt{MaxPool2D}} & {\scriptsize{}$56 \times 56$} & {\scriptsize{}$64$} & {\scriptsize{}$64$} & {\scriptsize{}$3 \times 3$} & {\scriptsize{}$2 \times 2$} & {\scriptsize{}$1 \times 1$} & {\scriptsize{}\xmark} & {\scriptsize{}\xmark} & {\scriptsize{}\xmark} &  &  &  \tabularnewline
			{\scriptsize{}\texttt{(FB) ResBlock}$\times 2$} & {\scriptsize{}$56 \times 56$} & {\scriptsize{}$64$} & {\scriptsize{}$64$} & {\scriptsize{}$3 \times 3$} & {\scriptsize{}$1 \times 1$} & {\scriptsize{}$1 \times 1$} & {\scriptsize{}\xmark} & {\scriptsize{}\cmark} & {\scriptsize{}\cmark}& \almosttopline & \halfline  & \halfline {\scriptsize{}$\times 4$}\tabularnewline
			{\scriptsize{}\texttt{(FB) ResBlock}$\times 2$} & {\scriptsize{}$28 \times 28$} & {\scriptsize{}$64$} & {\scriptsize{}$128$} & {\scriptsize{}$3 \times 3 $} & {\scriptsize{}$2 \times 2 $} & {\scriptsize{}$1 \times 1$} & {\scriptsize{}\xmark} & {\scriptsize{}\cmark} & {\scriptsize{}\cmark} & \arrvline & \halfline & \halfline {\scriptsize{}$\times 4$}\tabularnewline
			{\scriptsize{}\texttt{(FB) ResBlock}$\times 2$} & {\scriptsize{}$14 \times 14$} & {\scriptsize{}$128$} & {\scriptsize{}$256$} & {\scriptsize{}$3 \times 3$} & {\scriptsize{}$2 \times 2$} & {\scriptsize{}$1 \times 1$} & {\scriptsize{}\xmark} & {\scriptsize{}\cmark} & {\scriptsize{}\cmark} & \arrvline & \halfline  & \halfline {\scriptsize{}$\times 4$}\tabularnewline
			{\scriptsize{}\texttt{(FB) ResBlock}$\times 2$} & {\scriptsize{}$7 \times 7$} & {\scriptsize{}$256$} & {\scriptsize{}$512$} & {\scriptsize{}$3 \times 3$} & {\scriptsize{}$2 \times 2$} & {\scriptsize{}$1 \times 1$} & {\scriptsize{}\xmark} & {\scriptsize{}\cmark} & {\scriptsize{}\cmark} & \almostline & \halfline  & \halfline {\scriptsize{}$\times 4$}\tabularnewline
			{\scriptsize{}\texttt{AvgPool2D}} & {\scriptsize{}$1 \times 1$} & {\scriptsize{}$512$} & {\scriptsize{}$512$} & {\scriptsize{}$7 \times 7$} & {\scriptsize{}$0 \times 0$} & {\scriptsize{}$0 \times 0$} & {\scriptsize{}\xmark} & {\scriptsize{}\xmark} & {\scriptsize{}\xmark}\tabularnewline
			{\scriptsize{}\texttt{Linear}} & {\scriptsize{}$1,000$} & {\scriptsize{}$512$} & {\scriptsize{}$1,000$} & {\scriptsize{}---} & {\scriptsize{}---} & {\scriptsize{}---} & {\scriptsize{}\cmark} & {\scriptsize{}\xmark} & {\scriptsize{}\xmark}\tabularnewline
			\hline
	\end{tabular}}
	\caption{ResNet$18$ for ImageNet with \texttt{(FB) ResBlock}$\times 2$, shown in
		\cref{fig:res_block}. Last three columns declare \texttt{coarse}, \texttt{medium} and \texttt{fine} FB sharing schemes. Each \halfline{} connected by another \halfline{} corresponds to exactly one FB $\F$ shared for all filters in the corresponding layers. Note, the first convolutional layer has kernel size $7 \times 7$ and we do not use a FB formulation for this layer.}
	\label{tab:ResNetImageNet}
\end{table*}

\begin{table*}[tb]
	\def\arrvline{\kern0pt \hfill{\vrule width 5pt}\hfill \kern0pt}
	\def\smallline{\kern0pt \hfill{\vrule width 1pt}\hfill \kern0pt}
	\def\halfline{\rule[-1pt]{5pt}{7pt}}
	\def\almostline{\rule[-1pt]{5pt}{9.5pt}}
	\def\almosttopline{\rule[-4pt]{5pt}{10pt}}
	\centering
	\makebox[\textwidth]{
		{\scriptsize{}}%
		\begin{tabular}{@{}ccccccccccccccll@{}}
			\hline
			{\scriptsize{}Module} & {\scriptsize{}Output Size} & {\scriptsize{}$\ci$} & {\scriptsize{}$c_{mid}$} & {\scriptsize{}$\co$} & {\scriptsize{}$n_b$} & {\scriptsize{}$K$} & {\scriptsize{}Stride} & {\scriptsize{}Padding} & {\scriptsize{}Bias} & {\scriptsize{}BN} & {\scriptsize{}ReLU} & {\begin{tabular}[x]{@{}c@{}}\scriptsize{}\texttt{Coarse} \\ \scriptsize{}adapt\end{tabular}} & {\scriptsize{}\texttt{Medium}}  & {\scriptsize{}\texttt{Fine}} \tabularnewline
			\hline 
			{\scriptsize{}\texttt{(FB) Conv2D}} & {\scriptsize{}$112 \times 112$} & {\scriptsize{}$3$} & {\scriptsize{}---} & {\scriptsize{}$64$} & {\scriptsize{}---} & {\scriptsize{}$7 \times 7$} & {\scriptsize{}$2 \times 2 $} & {\scriptsize{}$3 \times 3$} & {\scriptsize{}\xmark} & {\scriptsize{}\cmark} & {\scriptsize{}\cmark} &\halfline &   \halfline  & \halfline{\scriptsize{}$\times 1$}     \tabularnewline
			{\scriptsize{}\texttt{MaxPool2D}} & {\scriptsize{}$56 \times 56$} & {\scriptsize{}$64$} & {\scriptsize{}---}  & {\scriptsize{}$64$}  & {\scriptsize{}---} & {\scriptsize{}$3 \times 3$} & {\scriptsize{}$2 \times 2$} & {\scriptsize{}$1 \times 1$} & {\scriptsize{}\xmark} & {\scriptsize{}\xmark} & {\scriptsize{}\xmark} &  &  &    \tabularnewline
			{\scriptsize{}\texttt{(FB) BottleneckBlock}} & {\scriptsize{}$56 \times 56$} & {\scriptsize{}$64$}  & {\scriptsize{}$64$} & {\scriptsize{}$256$}  & {\scriptsize{}$3$} & {\scriptsize{}$3 \times 3$} & {\scriptsize{}$1 \times 1$} & {\scriptsize{}$1 \times 1$} & {\scriptsize{}\xmark} & {\scriptsize{}\cmark} & {\scriptsize{}\cmark} & \almosttopline  & \halfline  & \halfline {\scriptsize{}$\times 3$}\tabularnewline
			{\scriptsize{}\texttt{(FB) BottleneckBlock}} & {\scriptsize{}$28 \times 28$} & {\scriptsize{}$256$} & {\scriptsize{}$128$}  & {\scriptsize{}$512$}  & {\scriptsize{}$4$} & {\scriptsize{}$3 \times 3 $} & {\scriptsize{}$2 \times 2 $} & {\scriptsize{}$1 \times 1$} & {\scriptsize{}\xmark} & {\scriptsize{}\cmark} & {\scriptsize{}\cmark} & \arrvline & \halfline & \halfline {\scriptsize{}$\times 4$}\tabularnewline
			{\scriptsize{}\texttt{(FB) BottleneckBlock}} & {\scriptsize{}$14 \times 14$} & {\scriptsize{}$512$} & {\scriptsize{}$256$}  &   {\scriptsize{}$1,024$}  & {\scriptsize{}$6$} & {\scriptsize{}$3 \times 3$} & {\scriptsize{}$2 \times 2$} & {\scriptsize{}$1 \times 1$} & {\scriptsize{}\xmark} & {\scriptsize{}\cmark} & {\scriptsize{}\cmark} &\arrvline  & \halfline  & \halfline {\scriptsize{}$\times 6$}\tabularnewline
			{\scriptsize{}\texttt{(FB) BottleneckBlock}} & {\scriptsize{}$7 \times 7$} & {\scriptsize{}$1,024$} & {\scriptsize{}$512$}   & {\scriptsize{}$2,048$}  & {\scriptsize{}$3$} & {\scriptsize{}$3 \times 3$} & {\scriptsize{}$2 \times 2$} & {\scriptsize{}$1 \times 1$} & {\scriptsize{}\xmark} & {\scriptsize{}\cmark} & {\scriptsize{}\cmark} & \almostline & \halfline  & \halfline {\scriptsize{}$\times 3$}\tabularnewline
			{\scriptsize{}\texttt{AvgPool2D}} & {\scriptsize{}$1 \times 1$} & {\scriptsize{}$512$}& {\scriptsize{}---}  & {\scriptsize{}$512$}  & {\scriptsize{}---} & {\scriptsize{}$7 \times 7$} & {\scriptsize{}$0 \times 0$} & {\scriptsize{}$0 \times 0$} & {\scriptsize{}\xmark} & {\scriptsize{}\xmark} & {\scriptsize{}\xmark}\tabularnewline
			{\scriptsize{}\texttt{Linear}} & {\scriptsize{}$1,000$} &  {\scriptsize{}$512$} & {\scriptsize{}---} & {\scriptsize{}$1,000$} & {\scriptsize{}---} & {\scriptsize{}---} & {\scriptsize{}---} & {\scriptsize{}---} & {\scriptsize{}\cmark} & {\scriptsize{}\xmark} & {\scriptsize{}\xmark}\tabularnewline
			\hline
	\end{tabular}}
	\caption{ResNet$50$ for ImageNet with \texttt{(FB) BottleneckBlock}, shown in
		\cref{fig:res_block_bottleneck}. Last three columns declare the adapted \texttt{coarse}, \texttt{medium} and \texttt{fine} FB sharing schemes. Each \halfline{} connected by another \halfline{} corresponds to exactly one FB $\F$ shared for all filters in the corresponding layers.}
	\label{tab:ResNet50ImageNet}
\end{table*}

\begin{figure}[t]
	\centering
			\tikzset{My Style/.style={draw, black, rectangle, minimum size=.6cm, minimum width=6.5cm, node distance = 0pt}}
	\tikzset{My Style2/.style={rectangle, minimum size=.5cm, minimum width=5.5cm, node distance = 0pt}}
	\tikzset{My Header/.style={minimum size=.5cm, minimum width=6.5cm, node distance = 0pt}}
			\begin{tikzpicture}
			\node (img) {};
			\node[My Style2, below=of img] (arrow1) {};
			\node[My Style2, below=of arrow1] (arrow2) {};
			\node[My Style, below=of arrow2, fill=brown!20] (input1) {$3\times 3$ \texttt{(FB) Conv2D}$(c_{in}, c_{out}, s \times s)$}; 
			\node[below =0pt of img] (dum_vert0){};
			\node[above =0pt of input1] (dum_vert1){};
			\draw[->, thick] (dum_vert0.center) -- (dum_vert1.center);
			
			\node[My Style, below=of input1, fill=gray!20] (input2) {\texttt{BatchNorm2D}}; 
			\node[My Style, below=of input2, fill=green!10] (input3) {\texttt{ReLU}};
			
			\node[My Style, below=of input3, fill=brown!20] (input4) {$3\times 3$ \texttt{(FB) Conv2D}$(c_{out}, c_{out}, 1\times 1)$}; 
			\node[My Style, below=of input4, fill=gray!20] (input5) {\texttt{BatchNorm2D}};

			\node[My Style2, below=of input5] (arrow3) {};
			\node[My Style2, below=of arrow3] (arrow3half) {};
			\node[My Style2, below=of arrow3half] (arrow4) {\scalebox{2.5}{$\oplus$}};
			\node[My Style2, below=of arrow4] (arrow5) {};
			\node[My Style2, below=of arrow5] (arrow5half) {};
			
			\node[My Style, below=of arrow5half, fill=green!10] (input6) {\texttt{ReLU}};
			\node[My Style2, below=of input6] (arrow6) {};
			\node[My Style2, below=of arrow6] (arrow6half) {};
			\node[left = of input3] (res_dum1) {};
			\node[My Style, left = of res_dum1, fill=brown!10] (res1) {$1\times1$ \texttt{Conv2D}$(c_{in}, c_{out}, s\times s)$};
			\node[My Style, below = of res1, fill=gray!20] (res_bn1) {\texttt{BatchNorm2D}};
			
			\node[right = of input3] (res_dum11) {};
			\node[My Style, right = of res_dum11, fill=brown!10] (res11) {\texttt{Identity}};

			\draw[->, thick] (arrow1.west) to [out=180, in=90] (res1.north) {};
			\draw[->, thick] (arrow1.east) to [out=0, in=90] (res11.north) {};
			\draw[-, thick] (arrow1.west) -- (arrow1.center);
			\draw[-, thick] (arrow1.east) -- (arrow1.center);
						
			\draw[-, thick] (res_bn1.south) to [out=-90, in=180] (arrow4.west) {};
			\draw[-, thick] (res11.south) to [out=-90, in=0] (arrow4.east) {};	
		
			\node[left = -70pt of arrow4] (dum_odot1){};
			\node[right = -70pt of arrow4] (dum_odot11){};
			\draw[->, thick] (arrow4.west) -- (dum_odot1.center);
			\draw[->, thick] (arrow4.east) -- (dum_odot11.center);
			\node[below =0pt of input5] (dum_vert2){};
			\draw[->, thick] (dum_vert2.center) -- (arrow4.north) {}; 
			
			\draw[-, thick] (arrow6.center) -- (arrow6.west) ;
			\node[My Style, below=of arrow6half, fill=blue!10] (input7) {$3\times 3$ \texttt{(FB) Conv2D}$(c_{out}, c_{out}, 1\times 1)$}; 
			
			\node[above =0pt of input6] (dum_vert3){};
			\draw[->, thick] (arrow4.south) -- (dum_vert3.center) {}; 
			\node[below =0pt of input6] (dum_vert4){};
			\node[above =0pt of input7] (dum_vert5){};
			\draw[->, thick] (dum_vert4.center) -- (dum_vert5.center) {}; 
			
			\node[My Style, below=of input7, fill=gray!20] (input8) {\texttt{BatchNorm2D}}; 
			\node[My Style, below=of input8, fill=green!10] (input9) {\texttt{ReLU}};
			
			\node[My Style, below=of input9, fill=blue!10] (input10) {$3\times 3$ \texttt{(FB) Conv2D}$(c_{out}, c_{out}, 1\times 1)$}; 
			
			\node[My Style, below=of input10, fill=gray!20] (input11) {\texttt{BatchNorm2D}}; 
			
			\node[My Style2, below=of input11] (arrow7) {};
			\node[My Style2, below=of arrow7] (arrow7half) {};
			\node[My Style2, below=of arrow7half] (arrow8) {\scalebox{2.5}{$\oplus$}};
			\node[My Style2, below=of arrow8] (arrow9) {};
			\node[My Style2, below=of arrow9] (arrow9half) {};
			\node[My Style, below=of arrow9half, fill=green!10] (input12) {\texttt{ReLU}};
			\node[My Style2, below=of input12] (arrow10) {};
			\node[My Style2, below=of arrow10] (arrow10half) {};
			
			\node[left = of input9] (res_dum2) {};
			\node[My Style, left = of res_dum2, fill=brown!10] (res2) {\texttt{Identity}};
			\draw[->, thick] (arrow6.west) to [out=180, in=90] (res2.north) {};
			
			\draw[-, thick] (res2.south) to [out=-90, in=180] (arrow8.west) {};
			\node[left = -70pt of arrow8] (dum_odot2){};
			\draw[->, thick] (arrow8.west) -- (dum_odot2.center);
			
			\node[below =0pt of input11] (dum_vert6){};			
			\draw[->, thick] (dum_vert6.center) -- (arrow8.north) {}; 
			\node[above =0pt of input12] (dum_vert7){};			
			\draw[->, thick] (arrow8.south) -- (dum_vert7.center) {}; 
			\node[below =0pt of input12] (dum_vert8){};		
			\node[below =0pt of arrow10half] (dum_vert9){};										
			\draw[->, thick] (dum_vert8.center) -- (arrow10half.south) {}; 
			
			\node[My Header,above left=25pt and -90pt of img] (t1) {\large \textbf {\texttt{(FB) ResBlock}$\times 2 (c_{in}, c_{out}, s\times s)$}};
			
			\node[above left=-10pt and 5pt of img] (t1) {$\leftarrow$ $s > 2$ or $c_{in} \neq c_{out}$ $\leftarrow$};
			\node[above right=-10pt and 5pt of img] (t1) {$\rightarrow$ $s = 1$ and $c_{in} = c_{out}$ $\rightarrow$};
			\end{tikzpicture} 	
	\caption{\label{fig:res_block}Architecture of a \texttt{(FB) ResBlock}$\times 2$ with $\ci$
		input channels, $\co$ output channels and stride $s \times s$
		for the first (FB) convolution. The first residual connection is is a standard $1\times1$\underline{}
		\texttt{2D Convolution} followed by a \texttt{BatchNorm2D} layer if $s > 1$ or $\ci \neq \co$. However, if $\ci = \co$ and $s = 1$, the first residual connection is simply an identity mapping. The second residual connection is always an identity mapping.
		All \texttt{(FB) Conv2D} layers do not have biases.} 
\end{figure}

\begin{figure}[tb]
	\centering
			\tikzset{My Style/.style={draw, black, rectangle, minimum size=.6cm, minimum width=6.5cm, node distance = 0pt}}
	\tikzset{My Style2/.style={rectangle, minimum size=.5cm, minimum width=5.5cm, node distance = 0pt}}
	\tikzset{My Header/.style={minimum size=.5cm, minimum width=6.5cm, node distance = 0pt}}
			\begin{tikzpicture}
			\node (img) {};
			\node[My Style2, below=of img] (arrow1) {};
			\node[My Style2, below=of arrow1] (arrow2) {};
			\node[My Style, below=of arrow2, fill=brown!20] (input1) {$1\times 1$ \texttt{Conv2D}$(c_{in}, c_{mid}, s \times s)$}; 
			\node[below =0pt of img] (dum_vert0){};
			\node[above =0pt of input1] (dum_vert1){};
			\draw[->, thick] (dum_vert0.center) -- (dum_vert1.center);
			
			\node[My Style, below=of input1, fill=gray!20] (input2) {\texttt{BatchNorm2D}}; 
			\node[My Style, below=of input2, fill=green!10] (input3) {\texttt{ReLU}};
			\node[My Style, below=of input3, fill=brown!40] (input31) {$3\times 3$ \texttt{(FB) Conv2D}$(c_{mid}, c_{mid}, 1 \times 1)$}; 
			\node[My Style, below=of input31, fill=gray!20] (input32) {\texttt{BatchNorm2D}}; 
			\node[My Style, below=of input32, fill=green!10] (input33) {\texttt{ReLU}};
			
			\node[My Style, below=of input33, fill=brown!20] (input4) {$1 \times 1$ \texttt{Conv2D}$(c_{mid}, c_{out}, 1\times 1)$}; 
			\node[My Style, below=of input4, fill=gray!20] (input5) {\texttt{BatchNorm2D}};

			\node[My Style2, below=of input5] (arrow3) {};
			\node[My Style2, below=of arrow3] (arrow3half) {};
			\node[My Style2, below=of arrow3half] (arrow4) {\scalebox{2.5}{$\oplus$}};
			\node[My Style2, below=of arrow4] (arrow5) {};
			\node[My Style2, below=of arrow5] (arrow5half) {};
			
			\node[My Style, below=of arrow5half, fill=green!10] (input6) {\texttt{ReLU}};
			\node[My Style2, below=of input6] (arrow6) {};
			\node[My Style2, below=of arrow6] (arrow6half) {};
			\node[left = of input31] (res_dum1) {};
			\node[My Style, left = of res_dum1, fill=brown!10] (res1) {$1\times1$ \texttt{Conv2D}$(c_{in}, c_{out}, s\times s)$};
			\node[My Style, below = of res1, fill=gray!20] (res_bn1) {\texttt{BatchNorm2D}};
			

			\draw[->, thick] (arrow1.west) to [out=180, in=90] (res1.north) {};
			\draw[-, thick] (arrow1.west) -- (arrow1.center);
						
			\draw[-, thick] (res_bn1.south) to [out=-90, in=180] (arrow4.west) {};
		
			\node[left = -70pt of arrow4] (dum_odot1){};
			\node[right = -70pt of arrow4] (dum_odot11){};
			\draw[->, thick] (arrow4.west) -- (dum_odot1.center);
			\node[below =0pt of input5] (dum_vert2){};
			\draw[->, thick] (dum_vert2.center) -- (arrow4.north) {}; 
			
			\draw[-, thick] (arrow6.center) -- (arrow6.west) ;
			\node[My Style, below=of arrow6half, fill=blue!10] (input7) {$1\times 1$ \texttt{Conv2D}$(c_{out}, c_{mid}, 1\times 1)$}; 
			
			\node[above =0pt of input6] (dum_vert3){};
			\draw[->, thick] (arrow4.south) -- (dum_vert3.center) {}; 
			\node[below =0pt of input6] (dum_vert4){};
			\node[above =0pt of input7] (dum_vert5){};
			\draw[->, thick] (dum_vert4.center) -- (dum_vert5.center) {}; 
			
			\node[My Style, below=of input7, fill=gray!20] (input8) {\texttt{BatchNorm2D}}; 
			\node[My Style, below=of input8, fill=green!10] (input9) {\texttt{ReLU}};

			\node[My Style, below=of input9, fill=blue!30] (input91) {$3\times 3$ \texttt{(FB) Conv2D}$(c_{mid}, c_{mid}, 1\times 1)$}; 
			\node[My Style, below=of input91, fill=gray!20] (input92) {\texttt{BatchNorm2D}}; 
			\node[My Style, below=of input92, fill=green!10] (input93) {\texttt{ReLU}};
			
			\node[My Style, below=of input93, fill=blue!10] (input10) {$1\times 1$ \texttt{Conv2D}$(c_{mid}, c_{out}, 1\times 1)$}; 
			
			\node[My Style, below=of input10, fill=gray!20] (input11) {\texttt{BatchNorm2D}}; 
			
			\node[My Style2, below=of input11] (arrow7) {};
			\node[My Style2, below=of arrow7] (arrow7half) {};
			\node[My Style2, below=of arrow7half] (arrow8) {\scalebox{2.5}{$\oplus$}};
			\node[My Style2, below=of arrow8] (arrow9) {};
			\node[My Style2, below=of arrow9] (arrow9half) {};
			\node[My Style, below=of arrow9half, fill=green!10] (input12) {\texttt{ReLU}};
			\node[My Style2, below=of input12] (arrow10) {};
			\node[My Style2, below=of arrow10] (arrow10half) {};
			
			\node[left = of input91] (res_dum2) {};
			\node[My Style, left = of res_dum2, fill=brown!10] (res2) {\texttt{Identity}};
			\draw[->, thick] (arrow6.west) to [out=180, in=90] (res2.north) {};
			
			\draw[-, thick] (res2.south) to [out=-90, in=180] (arrow8.west) {};
			\node[left = -70pt of arrow8] (dum_odot2){};
			\draw[->, thick] (arrow8.west) -- (dum_odot2.center);
			
			\node[below =0pt of input11] (dum_vert6){};			
			\draw[->, thick] (dum_vert6.center) -- (arrow8.north) {}; 
			\node[above =0pt of input12] (dum_vert7){};			
			\draw[->, thick] (arrow8.south) -- (dum_vert7.center) {}; 
			\node[below =0pt of input12] (dum_vert8){};		
			\node[below =0pt of arrow10half] (dum_vert9){};										
			\draw[dotted, thick] (dum_vert8.center) -- (arrow10half.south) {}; 
			\node[below=-0 pt of arrow10half] {\LARGE Repeat last block $n_b-2$ times};
			\node[My Header,above left=10pt and -20pt of img] (t1) {\large \textbf {\texttt{(FB) BottleneckBlock}$ (c_{in}, c_{mid}, c_{out}, n_b, s\times s)$}};
			
			\end{tikzpicture} 	
	\caption{\label{fig:res_block_bottleneck}Architecture of a \texttt{(FB) BottleneckBlock} with $\ci$
		input channels, $c_{mid}$ middle channels and $\co$ output channels and stride $s \times s$
		for the first $1 \times 1$ (FB) convolution. The number of small blocks forming the \texttt{(FB) BottleneckBlock} is given by $n_b$. The first residual connection is is a standard $1\times1$\underline{}
		\texttt{2D Convolution} followed by a \texttt{BatchNorm2D} layer. The following residual connections are always identity mappings.
		All \texttt{(FB) Conv2D} layers do not have biases.} 
\end{figure}

\begin{figure}
	\centering
	\includegraphics[width=.45\textwidth]{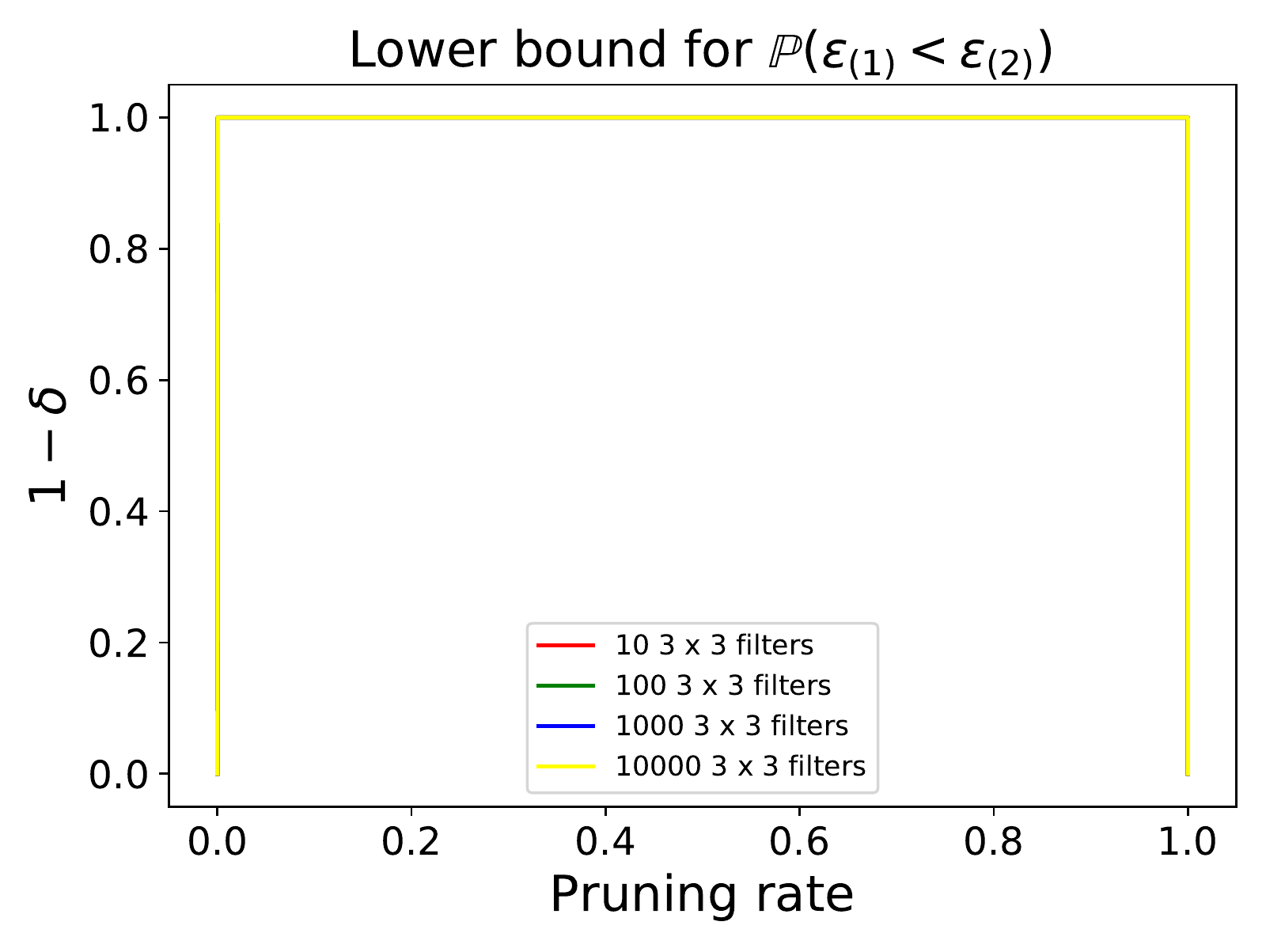}
	\caption{Lower bound $1 - \delta$ for $\mathbb{P}(\varepsilon_{(1)} < \varepsilon_{(2)})$, \ie the optimization problem \eqref{eq:spd_fb_appendix} having a \emph{smaller} solution than the problem \eqref{eq:spd_standard_appendix}. Computed for varying numbers $n$ of $3 \times 3$ filters in a layer and varying pruning rates $p \in [0,1]$.} 
	\label{fig:probabilities}
\end{figure}

\section{Proof of Theorem \ref{thm:spd}}\label{sec:proofs}
SDL optimizes a dictionary $\Fb \in \R^{m \times M}$ jointly with its coefficients $R \in \R^{M \times n}$ \wrt the non-convex problem
\begin{equation}
\inf_{\Fb, R}\Vert U - \Fb \cdot R \Vert_F \;\; \text{s.t.} \; \Vert R \Vert_0 \leq s \label{eq:spd_fb_appendix}\;,
\end{equation}
for a target $U \in \R^{m \times n}$ and sparsity constraint $s$. In our context $U$ corresponds to a convolutional layer, the dictionary $\Fb$ to the layer's FB (FD) $\F$ and $R$ to the FB (FD) coefficients. Standard magnitude pruning can be seen as a special case of SDL where the dictionary $\Fb$ is fixed to form the standard basis, \ie $\Fb = \id_{\R^m}$. Accordingly,
\begin{equation}
\inf_{\P } \Vert U - \P  \Vert_F \;\; \text{s.t.} \; \Vert R \Vert_0 \leq s \; \label{eq:spd_standard_appendix}
\end{equation}
is minimized. 

\begin{thm}\label{thm:spd_appendix}
	Let $1 < m \leq M$, $0 < s < m\cdot n$ and $U_{i,j} \sim \mathcal{N}(0,1)$ \iid. Further assume that $\varepsilon_{(1)}$ is the infimum of \cref{eq:spd_fb_appendix} and $\varepsilon_{(2)}$ the minimum of \cref{eq:spd_standard_appendix}. Assume $\Pa$ to be the minimizer for \cref{eq:spd_standard_appendix}. Then $\varepsilon_{(1)} < \varepsilon_{(2)}$ holds with probability $\mathbb{P} = 1$.
	
	If furthermore $\supp R$ for \cref{eq:spd_fb_appendix} is fixed to be equal to $\supp \Pa$, then $\varepsilon_{(1)} \leq \varepsilon_{(2)}$ and strict inequality holds with $\mathbb{P} \geq 1 - \delta$, where
	\begin{equation}
	\delta = \begin{cases}
	0 \; & , \; \text{if} \; s \not\equiv 0 (\textrm{mod}\ m)\; \\
	\frac{\binom{n}{\frac{s}{m}}}{\binom{m\cdot n}{s}} \; & , \; \text{if} \; s \equiv 0 (\textrm{mod}\ m)\; 
	\end{cases}\;.
	\end{equation}
\end{thm}

\Cref{fig:probabilities} shows the probability of the solution to \cref{eq:spd_standard_appendix} being strictly bigger than the solution of \cref{eq:spd_fb_appendix} if $\supp R$ is restricted to be $\supp \Pa$. Precisely, it shows $1 - \delta$ for varying pruning rates $p$. It can be seen that, except for the trivial case of a network being completely pruned or being not pruned at all, $\delta$ is numerically equal to zero, even for a network with only $100$ filters.\footnote{The minimum $\delta$ we computed for non-trivial pruning rates $p \in (0, 1)$ for a network with $n=10$ filters of size $3 \times 3$ was given by $\delta \leq 10^{-10}$. For $n \geq 100$, numerically $\delta=0$ for all $p \in (0,1)$.} Thus, despite $\delta > 0$, numerically the chance of $\varepsilon_{(1)} = \varepsilon_{(2)}$ \emph{is} equal to zero.

The proof of \cref{thm:spd_appendix} is split in several parts.
\begin{itemize}[noitemsep,topsep=0pt]
	\item \Cref{lem:solve_standard} shows that \cref{eq:spd_standard_appendix} always has a minimum and constructs the minimizing $\Pa$.
	\item \Cref{lem:approximator} will show that for each feasible point $\Phi_0$ for \cref{eq:spd_standard_appendix}, there exists an equivalent feasible point $(\Fb_0, R_0)$ for \cref{eq:spd_fb_appendix} with the same sparsity $\Vert R_0 \Vert_0 = \Vert \Phi_0 \Vert_0$ and distance $\Vert U - {\Fb}_0 \cdot {R}_0\Vert_F = \Vert U - \Phi_0 \Vert_F$.
	\item Consequently, \cref{cor:trivially_leq} shows that the solution to \cref{eq:spd_fb_appendix} is always smaller or equal to the solution of \cref{eq:spd_standard_appendix}. 
	\item The first part of the proof of \cref{thm:spd_appendix} shows that the solution obtained by \cref{eq:spd_standard_appendix} can only with a small chance ${\delta}_0 \leq \delta$ be the optimum of \cref{eq:spd_fb_appendix}. This is based on two facts, first we construct the equivalent point $({\Fb}_0, R_0)$ to $\Pa$ according to \cref{lem:approximator}. Then, we show that with a probability of at most $\delta$, $({\Fb}_0, R_0)$ fulfills a necessary condition for solving \cref{eq:spd_fb_appendix}. This condition is given by ${\Fb}_0$ yielding a local optimum\footnote{By convexity of $f$ it is therefore a global minimum.} of the smooth, convex function
	\begin{equation}
	f : \R^{m \times M} \rightarrow \R \; , \; \Fb \mapsto \Vert U - \Fb \cdot R_0 \Vert_F^2 \;,
	\end{equation}
	which is evaluated by looking at the probability of ${\Fb}_0$ being a root of $\frac{\partial f}{\partial \Fb}$.
	\item The second part of \cref{thm:spd_appendix} adapts $(\Fb_0, R_0)$ \emph{if} $\supp R_0$ is not fixed to be equal to $\supp \Pa$. By setting one column of $\Pa$ as a new basis element, the number of coefficients needed to match $\Pa = \Fb^\ast \cdot R^\ast$ with the adapted $(\Fb^\ast, R^\ast)$ is reduced. This of course provides new unused coefficients which are used to \emph{better approximate} the target $U$.
\end{itemize}

\begin{lem}\label{lem:solve_standard}
	The optimization problem \eqref{eq:spd_standard_appendix} always has a solution $\Pa \in \R^{m \times n}$ obtained by
	\begin{equation}
	\begin{aligned}\label{eq:standard_ast}
	& \Pa = (\Pa_{i,j})_{i,j} \; \, \; \text{with} \\ &\Pa_{i,j} = \begin{cases}
	U_{i,j} \;, & \text{if} \; (i,j) \in \tops \\
	0 \;, & \text{else}
	\end{cases}\;.
	\end{aligned}
	\end{equation}
	Here,
	\begin{equation}
	\begin{aligned}
	\tops := \{&(i_0, j_0) \in \{1, \ldots, m\} \times \{1, \ldots, n\}  : U_{i_0,j_0} \\ & \text{belongs to the top} \; s \;  \text{magnitudes of} \; \{U_{i,j}\}  \} 
	\end{aligned}
	\end{equation}
	defines the indices corresponding to the $s$ highest magnitudes of $U$.
\end{lem}

\begin{proof}[Proof of \Cref{lem:solve_standard}]
	To solve \cref{eq:spd_standard_appendix}, we rewrite the optimization problem into its equivalent, squared form
	\begin{equation}\label{eq:standard_squared}
	\inf_{\P \in \R^{m \times n}} \Vert U - {\P}  \Vert_F^2 \;\; \text{s.t.} \; \Vert {\P} \Vert_0 \leq s \;.
	\end{equation}	
	The problem \eqref{eq:standard_squared} is equivalent to
	\begin{equation}\label{eq:standard_split}
	\inf_{\P \in \bigcup_{k=1}^r \sc_k} \Vert U - {\P}  \Vert_F^2  = \min_{k \in \{1, \ldots, r\}} \inf_{\P \in \sc_k}  \Vert U - {\P}  \Vert_F^2
	\end{equation}	
	with $r = \binom{n \cdot m}{s}$,
	\begin{equation}
	\begin{aligned}
	&\sc_k = \{A \in \R^{m \times n} : \supp A \subset S_k \}, \; \\ &S_k \subset \{1, \ldots, m\} \times \{1, \ldots, n\} \; , \# S_k  = s
	\end{aligned}
	\end{equation}
	satisfying $S_k \neq S_j$ for $k \neq j$ and $\bigcup_{k=1}^r \sc_k = \{A \in \R^{m\times n} : \Vert A \Vert_0 \leq s\}$.
	
	In order to solve \cref{eq:standard_split}, we minimize for each $k$
	\begin{equation}\label{eq:spd_standard_individual}
	\inf_{\P \in \sc_k} \Vert U - \P \Vert_F^2 = \inf_{\P \in \sc_k} \sum_{i,j} (U_{i,j} - \P_{i,j} )^2 
	\end{equation}
	individually. The problem \eqref{eq:spd_standard_individual} is minimized by $\P^{(k)\ast}$ with
	\begin{equation}
	\P^{(k)\ast}_{i,j} = \begin{cases}
	U_{i,j} \;, & \text{if} \; (i,j) \in S_k \\ 0 \;, & \text{else}
	\end{cases} \;.
	\end{equation}
	Thus, the minimum of \cref{eq:spd_standard_individual} for a $k \in \{1, \ldots, r\}$ is given by
	\begin{equation}
	\Vert U - \P^{(k)\ast} \Vert_F^2 = \sum_{(i,j) \notin S_k} U_{i,j}^2\label{eq:solve_subset_prob}
	\end{equation}
	\Cref{eq:solve_subset_prob} leads to the solution of \cref{eq:standard_split}, given by
	\begin{equation}
	\min_{k\in\{1, \ldots, r\}} \sum_{(i,j) \notin S_k} U_{i,j}^2  = \max_{k \in \{1, \ldots, r\}} \sum_{(i,j) \in S_k} U_{i,j}^2 
	\end{equation}
	which is reached by choosing $k$ such that $S_k = \tops$.
\end{proof}

\begin{lem}\label{lem:approximator}
	Let $m \leq M$, then for each $\P \in \R^{m \times n}$ there exists a ${\Fb} \in \R^{m \times M}$ and a ${R} \in \R^{M \times n}$ with $\Vert {R} \Vert_0 = \Vert \P \Vert_0$ and ${\Fb} \cdot {R} = \P$.
\end{lem}
\begin{proof}[Proof of \Cref{lem:approximator}]
	Let $\P = (\P_{i,j})_{i,j} \in \R^{m \times n}$ be given. Now, we define ${R} \in \R^{M \times n}$ and ${\Fb} \in \R^{m \times M}$ via
	\begin{equation}\label{eq:construct_last}
	{R}_{i,j} = \begin{cases}
	\P_{i,j} \; & , \text{if} \; i \leq m \\ 0 \; & , \text{else}
	\end{cases}
	\end{equation}
	and
	\begin{equation}\label{eq:construct_fast}
	{\Fb}_{i,j} = \begin{cases}
	1 \; & , \text{if} \; i = j \; \text{and} \; j \leq m \\ 0 \; & , \text{else}
	\end{cases}.
	\end{equation}
	By construction of ${R}$ and ${\Fb}$, it holds $\P = {\Fb} \cdot {R}$ and $\Vert {R} \Vert_0 = \Vert \P \Vert_0$. 
\end{proof}

\begin{cor}\label{cor:trivially_leq}
	Let $m \leq M$, $\varepsilon_{(1)}$ be the infimum of \cref{eq:spd_fb_appendix} and $\varepsilon_{(2)}$ be the minimum of \cref{eq:spd_standard_appendix}, respectively. Then it holds $\varepsilon_{(1)} \leq \varepsilon_{(2)}$.
\end{cor}
\begin{proof}[Proof of \Cref{cor:trivially_leq}]
	By \cref{lem:solve_standard}, \cref{eq:spd_standard_appendix} is always minimized by a $\Pa \in \R^{m \times n}$. By using \cref{lem:approximator}, since $m \leq M$, there exists ${\Fb}_0 \in \R^{m \times M}$ and ${R}_0 \in \R^{M \times n}$ with $\Pa = {\Fb}_0 \cdot {R}_0$ and $\Vert {R}_0 \Vert_0 = \Vert \Pa \Vert_0 = s$. Thus, $({\Fb}_0, {R}_0)$ is feasible for \cref{eq:spd_fb_appendix} and consequently $\varepsilon_{(1)} \leq \varepsilon_{(2)}$.
\end{proof}

\begin{proof}[Proof of \Cref{thm:spd_appendix}]
	\textbf{First part of proof with $\supp R = \supp \Pa$.}
	W.l.o.g. we assume $\varepsilon_{(1)} = \varepsilon_{(2)}$. By \cref{cor:trivially_leq}, $\varepsilon_{(1)} \leq \varepsilon_{(2)}$ always holds. If $\varepsilon_{(1)} < \varepsilon_{(2)}$, we would be finished with the proof. Furthermore, we assume w.l.o.g. $m = M$, since otherwise we just fill the corresponding entries in $R_{i,j}$ and $\Fb_{k,i}$ for index values $m < i \leq M$ and arbitrary $j,k$ with zeros.
	
	Therefore, let $\varepsilon_{(1)} = \varepsilon_{(2)}$. Let $\Pa \in \R^{m \times n}$ solve \cref{eq:spd_standard_appendix}. Then, there exists an equivalent feasible point $({\Fb}_0, {R}_0) \in \R^{m \times m} \times \R^{m \times n}$ for \cref{eq:spd_fb_appendix}, constructed according to \cref{eq:construct_last,eq:construct_fast} in the proof of \cref{lem:approximator}. \Ie, $\Pa = {\Fb}_0 \cdot {R}_0$ and $\Vert \Pa \Vert_0 = \Vert {R}_0 \Vert_0$. By the assumption $\varepsilon_{(1)} = \varepsilon_{(2)}$, $({\Fb}_0, {R}_0)$ also solves \cref{eq:spd_fb_appendix}. Especially, ${\Fb}_0$ defines a \emph{global} minimum of the smooth, convex function
	\begin{equation}
	f : \R^{m \times m} \rightarrow \R, \Fb \mapsto \Vert U - \Fb \cdot {R}_0\Vert_F^2 \;.
	\end{equation}
	Note, minimizing $f$ is, contrarily to \cref{eq:spd_fb_appendix}, a convex problem. 
	
	A necessary, and by convexity of $f$ even sufficient, condition for ${\Fb}_0$ to minimize $f$ is given by
	\begin{equation}\label{eq:necessary_condition}
	\left. \frac{\partial f}{\partial \Fb}\right \vert_{\Fb = {\Fb}_0} = 0 \in \R^{m \times m}\;.
	\end{equation}
	It holds
	\begin{equation}\label{eq:partial_derivative}
	\frac{\partial f}{\partial \Fb} = \frac{\partial}{\partial \Fb} \Vert U - \Fb \cdot {R}_0 \Vert_F^2 = 2 \cdot (\Fb \cdot {R}_0 - U ) \cdot {R}_0^T \;.
	\end{equation}
	By combining \cref{eq:necessary_condition,eq:partial_derivative}, we get a necessary condition for ${\Fb}_0$ yielding a minimum for $f$, given by
	\begin{equation}\label{eq:necessary_condition2}
	(U - {\Fb}_0 \cdot {R}_0) \cdot {R}_0^T = 0\;.
	\end{equation}
	Consequently, \cref{eq:necessary_condition2} is a necessary condition for $({\Fb}_0, {R}_0)$ to define the minimum for \cref{eq:spd_fb_appendix}.
	From the construction of ${\Fb}_0$ and ${R}_0$ we know that ${\Fb}_0 \cdot {R}_0 = \Pa$, ${\Fb}_0 = \id_{\R^m}$ and ${R}_0 = \Pa$. By \cref{lem:solve_standard}, $\Pa$ is given by $\Pa_{i,j} = \chi_{\{(i,j) \in \tops\}} \cdot U_{i,j}$ with the \emph{characteristic function} $\chi_{\{\cdot \}}$. Combining this with \cref{eq:necessary_condition2} leads to the necessary condition
	\begin{equation}\label{eq:necessary_condition3}
	\hat{U} \cdot \check{U}^T = 0
	\end{equation}
	with
	\begin{equation}
	\hat{U}_{i,j} = \begin{cases} U_{i,j} , \; &\text{if} \; (i,j) \notin \tops \\  0 , \; &\text{if} \; (i,j) \in \tops 
	\end{cases}
	\end{equation}
	and
	\begin{equation}
	\check{U}_{i,j} = \begin{cases} U_{i,j} , \; & \text{if} \; (i,j) \in \tops \\  0 , \;& \text{if} \; (i,j) \notin \tops \;.
	\end{cases}
	\end{equation}
	In the following we will compute an upper bound $\delta$ for the probability $\mathbb{P}(\hat{U} \cdot \check{U}^T = 0)$. By using the fact that $\hat{U} \cdot \check{U}^T = 0$ is a necessary condition for $({\Fb}_0, {R}_0)$ being a minimizer to \cref{eq:spd_fb_appendix}, which is equivalent to $\varepsilon_{(1)} = \varepsilon_{(2)}$, we finally get
	\begin{align}
	\mathbb{P}(\varepsilon_{(1)} < \varepsilon_{(2)}) & = 1 - \mathbb{P}(\varepsilon_{(1)} \geq \varepsilon_{(2)}) \\
	& = 1 - \mathbb{P}(\varepsilon_{(1)} = \varepsilon_{(2)}) \\
	& \geq 1 - \mathbb{P} ( \hat{U} \cdot \check{U}^T = 0) \label{eq:estimation_prob} \;. 
	\end{align}
	Thus, the last step is to find an upper bound $\delta \geq \mathbb{P} ( \hat{U} \cdot \check{U}^T = 0)$. In order to compute $\delta$, we have a closer look on $\hat{U} \cdot \check{U}^T $. It holds
	\begin{align}
	(\hat{U} \cdot \check{U}^T )_{i,j} & = \sum_{k=1}^n \hat{U}_{i,k} \check{U}_{j,k} \\ & = \begin{cases}\sum_{k \in \tij} U_{i,k} U_{j,k} , \; \text{if} \; \tij \neq \emptyset \\ 0 , \; \text{else}\end{cases}\;, 
	\end{align}
	where for each $(i,j) \in \{1, \ldots, m\}^2$, 
	\begin{align}
	\tij := \{k \in \{1, \ldots, n\} : & (i,k) \notin \tops \; \text{and} \notag \\
	& (j,k) \in \tops \} \;.
	\end{align}
	Now assume $S \subset (\{1 ,\ldots, m\} \times \{1, \ldots, n\})^2$ with $S\neq \emptyset$ to be given, then
	\begin{equation}\label{eq:prob_continuous_zero}
	\mathbb{P}(\sum_{(i_1, j_1), (i_2, j_2) \in S} U_{i_1,j_1} \cdot U_{i_2,j_2} = 0) = 0\;.
	\end{equation}
	This equality holds since for each $S \subset (\{1 ,\ldots, m\} \times \{1, \ldots, n\})^2$ with $S\neq \emptyset$, $\sum_{(i_1, j_1), (i_2, j_2) \in S} U_{i_1,j_1} \cdot U_{i_2,j_2}$ follows a continuous probability distribution. 
	
	Consequently,
	\begin{align}
	& &&\mathbb{P}(\hat{U} \cdot \check{U}^T = 0) \\ & = &&\mathbb{P} (\forall (i,j): \tij = \emptyset \notag \\ & &&\lor ( \tij \neq \emptyset \land \sum_{k \in \tij} U_{i,k} \cdot U_{j,k} = 0) ) \\
	& \leq && \mathbb{P} ((\forall (i,j): \tij = \emptyset  ) \notag \\ & &&\lor (\exists S \subset (\{1 ,\ldots, m\} \times \{1, \ldots, n\})^2 \setminus \emptyset: \notag \\ & && \sum_{(i_1, j_1), (i_2, j_2) \in S} U_{i_1,j_1} \cdot U_{i_2,j_2} = 0) ) \\
	& \leq &&  \mathbb{P} (\forall (i,j): \tij = \emptyset)  \notag \\ & && + \left ( \sum_{\substack{S \subset (\{1 ,\ldots, m\} \times \{1, \ldots, n\})^2  \\  S \neq \emptyset}} \right . \notag \\ & && \; \; \; \; \; \left . \mathbb{P}  ( \sum_{(i_1, j_1), (i_2, j_2) \in S} U_{i_1,j_1} \cdot U_{i_2,j_2} = 0) \right) \label{eq:ineq_subad} \\  &= &&\mathbb{P} (\forall (i,j): \tij = \emptyset)\;, \label{eq:equal_zero_prob}
	\end{align}
	where the inequality \eqref{eq:ineq_subad} uses	the subadditivity of probability measures and the final equality \eqref{eq:equal_zero_prob} is achieved by using \cref{eq:prob_continuous_zero}. By looking at the definition of $\tij$, we see that $\forall (i,j): \tij = \emptyset$ only happens if for each $k \in \{1, \ldots, n\}$ either
	\begin{equation}
	\forall i \in \{1, \ldots, m\} : (i,k) \in \tops 	
	\end{equation}
	or
	\begin{equation}
	\forall i \in \{1, \ldots, m\} : (i,k) \notin \tops
	\end{equation}
	holds true. Otherwise, if $k \in \{1, \ldots, n\}$, $i,j \in \{1,\ldots,m\}$ exist with $(i,k) \notin \tops$ and $(j,k) \in \tops$, obviously $\tij \neq \emptyset$. This shows, that $\varepsilon_{(1)} = \varepsilon_{(2)}$ is only possible in the trivial case, where each of the $n$ filters (with $m$ coefficients) is either completely pruned or not pruned at all.
	
	Therefore, we need to compute the probability 
	\begin{equation}\label{eq:total_prob_formula}
	\begin{aligned}
	\mathbb{P} ( \forall k: & &&( \forall i: (i,k) \in \tops )  \\  & \lor && (\forall i : (i,k) \notin \tops)  ) \;.
	\end{aligned}
	\end{equation}
	Due to the \iid assumption of the $U_{i,k}$, all $(i,k)$ have the same probability of being in $\tops$. Thus, deciding $(i,k) \in \tops$ or $(i,k) \notin \tops$ for all $(i,k) \in \{1, \ldots, m\} \times \{1, \ldots, n\}$ together can equivalently be modeled with choosing a subset of size $s$ from a set of size $m \cdot n$, where each subset has the same probability of being sampled, \ie with probability $\frac{1}{\binom{m \cdot n}{s}}$.
	
	Furthermore, \cref{eq:total_prob_formula} is only possible \emph{if} $s = \alpha \cdot m$ for some $\alpha \in \mathbb{N}$. Otherwise, there needs to exists at least one $k,i,j$ such that $(i,k) \notin \tops$ and $(j,k) \in \tops$. Assuming $s = \alpha \cdot m$ for some $\alpha$, there exist exactly $\binom{n}{\alpha}$ different choices to find $\alpha$ many $k$ that satisfy $\forall i: (i,k) \in \tops$ which, by the discussion above, all have similar probability.
	
	Altogether, the probability \cref{eq:total_prob_formula} is given by
	\begin{equation}
	\delta = \begin{cases}
	0 \; & , \; \text{if} \; s \not\equiv 0 (\textrm{mod}\ m)\; \\
	\frac{\binom{n}{\frac{s}{m}}}{\binom{m\cdot n}{s}} \; & , \; \text{if} \; s \equiv 0 (\textrm{mod}\ m)\; 
	\end{cases}\;.
	\end{equation} 
	Finally, 
	\begin{align}
	\mathbb{P}(\hat{U} \cdot \check{U}^T = 0) & \leq && \mathbb{P} (\forall (i,j): \tij = \emptyset) \\ & = && \mathbb{P}(\forall k: ( \forall i: (i,k) \in \tops ) \notag \\ & && \lor (\forall i : (i,k) \notin \tops) ) \\  & \leq && \delta  \;.
	\end{align}
	Using the estimation in \cref{eq:estimation_prob}, we finally get
	\begin{equation}
	\mathbb{P}(\varepsilon_{(1)} < \varepsilon_{(2)}) \geq 1 - \mathbb{P}(\hat{U} \cdot \check{U}^T) \geq 1 - \delta  \;,
	\end{equation}
	which finishes the first part of the proof where $\supp R$ is fixed to be equal to $\supp \Pa$.
	
	\textbf{Second part of proof with arbitrary $\supp R$.}
	As shown in the first part of the proof, \cref{eq:total_prob_formula} is a necessary condition for $\varepsilon_{(1)} = \varepsilon_{(2)}$. This means that all columns of $\Pa$ are either $\Pa_{:,k} = 0 \in \R^m$ or $\Vert \Pa_{:,k} \Vert_0 = m$ which we therefore will assume from now on.\footnote{For a matrix $A \in \R^{d_1 \times d_2}$, the $j$\textsuperscript{th} column $A_{:,j}$ is given by $(A_{i,j})_i \in \R^{d_1}$.}
	
	By assumption, $0 < s$ and therefore, there exists a $k$ with $\Vert \Pa_{:,k} \Vert_0 = m$. Now, set ${\hat{\Fb}}_{:,1} = \Pa_{:,k}$ and ${\hat{\Fb}}_{:,j} = ({\Fb}_0)_{:,j}$ for all other $j$. Then, with $\mathbb{P}=1$, ${\hat{\Fb}}$ still forms a basis. 
	
	Setting ${\hat{R}}_{i,k} = \delta_{i,1}$ for all $i \in \{1, \ldots, m\}$ yields $({\hat{\Fb}} \cdot {\hat{R}})_{:,k} = U_{:,k} = \Pa_{:,k}$. For all other $j \neq k$ with $\Vert \Pa_{:,j} \Vert_0 = m$ there exists a ${\hat{R}}_{:,j}$ with $({\hat{\Fb}} \cdot {\hat{R}})_{:,j} = \Pa_{:,j} = U_{:,j}$ since $\hat{\Fb}$ forms a basis. 
	
	Setting the remaining ${\hat{R}}_{:,j} = 0$ leads to ${\hat{\Fb}} \cdot {\hat{R}} = \Pa$ and $\Vert {\hat{R}} \Vert_0 \leq \Vert \Pa \Vert_0 - (m-1) < \Vert \Pa \Vert_0$. The last inequality holds, since $m > 1$ is assumed. 
	
	Finally, one of the (at least) remaining $m-1$ coefficients which were not spend up to now can be used to better approximate one column of $U$ which is completely zeroed in $\Pa$. Such a column $j_0$ must fulfill $U_{:,j_0} \neq 0$ and $\Pa_{:,j_0}= 0$. Since $s < m \cdot n$ and $U_{i,j}$ \iid $\mathcal{N}(0,1)$, such a column $j_0$ exists with $\mathbb{P} = 1$. Since ${\hat{\Fb}}$ forms a basis, we can find some $l, \l_l$ such that 
	\begin{equation}
	\Vert U_{:,j_0} - \l_l {\hat{\Fb}}_{:,l} \Vert_2 < \Vert U_{:,j_0} \Vert_2\;.
	\end{equation}
	Setting ${\hat{R}}_{l,j_0} = \l_l$ leads to
	\begin{align}
	\Vert U - {\hat{\Fb}} \cdot {\hat{R}} \Vert_F^2 =  & \Vert U - \Pa \Vert_F^2 \\   & -(\Vert U_{:,j_0} \Vert_2^2  - \Vert U_{:,j_0} - \l_l {\hat{\Fb}}_{:,l} \Vert_2^2)  \\  <& \Vert U - \Pa \Vert_F^2\;,
	\end{align}
	which finishes the proof.
\end{proof}

\end{appendices}

\end{document}